\documentclass{article}

\usepackage[eandd,final]{neurips_2026}

\usepackage[utf8]{inputenc}
\usepackage[T1]{fontenc}
\usepackage[colorlinks=true,linkcolor=black,citecolor=black]{hyperref}
\usepackage{placeins}
\usepackage{needspace}
\usepackage{url}
\usepackage{booktabs}
\usepackage{amsfonts}
\usepackage{amsmath}
\let\phi\varphi  
\usepackage{nicefrac}
\usepackage{microtype}
\usepackage[table]{xcolor}
\usepackage{graphicx}
\usepackage{tikz}
\usepackage{array}
\usepackage{tabularx}
\usepackage{makecell}
\newcolumntype{Y}{>{\raggedright\arraybackslash}X}
\title{Different Models, Shared Failures: Auditing Frozen-Encoder Pools for Medical Segmentation}
\hypersetup{pdftitle={Different Models, Shared Failures: Auditing Frozen-Encoder Pools for Medical Segmentation}, pdfauthor={Eunseob Choi and Kyeonghun Kim and Hyuk-Jae Lee and Nam-Joon Kim}}

\author{%
  \textbf{Eunseob Choi}\textsuperscript{1}\quad
  \textbf{Kyeonghun Kim}\textsuperscript{2}\quad
  \textbf{Hyuk-Jae Lee}\textsuperscript{3}\quad
  \textbf{Nam-Joon Kim}\textsuperscript{3}\thanks{Corresponding author.}\\[5pt]
  {\small\textsuperscript{1}Gwangju Institute of Science and Technology}\\[1pt]
  {\small\textsuperscript{2}OUTTA\qquad\textsuperscript{3}Seoul National University}\\[5pt]
  {\footnotesize\texttt{eunseobchoi@gm.gist.ac.kr}\quad\texttt{kyeonghun.kim@outta.ai}}\\[-1pt]
  {\footnotesize\texttt{hjlee@capp.snu.ac.kr}\quad\texttt{knj01@snu.ac.kr}}
}

\DeclareRobustCommand{\mn}{\ensuremath{-}}  
\newcommand{\GapKvasirPrimary}{0.265}
\newcommand{\KKvasir}{11}
\newcommand{\CrossRbarKvasir}{0.675}
\newcommand{\NTestKvasir}{120}
\newcommand{\GapKvasirPrimaryCI}{[0.217, 0.317]}
\newcommand{\GapAcdcPrimary}{0.317}

\newcommand{\KAcdc}{10}
\newcommand{\WithinRbarAcdc}{0.955}
\newcommand{\NTestAcdc}{977}
\newcommand{\GapAcdcPrimaryCI}{[0.289, 0.344]}

\newcommand{\KBrats}{10}
\newcommand{\NTestBrats}{3365}
\newcommand{\GapCupPrimary}{0.557}
\newcommand{\KCup}{11}
\newcommand{\WithinRbarCup}{0.867}
\newcommand{\CrossRbarCup}{0.310}
\newcommand{\NTestCup}{95}
\newcommand{\GapCupPrimaryCI}{[0.500, 0.609]}
\newcommand{\GapDiscPrimary}{0.640}
\newcommand{\KDisc}{11}
\newcommand{\CrossRbarDisc}{0.230}
\newcommand{\GapDiscPrimaryCI}{[0.570, 0.689]}

\newcommand{\DiceCupResnetEighteen}{0.304}
\newcommand{\DiceCupEfficientnetBZero}{0.566}
\newcommand{\KvasirSplitTrainN}{880}
\newcommand{\KvasirSplitTestN}{120}
\newcommand{\BratsTrainPatients}{584}
\newcommand{\BratsTestSlices}{3365}
\newcommand{\BratsTestSessions}{339}
\newcommand{\BratsTestPatients}{146}
\newcommand{\GapKvasirDinoPairsAsCross}{0.300}
\newcommand{\GapCupDinoPairsAsCross}{0.639}
\newcommand{\GapMetaMean}{0.424}
\newcommand{\GapMetaMedian}{0.340}
\newcommand{\GapMetaMin}{0.265}
\newcommand{\GapMetaMax}{0.640}
\newcommand{\KappaNeverFailConvention}{both-never-fail pairs scored $\kappa=1$ as in the official implementation}
\newcommand{\GapKvasirFloorForty}{0.225}
\newcommand{\KKvasirFloorForty}{10}
\newcommand{\GapKvasirLooRange}{0.225--0.308}
\newcommand{\PhiKvasirGapPThirtyThree}{0.290}
\newcommand{\KappaGapKvasirPThirtyThree}{0.329}
\newcommand{\MeffFourPoolKvasirPThirtyThree}{1.559}
\newcommand{\MeffFourPoolKvasirFixedEightyFive}{1.865}
\newcommand{\NAcdcSubjects}{50}
\newcommand{\GapAcdcSubject}{0.493}
\newcommand{\GapAcdcSubjectCI}{[0.391, 0.594]}
\newcommand{\NBratsSubjects}{146}
\newcommand{\GapBratsSubject}{0.204}
\newcommand{\GapBratsSubjectCI}{[0.160, 0.253]}
\newcommand{\GapBratsIccMinusPearson}{0.095}

\newcommand{\GapCupFloorThirtyFive}{0.541}
\newcommand{\KCupFloorThirtyFive}{10}
\newcommand{\GapCupIccMinusPearson}{0.038}
\newcommand{\GapCupLooRange}{0.535--0.653}
\newcommand{\MeffFourPoolCupFixedEightyFiveNSupported}{13{,}440}
\newcommand{\MeffFourPoolCupFixedEightyFiveNPools}{84{,}480}
\newcommand{\PhiDiscGapPThirtyThree}{0.576}
\newcommand{\KappaGapDiscPThirtyThree}{0.582}
\newcommand{\MeffFourPoolDiscPThirtyThree}{2.918}
\newcommand{\MeffFourPoolDiscPThirtyThreeNSupported}{48{,}384}
\newcommand{\MeffFourPoolDiscPThirtyThreeNPools}{84{,}480}
\newcommand{\MeffFourPoolDiscFixedEightyFive}{3.001}
\newcommand{\VitFunctionalKvasir}{5/5}
\newcommand{\VitSeedFloorKvasir}{0.981}
\newcommand{\VitCrossPretrainKvasir}{0.774}
\newcommand{\VitSeedFloorKvasirCI}{[0.975, 0.985]}
\newcommand{\VitCrossPretrainKvasirCI}{[0.708, 0.826]}
\newcommand{\VitCrossRefKvasir}{0.675}
\newcommand{\VitDiffKvasir}{0.207}
\newcommand{\VitDiffKvasirCaseCI}{[0.158, 0.270]}
\newcommand{\VitSeedPairsKvasir}{30}
\newcommand{\VitCrossPretrainPairsKvasir}{160}
\newcommand{\SameBundleSeedPairsKvasir}{66}
\newcommand{\TotalSamePairsKvasir}{82}
\newcommand{\VitFunctionalAcdc}{5/5}
\newcommand{\VitSeedFloorAcdc}{0.993}
\newcommand{\VitCrossPretrainAcdc}{0.779}
\newcommand{\VitSeedFloorAcdcCI}{[0.992, 0.994]}
\newcommand{\VitCrossPretrainAcdcCI}{[0.752, 0.802]}
\newcommand{\VitCrossRefAcdc}{0.638}
\newcommand{\VitDiffAcdc}{0.215}
\newcommand{\VitDiffAcdcCaseCI}{[0.192, 0.240]}
\newcommand{\VitDiffAcdcClusterCI}{[0.187, 0.240]}
\newcommand{\VitSeedPairsAcdc}{30}
\newcommand{\VitCrossPretrainPairsAcdc}{160}
\newcommand{\SameBundleSeedPairsAcdc}{60}
\newcommand{\TotalSamePairsAcdc}{76}
\newcommand{\SamePairSharePercentAcdc}{78.9}
\newcommand{\VitFunctionalBrats}{5/5}
\newcommand{\VitSeedFloorBrats}{0.988}
\newcommand{\VitCrossPretrainBrats}{0.709}
\newcommand{\VitSeedFloorBratsCI}{[0.988, 0.989]}
\newcommand{\VitCrossPretrainBratsCI}{[0.691, 0.728]}
\newcommand{\VitCrossRefBrats}{0.587}
\newcommand{\VitDiffBrats}{0.279}
\newcommand{\VitDiffBratsCaseCI}{[0.261, 0.297]}
\newcommand{\VitDiffBratsClusterCI}{[0.231, 0.337]}
\newcommand{\VitSeedPairsBrats}{30}
\newcommand{\VitCrossPretrainPairsBrats}{160}
\newcommand{\VitFunctionalCup}{5/5}
\newcommand{\VitSeedFloorCup}{0.960}
\newcommand{\VitCrossPretrainCup}{0.358}
\newcommand{\VitSeedFloorCupCI}{[0.943, 0.972]}
\newcommand{\VitCrossPretrainCupCI}{[0.244, 0.473]}
\newcommand{\VitCrossRefCup}{0.310}
\newcommand{\VitDiffCup}{0.601}
\newcommand{\VitDiffCupCaseCI}{[0.492, 0.711]}
\newcommand{\VitSeedPairsCup}{30}
\newcommand{\VitCrossPretrainPairsCup}{160}
\newcommand{\SamePairSharePercentCup}{80.5}
\newcommand{\VitFunctionalDisc}{5/5}
\newcommand{\VitSeedFloorDisc}{0.971}
\newcommand{\VitCrossPretrainDisc}{0.242}
\newcommand{\VitSeedFloorDiscCI}{[0.965, 0.976]}
\newcommand{\VitCrossPretrainDiscCI}{[0.168, 0.348]}
\newcommand{\VitCrossRefDisc}{0.230}
\newcommand{\VitDiffDisc}{0.729}
\newcommand{\VitDiffDiscCaseCI}{[0.623, 0.804]}
\newcommand{\VitSeedPairsDisc}{30}
\newcommand{\VitCrossPretrainPairsDisc}{160}
\newcommand{\PerRaterCupGtConsensusAboveSingleMax}{0.012}
\newcommand{\XckptCrossFamilyRbarKvasir}{0.650}
\newcommand{\XckptGapKvasir}{0.086}
\newcommand{\XckptSameBundleRbarAcdc}{0.989}
\newcommand{\XckptSameFamilyRbarAcdc}{0.745}
\newcommand{\XckptSubjectGapAcdc}{0.315}
\newcommand{\XckptSubjectGapAcdcCI}{[0.224, 0.411]}
\newcommand{\XckptSubjectGapBrats}{0.103}
\newcommand{\XckptSubjectGapBratsCI}{[0.070, 0.138]}
\newcommand{\XckptSameBundleRbarCup}{0.959}
\newcommand{\XckptSameFamilyRbarCup}{0.400}
\newcommand{\XckptGapPrimaryPoolCup}{0.045}
\newcommand{\XckptGapPrimaryPoolCupCI}{[\mn0.044, 0.126]}
\newcommand{\XckptCrossFamilyRbarDisc}{0.194}
\newcommand{\XckptGapDisc}{0.213}
\newcommand{\XckptGapPrimaryPoolDisc}{0.165}
\newcommand{\XckptGapPrimaryPoolDiscCI}{[0.016, 0.281]}
\newcommand{\MedEncPooledKvasirDiff}{\mn0.058}
\newcommand{\MedEncMutualKvasir}{0.647}
\newcommand{\MedEncExtPoolKvasirCross}{0.649}
\newcommand{\MedEncPooledCupDiff}{0.047}
\newcommand{\MedEncExtPoolCupCross}{0.338}
\newcommand{\MedEncPooledDiscDiff}{\mn0.064}
\newcommand{\MedEncMutualDisc}{0.136}
\newcommand{\MedEncExtPoolDiscCross}{0.197}
\newcommand{\PixelGapKvasir}{0.343}
\newcommand{\PixelGapAcdc}{0.394}
\newcommand{\PixelGapAcdcClusterCI}{[0.380, 0.409]}

\newcommand{\PixelGapCup}{0.355}
\newcommand{\PixelGapDisc}{0.443}
\newcommand{\ItemRawGapAcdc}{0.328}
\newcommand{\ItemPanelsAcdc}{70}
\newcommand{\UnetSkipGapKvasir}{0.158}
\newcommand{\UnetSkipGapKvasirCI}{[0.115, 0.208]}
\newcommand{\UnetSkipGapAcdc}{0.084}
\newcommand{\UnetSkipGapAcdcCI}{[0.068, 0.100]}
\newcommand{\UnetSkipWithinRbarCup}{0.806}
\newcommand{\UnetSkipCrossRbarCup}{0.637}
\newcommand{\UnetSkipGapCup}{0.169}
\newcommand{\UnetSkipGapCupCI}{[0.136, 0.209]}
\newcommand{\UnetSkipGapDisc}{0.252}
\newcommand{\UnetSkipGapDiscCI}{[0.169, 0.339]}
\newcommand{\RandInitGapAcdc}{0.133}
\newcommand{\RandInitGapAcdcCI}{[0.089, 0.179]}
\newcommand{\RandInitGapCup}{0.317}
\newcommand{\RandInitGapCupCI}{[0.193, 0.430]}
\newcommand{\PrimaryNineBundleKKvasir}{9}
\newcommand{\PrimaryNineBundleGapKvasir}{0.277}
\newcommand{\PrimaryNineBundleGapKvasirCI}{[0.223, 0.335]}
\newcommand{\PrimaryNineBundleKAcdc}{8}
\newcommand{\PrimaryNineBundleGapAcdc}{0.325}
\newcommand{\PrimaryNineBundleGapAcdcCI}{[0.294, 0.355]}
\newcommand{\PrimaryNineBundleKCup}{9}
\newcommand{\PrimaryNineBundleGapCup}{0.521}
\newcommand{\PrimaryNineBundleGapCupCI}{[0.468, 0.576]}
\newcommand{\PrimaryNineBundleKDisc}{9}
\newcommand{\PrimaryNineBundleGapDisc}{0.635}
\newcommand{\PrimaryNineBundleGapDiscCI}{[0.552, 0.699]}
\newcommand{\NnunetAcdcEpochs}{100}
\newcommand{\NnunetAcdcModels}{10}
\newcommand{\NnunetAcdcNTest}{100}
\newcommand{\NnunetAcdcMeanDice}{0.942}
\newcommand{\NnunetAcdcWithinRbar}{0.957}
\newcommand{\NnunetAcdcCrossRbar}{0.894}
\newcommand{\NnunetAcdcGap}{0.063}
\newcommand{\NnunetAcdcGapCI}{[0.032, 0.094]}
\newcommand{\NnunetAcdcMeff}{1.244}
\newcommand{\NnunetAcdcMeffCI}{[1.148, 1.373]}
\newcommand{\NnunetCupEpochs}{100}
\newcommand{\NnunetCupModels}{10}
\newcommand{\NnunetCupNTest}{95}
\newcommand{\NnunetCupMeanDice}{0.863}
\newcommand{\NnunetCupWithinRbar}{0.825}
\newcommand{\NnunetCupCrossRbar}{0.758}
\newcommand{\NnunetCupGap}{0.067}
\newcommand{\NnunetCupGapCI}{[0.036, 0.119]}
\newcommand{\NnunetCupMeff}{1.264}
\newcommand{\NnunetCupMeffCI}{[1.150, 1.417]}
\newcommand{\FtGapLrEMinusFourKvasir}{0.182}
\newcommand{\FtGapLrEMinusFourKvasirCI}{[0.133, 0.237]}
\newcommand{\FtGapLrEMinusFiveKvasir}{0.238}
\newcommand{\FtGapLrEMinusFiveKvasirCI}{[0.172, 0.306]}
\newcommand{\FtGapExclCollapsedLrEMinusFourKvasirGroupedSens}{0.177}
\newcommand{\FtGapExclCollapsedLrEMinusFourKvasirGroupedSensCI}{[0.114, 0.255]}
\newcommand{\FtWithinRbarLrEMinusFourCup}{0.871}
\newcommand{\FtCrossRbarLrEMinusFourCup}{0.657}
\newcommand{\FtGapLrEMinusFourCup}{0.214}
\newcommand{\FtGapLrEMinusFourCupCI}{[0.171, 0.259]}
\newcommand{\FtWithinRbarLrEMinusFiveCup}{0.864}
\newcommand{\FtCrossRbarLrEMinusFiveCup}{0.581}
\newcommand{\FtGapLrEMinusFiveCup}{0.284}
\newcommand{\FtGapLrEMinusFiveCupCI}{[0.225, 0.347]}
\newcommand{\FtMeanDiceLrEMinusFiveCupEfficientnetBZero}{0.473}
\newcommand{\HeldoutSplitUnitIsic}{image-level analysis of the official ISIC-2018 Task 1 training (2594) / test (1000) split; partial patient identifiers are available}
\newcommand{\HeldoutNTestIsic}{1000}
\newcommand{\HeldoutSplitUnitHippocampus}{volume-level, not patient-level, 208/52 over the 260 training volumes: the 390 MSD volumes are left/right crops of 195 subjects and no subject mapping is released}
\newcommand{\HeldoutNTestHippocampus}{52}
\newcommand{\HeldoutSplitUnitSpleen}{volume-level random split into 33 training and 8 test volumes (seed 42)}
\newcommand{\HeldoutSplitUnitHeart}{volume-level random split into 16 training and 4 test volumes (seed 42)}
\newcommand{\HeldoutKHeart}{0}
\newcommand{\HeldoutKAvailHeart}{11}
\newcommand{\PoolSelSplitsKvasir}{200}
\newcommand{\PoolSelFreqCup}{51.5}
\newcommand{\PoolSelFreqDisc}{2.0}
\newcommand{\PoolSelFreqHippocampus}{1.5}
\newcommand{\PoolSelOverallLow}{910}
\newcommand{\PoolSelOverallSplits}{1400}
\newcommand{\PoolSelOverallUnsupported}{61}
\newcommand{\PoolSelOverallFreq}{65.0}
\newcommand{\PoolSelOverallFreqSupportValid}{68.0}
\newcommand{\SpleenPredHeads}{44}
\newcommand{\SpleenPredHeadsAllEmpty}{38}
\newcommand{\SpleenPredCaseHeads}{352}
\newcommand{\SpleenPredNonemptyCaseHeads}{6}
\newcommand{\SpleenTestCases}{8}
\newcommand{\SpleenGtEmptyCases}{4}
\newcommand{\SpleenEmptyGtDiceOnePairs}{174}
\newcommand{\SpleenNonemptyGtDiceZeroPairs}{176}
\newcommand{\KvasirOfficialTrainN}{880}
\newcommand{\KvasirOfficialTestN}{120}

\newcommand{\KvasirOfficialOverlap}{12}
\newcommand{\KvasirOfficialGap}{0.265}
\newcommand{\KvasirOfficialGapPrimaryCI}{[0.217, 0.317]}
\newcommand{\KvasirOfficialWithinRbar}{0.940}
\newcommand{\KvasirOfficialCrossRbar}{0.675}
\newcommand{\KvasirOfficialK}{11}
\newcommand{\KvasirOfficialXckptGap}{0.086}
\newcommand{\KvasirOfficialXckptGapCI}{[0.047, 0.126]}
\newcommand{\KvasirOfficialMedEncPooledDiff}{\mn0.058}
\newcommand{\KvasirOfficialMedEncPooledDiffCI}{[\mn0.106, \mn0.012]}
\newcommand{\KvasirOfficialMedEncExtPoolGap}{0.302}
\newcommand{\KvasirOfficialMedEncExtPoolGapCI}{[0.254, 0.357]}
\newcommand{\KvasirOfficialUnetSkipGap}{0.158}
\newcommand{\KvasirOfficialUnetSkipGapCI}{[0.115, 0.208]}
\newcommand{\GapKvasirGroupedSens}{0.323}
\newcommand{\GapKvasirGroupedSensCI}{[0.257, 0.403]}
\newcommand{\KKvasirGroupedSens}{11}
\newcommand{\WithinRbarKvasirGroupedSens}{0.933}
\newcommand{\CrossRbarKvasirGroupedSens}{0.610}
\newcommand{\XckptGapKvasirGroupedSens}{0.117}
\newcommand{\XckptGapKvasirGroupedSensCI}{[0.066, 0.169]}
\newcommand{\MedEncPooledKvasirDiffGroupedSens}{\mn0.050}
\newcommand{\MedEncPooledKvasirDiffGroupedSensCI}{[\mn0.099, \mn0.006]}
\newcommand{\MedEncExtPoolKvasirGapGroupedSens}{0.360}
\newcommand{\MedEncExtPoolKvasirGapGroupedSensCI}{[0.288, 0.442]}
\newcommand{\PixelGapKvasirGroupedSens}{0.390}
\newcommand{\PixelGapKvasirGroupedSensCI}{[0.370, 0.412]}
\newcommand{\UnetSkipGapKvasirGroupedSens}{0.173}
\newcommand{\UnetSkipGapKvasirGroupedSensCI}{[0.108, 0.251]}
\newcommand{\UnetSkipKKvasirGroupedSens}{11}
\newcommand{\KvasirOfficialDicePrimaryResnetEighteen}{0.356}
\newcommand{\MaskAlignMaxAbsDelta}{0.0043}
\newcommand{\MaskAlignOffsetPx}{0.47}
\newcommand{\MaskAlignOffsetPxRange}{0.44--0.49}
\newcommand{\GpuHoursPrimary}{8.6}
\newcommand{\GpuHoursXckpt}{4.4}
\newcommand{\GpuHoursMedEnc}{0.3}
\newcommand{\GpuHoursUnetSkip}{25.7}
\newcommand{\GpuHoursFt}{29.3}
\newcommand{\GpuHoursNnunet}{24.2}
\newcommand{\GpuHoursRandInit}{3.5}
\newcommand{\GpuHoursPerRater}{0.2}
\newcommand{\GpuHoursPixelReplay}{0.4}
\newcommand{\GpuHoursHeldout}{0.6}
\newcommand{\GpuHoursMaskAlign}{1.2}
\newcommand{\GpuHoursKvasirOfficial}{0.7}
\newcommand{\GpuHoursOfficialSplits}{15.3}
\newcommand{\GpuHoursTotal}{114.3}
\newcommand{\GpuHoursDistinct}{113.3}
\newcommand{\GpuModel}{NVIDIA A100-SXM4-80GB}
\newcommand{\GpuCount}{7}
\newcommand{\AmosOrgans}{13}

\newcommand{\AmosMinOrgans}{11}
\newcommand{\AmosMaxOrgans}{13}
\newcommand{\AmosCTN}{60}
\newcommand{\AmosCTSameR}{0.988}
\newcommand{\AmosCTCrossR}{0.976}
\newcommand{\AmosCTGap}{0.012}
\newcommand{\AmosCTGapCI}{[0.001, 0.029]}
\newcommand{\AmosCTMeff}{1.071}
\newcommand{\AmosCTThreshold}{0.798}
\newcommand{\AmosCTKappaGap}{0.038}
\newcommand{\AmosCTKappaGapCI}{[\mn0.067, 0.173]}
\newcommand{\AmosCTQGap}{0.004}
\newcommand{\AmosCTQGapCI}{[\mn0.004, 0.027]}
\newcommand{\AmosCTDisagreementGap}{\mn0.017}
\newcommand{\AmosCTDisagreementGapCI}{[\mn0.075, 0.025]}
\newcommand{\AmosCTDoubleFaultGap}{0.008}
\newcommand{\AmosCTDoubleFaultGapCI}{[\mn0.025, 0.050]}
\newcommand{\AmosMRIN}{60}
\newcommand{\AmosMRISameR}{0.975}
\newcommand{\AmosMRICrossR}{0.918}
\newcommand{\AmosMRIGap}{0.057}
\newcommand{\AmosMRIGapCI}{[0.026, 0.105]}
\newcommand{\AmosMRIMeff}{1.268}
\newcommand{\AmosMRIThreshold}{0.748}
\newcommand{\AmosMRIKappaGap}{0.143}
\newcommand{\AmosMRIKappaGapCI}{[\mn0.054, 0.351]}
\newcommand{\AmosMRIQGap}{0.009}
\newcommand{\AmosMRIQGapCI}{[\mn0.061, 0.082]}
\newcommand{\AmosMRIDisagreementGap}{\mn0.083}
\newcommand{\AmosMRIDisagreementGapCI}{[\mn0.192, 0.008]}
\newcommand{\AmosMRIDoubleFaultGap}{\mn0.033}
\newcommand{\AmosMRIDoubleFaultGapCI}{[\mn0.067, \mn0.008]}
\newcommand{\PipelineUnetCupKappaGap}{0.148}
\newcommand{\PipelineUnetCupKappaGapCI}{[0.111, 0.187]}
\newcommand{\PipelineUnetCupQGap}{0.097}
\newcommand{\PipelineUnetCupQGapCI}{[0.059, 0.150]}
\newcommand{\PipelineUnetCupDisagreementGap}{\mn0.070}
\newcommand{\PipelineUnetCupDisagreementGapCI}{[\mn0.089, \mn0.051]}
\newcommand{\PipelineUnetCupDoubleFaultGap}{0.025}
\newcommand{\PipelineUnetCupDoubleFaultGapCI}{[0.017, 0.034]}
\newcommand{\PipelineUnetDiscKappaGap}{0.226}
\newcommand{\PipelineUnetDiscKappaGapCI}{[0.177, 0.273]}
\newcommand{\PipelineUnetDiscQGap}{0.165}
\newcommand{\PipelineUnetDiscQGapCI}{[0.101, 0.258]}
\newcommand{\PipelineUnetDiscDisagreementGap}{\mn0.103}
\newcommand{\PipelineUnetDiscDisagreementGapCI}{[\mn0.127, \mn0.081]}
\newcommand{\PipelineUnetDiscDoubleFaultGap}{0.054}
\newcommand{\PipelineUnetDiscDoubleFaultGapCI}{[0.041, 0.068]}
\newcommand{\PipelineNnunetCupKappaGap}{0.099}
\newcommand{\PipelineNnunetCupKappaGapCI}{[0.043, 0.159]}
\newcommand{\PipelineNnunetCupQGap}{0.028}
\newcommand{\PipelineNnunetCupQGapCI}{[0.008, 0.060]}
\newcommand{\PipelineNnunetCupDisagreementGap}{\mn0.044}
\newcommand{\PipelineNnunetCupDisagreementGapCI}{[\mn0.071, \mn0.019]}
\newcommand{\PipelineNnunetCupDoubleFaultGap}{0.022}
\newcommand{\PipelineNnunetCupDoubleFaultGapCI}{[0.009, 0.036]}
\newcommand{\ScopeKvasirFrozenMeff}{1.559}
\newcommand{\ScopeKvasirFtFourMeff}{1.630}
\newcommand{\ScopeKvasirFtFiveMeff}{1.552}
\newcommand{\ScopeCupFrozenMeff}{2.420}
\newcommand{\ScopeCupFtFourMeff}{1.649}
\newcommand{\ScopeCupFtFiveMeff}{1.747}
\newcommand{\MeffFourPoolKvasirPThirtyThreeTwoDp}{1.56}
\newcommand{\MeffFourPoolDiscPThirtyThreeTwoDp}{2.92}
\newcommand{\KvasirOfficialNearDupGroupsWord}{three}

\newcommand{\VerifCalibClusterCIAcdc}{[0.234, 0.305]}
\newcommand{\VerifBetaClusterCIAcdc}{[0.191, 0.241]}
\newcommand{\VerifKappaClusterCIAcdc}{[0.416, 0.481]}
\newcommand{\VerifCalibClusterCIBrats}{[0.313, 0.430]}
\newcommand{\VerifBetaClusterCIBrats}{[0.194, 0.285]}
\newcommand{\VerifKappaClusterCIBrats}{[0.435, 0.503]}
\newcommand{\VerifClusterWidthRatioMin}{1.2}
\newcommand{\VerifClusterWidthRatioMax}{3.9}
\newcommand{\VerifHierLoWithSelfCup}{0.444}
\newcommand{\VerifHierLoNoSelfCup}{0.409}
\newcommand{\VerifHierLoWithSelfDisc}{0.495}
\newcommand{\VerifHierLoNoSelfDisc}{0.452}
\newcommand{\VerifHierSelfPairMaxDrop}{0.043}
\newcommand{\VerifKappaGapDiscExclConstant}{0.604}
\newcommand{\VerifNConstantMembersDisc}{2}
\newcommand{\VerifResidNullNPerm}{200}

\usepackage{collcell}
\makeatletter
\newcommand{\estci}[1]{\est@ci#1 [\@nil]\est@end}
\def\est@ci#1 [#2]#3\est@end{\ifx\@nil#2\relax\begin{tabular}[t]{@{}c@{}}#1\end{tabular}\else\begin{tabular}[t]{@{}c@{}}#1\\{}[#2]\est@rest#3\est@end\end{tabular}\fi}
\def\est@rest#1[\@nil]#2\est@end{#1}
\newcommand{\estpairs}[1]{\est@pr#1 (\@nil)\est@end}
\def\est@pr#1 (#2)#3\est@end{\ifx\@nil#2\relax#1\else\begin{tabular}[t]{@{}c@{}}#1\\(#2)\end{tabular}\fi}
\makeatother
\newcolumntype{E}{>{\collectcell\estci}c<{\endcollectcell}}
\newcolumntype{P}{>{\collectcell\estpairs}c<{\endcollectcell}}
\makeatletter
\newcommand{\estcik}[1]{\protected@edef\est@tmp{#1}\expandafter\est@k\est@tmp{} [\@nil](\@nil)\est@end}
\def\est@k#1 [#2]#3(#4)#5\est@end{\ifx\@nil#2\relax#1\else\begin{tabular}[t]{@{}c@{}}#1\\{}[#2]\\(#4)\end{tabular}\fi}
\makeatother
\newcolumntype{K}{>{\collectcell\estcik}c<{\endcollectcell}}
\newcolumntype{H}{>{\setbox0=\hbox\bgroup}c<{\egroup}@{}}

\definecolor{TableFocus}{HTML}{E8F0F5}
\definecolor{LinkText}{HTML}{1F4E79}
\hypersetup{urlcolor=LinkText}
\newcommand{\extlink}[2]{\href{#1}{\textcolor{LinkText}{#2}}}

\newcolumntype{T}{l}  
\ExplSyntaxOn
\cs_new_protected:Npn \celllabelkey #1#2
  { \tl_set:Nn #1 {#2} \tl_remove_all:Nn #1 {\relax} }
\ExplSyntaxOff
\makeatletter
\def\ctl@nine{UNet-skip readout, 9 bundles (CLIP excluded)}
\def\ctl@eleven{UNet-skip readout, 11 bundles}
\def\ctl@random{random-initialisation CNN}
\def\ctl@medical{medical-encoder extended pool (15 bundles)}
\newcommand{\ctllabel}[1]{\celllabelkey\ctl@tmp{#1}\ifx\ctl@tmp\ctl@nine UNet-skip (9)\else\ifx\ctl@tmp\ctl@eleven UNet-skip (11)\else\ifx\ctl@tmp\ctl@random Random-init CNN\else\ifx\ctl@tmp\ctl@medical Medical pool\else#1\fi\fi\fi\fi}
\newcommand{\taskshort}[1]{\celllabelkey\task@tmp{#1}\def\task@official{Kvasir (official)}\def\task@grouped{Kvasir (content-group split, sensitivity)}\ifx\task@tmp\task@official Kvasir\else\ifx\task@tmp\task@grouped Kvasir (grouped)\else#1\fi\fi}
\makeatother

\makeatletter
\def\su@core#1; ordered#2\su@end{#1\ifx\relax#2\relax\else)\fi}
\newcommand{\splitunit}[1]{\expandafter\su@core#1; ordered\su@end}
\makeatother
\newcommand{\PaperRepo}{CoFailMed}
\definecolor{PaperLink}{HTML}{244A68}
\hypersetup{colorlinks=true,linkcolor=black,citecolor=black,urlcolor=PaperLink,
  bookmarksopen=true,bookmarksnumbered=true}
\makeatletter
\renewcommand{\@maketitle}{%
  \vbox{\hsize\textwidth\linewidth\hsize
    \vbox to 0pt{%
      \vss
      \hbox to \textwidth{%
        \raisebox{-.5\height}{\includegraphics[height=4.5mm]{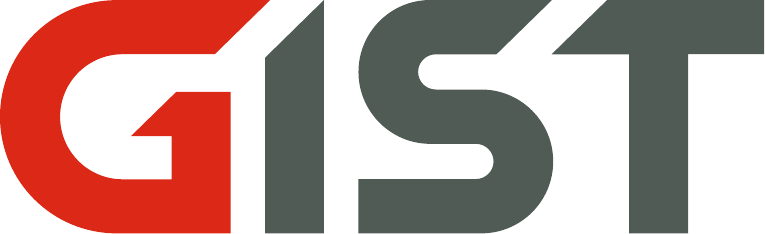}}%
        \hspace{6mm}%
        \raisebox{-.5\height}{\includegraphics[height=8mm]{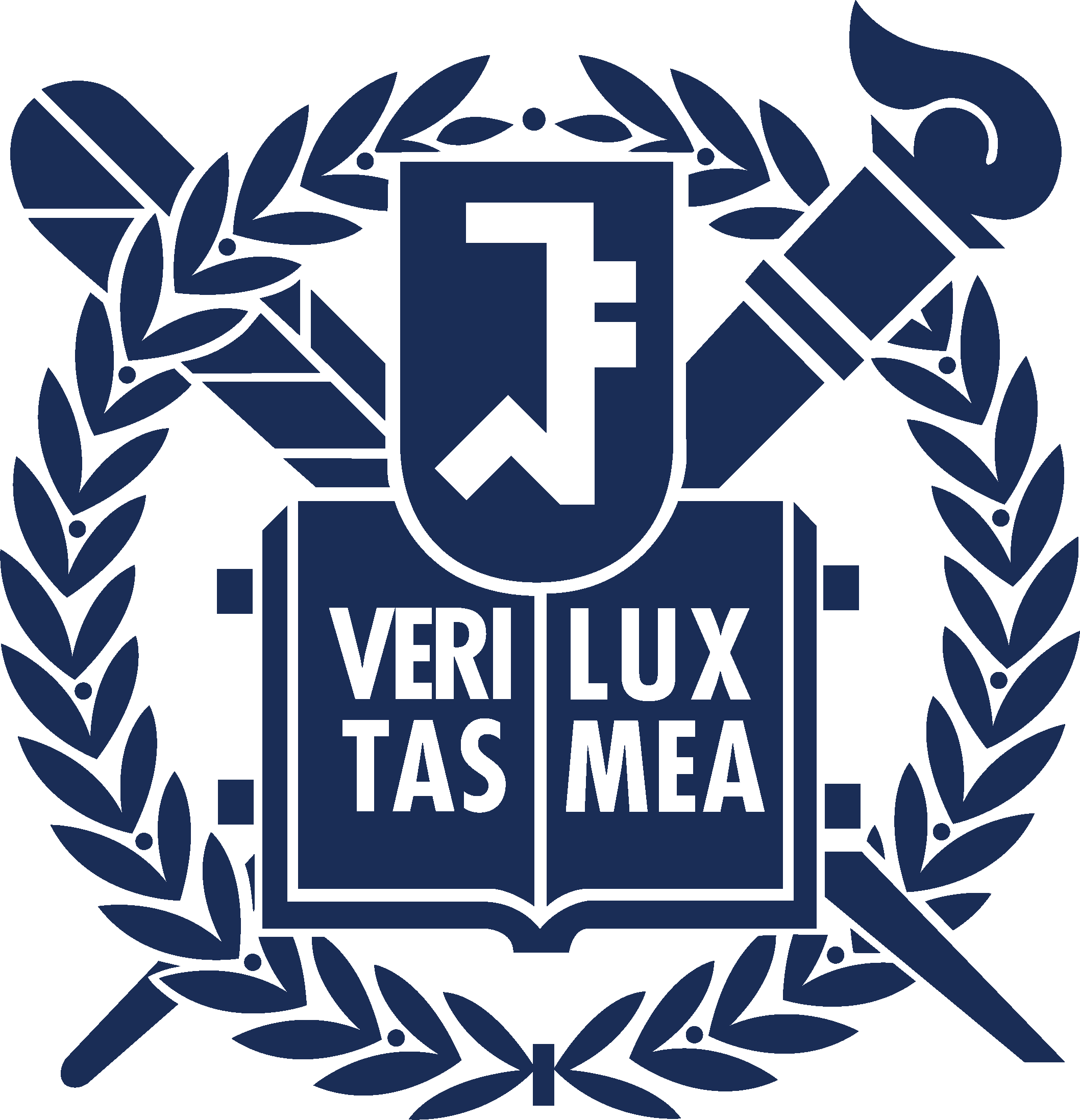}}%
        \hfil
      }%
      \vskip 23pt
    }%
    \vskip 2pt
    \centering
    {\fontsize{17}{20}\selectfont\bfseries\@title\par}
    \vskip 13pt
    {\normalsize\@author\par}
    \vskip 10pt
    {\fontsize{9}{11}\selectfont
      \textbf{Code \& Data:}\enspace
      \href{https://github.com/eunseobchoi/\PaperRepo}{\texttt{github.com/eunseobchoi/\PaperRepo}}\par}
    \vskip 10pt
    {\color{black!25}\hrule height .4pt}
    \vskip 6pt
  }%
}
\long\def\@makecaption#1#2{%
  \vskip\abovecaptionskip
  \sbox\@tempboxa{\small\textbf{#1.}\enspace #2}%
  \ifdim\wd\@tempboxa>\hsize
    {\small\textbf{#1.}\enspace #2\par}%
  \else
    \global\@minipagefalse
    \hb@xt@\hsize{\hfil\box\@tempboxa\hfil}%
  \fi
  \vskip\belowcaptionskip
}
\makeatother

\begin{document}

\maketitle

\begin{abstract}
Average segmentation accuracy and model count do not reveal whether models fail on the same cases. We present CoFailMed, an audit of correlated failures in pretrained-encoder pools for medical segmentation. Across five 2D tasks, we evaluate a controlled sample of 11 public pretrained encoders (not a census) using frozen features and separately trained $1{\times}1$ readouts with four seeds per encoder. Shared-checkpoint pairs have the highest mean per-case Dice correlation. An expanded 17-encoder analysis shows that distinct checkpoints within the same encoder family remain more correlated than cross-family pairs. At the pooled 33rd-percentile Dice threshold, four-member pools drawn from distinct encoders have mean effective sizes of \MeffFourPoolKvasirPThirtyThreeTwoDp--\MeffFourPoolDiscPThirtyThreeTwoDp, estimated from failure correlations in pools with enough failures and passes per member. On the tasks where they were run, richer readouts and fine-tuning of the primary pool narrow the Dice-correlation gap between pairs sharing a starting checkpoint (or the DINOv2-B/S release) and all other pairs. Yet dependence between different encoders can increase, and fine-tuning does not consistently increase effective pool size at 33rd-percentile thresholds recomputed for each training recipe. A smaller between-group gap can coexist with greater dependence. Pool evaluation should therefore report absolute dependence and threshold-specific effective size alongside accuracy and nominal pool size. The CoFailMed release provides aligned per-case traces, analysis scripts and study code to reproduce the central audit results.
\end{abstract}

\section{Introduction}

Medical-segmentation pipelines \citep{isensee2021nnunet,ferreira2024brats} and pathology foundation-model ensembles \citep{luo2025elf} combine architectures or backbones for complementary predictions. However, diversity measures need not track accuracy gains \citep{kuncheva2003diversity}. Mean Dice does not show whether pool members perform poorly on the same cases. We compare per-case Dice and thresholded failures across frozen and adapted encoder pools on matched test cases (Figure~\ref{fig:protocol}).

Models with different names can share checkpoints, pretraining, input normalisation and decoder training. In the expanded 17-bundle pool, we find a hierarchy of dependence: shared-checkpoint seed pairs have the highest mean Dice correlation, distinct checkpoints within a family form a weaker tier, and cross-family dependence varies across tasks. Correlation gaps need not track absolute levels: richer readouts and fine-tuning narrow the shared-checkpoint/release gap while cross-encoder correlation stays high or rises.

We contribute (i) a case-aligned co-failure audit with paired resampling (Section~\ref{sec:formal}), (ii) evidence on how checkpoint sharing, encoder family and adaptation relate to dependence on up to five tasks (Sections~\ref{sec:main}--\ref{sec:controls}), (iii) threshold-specific effective pool sizes (Section~\ref{sec:meff-results}) and (iv) the released traces, splits and code (Section~\ref{sec:artifact}).

\section{Related work}
\label{sec:related}

\paragraph{Monoculture and correlated errors.} Research on algorithmic monoculture examines risks from shared upstream components \citep{kleinberg2021monoculture,bommasani2022picking,toups2023ecosystem} and foundation-model homogenisation \citep[Sec.~1.1]{bommasani2021foundation}. Error consistency measures chance-corrected trial-level agreement \citep{geirhos2020errorconsistency}; other work quantifies error and performance correlations among models sharing algorithms, data or foundation-model components \citep{li2025quantifying}. LLM studies measure correlated errors, output homogeneity and prediction diversity \citep{jiang2025hivemind,kim2025correlated,gorecki2025monoculture,goel2025greatmodels}, while null-model choices affect how agreement is interpreted \citep{subjectivity2026}. In medical segmentation, pairwise Dice between ensemble predictions supports failure detection \citep{roy2019bayesian,zenk2025medseg}. Medical studies evaluate cross-centre robustness \citep{dejong2025pathologyrobust}, clinical benchmarks \citep{campanella2025clinical,neidlinger2025benchmarking}, and complementary domain-specialised features \citep{luo2025elf}. We instead measure case-level co-failure under checkpoint sharing across segmentation pools.

\paragraph{Representation similarity and ensemble diversity.} Checkpoint similarity varies across initialisations and weight-space basins \citep{fort2019loss,wortsman2022soups}. Transfer studies analyse feature reuse and shared prediction errors \citep{neyshabur2020what,gontijolopes2022norep}, and pretraining as a source of diversity \citep{deepensemble_transfer2020}. Representation-similarity studies document alignment across initialisations of the same architecture \citep{kornblith2019similarity,klabunde2025similarity}. Medical ensembles can promote diversity \citep{georgescu2023dipe}; in classification, diversity-promoting interventions can reduce deep-ensemble gains \citep{abe2024pathologies}. Diversity metrics and theory motivate our analysis \citep{kuncheva2003diversity,wood2023diversity}: we correlate per-case Dice scores to characterise shared variation in segmentation quality.

\paragraph{Medical segmentation benchmarks and foundation-model evaluations.} Segmentation benchmarks assess Dice \citep{isensee2021nnunet,isensee2024revisit}, while uncertainty studies assess reliability and calibration \citep{mehrtash2020calibration,jungo2019reliability}; related ensemble and uncertainty methods address classification and anomaly detection \citep{lakshminarayanan2017deepens,gal2016mcdropout,loraensemble2026,d2ue2024}. Failure-detection benchmarks ask whether a method identifies poor segmentations \citep{zenk2025medseg}. We compare failures across models, following case-level evaluation principles \citep{maierhein2024metrics}. Medical segmentation benchmarks \citep{medsegbench2024} and feature-transfer studies \citep{baharoon2023dinov2radiology,yang2023ccfv,kerssies2024benchmark} assess model suitability and performance. SAM-based, volumetric, video and in-context models broaden the candidate pool \citep{sam,huang2024sammedical,medsam2024,sammed2d2023,medicosam2025,gu2025melba,zhu2024medsam2,ma2025medsam2,he2025vista3d,gao2025showandsegment}. Specification-based evaluation assesses robustness against task-specific criteria \citep{xian2025robustness}. Our audit adds measurements of shared failures in frozen-encoder pools.

\section{Dependence measures}
\label{sec:formal}

\begin{figure}[tbp]
  \centering
  \includegraphics[width=\linewidth]{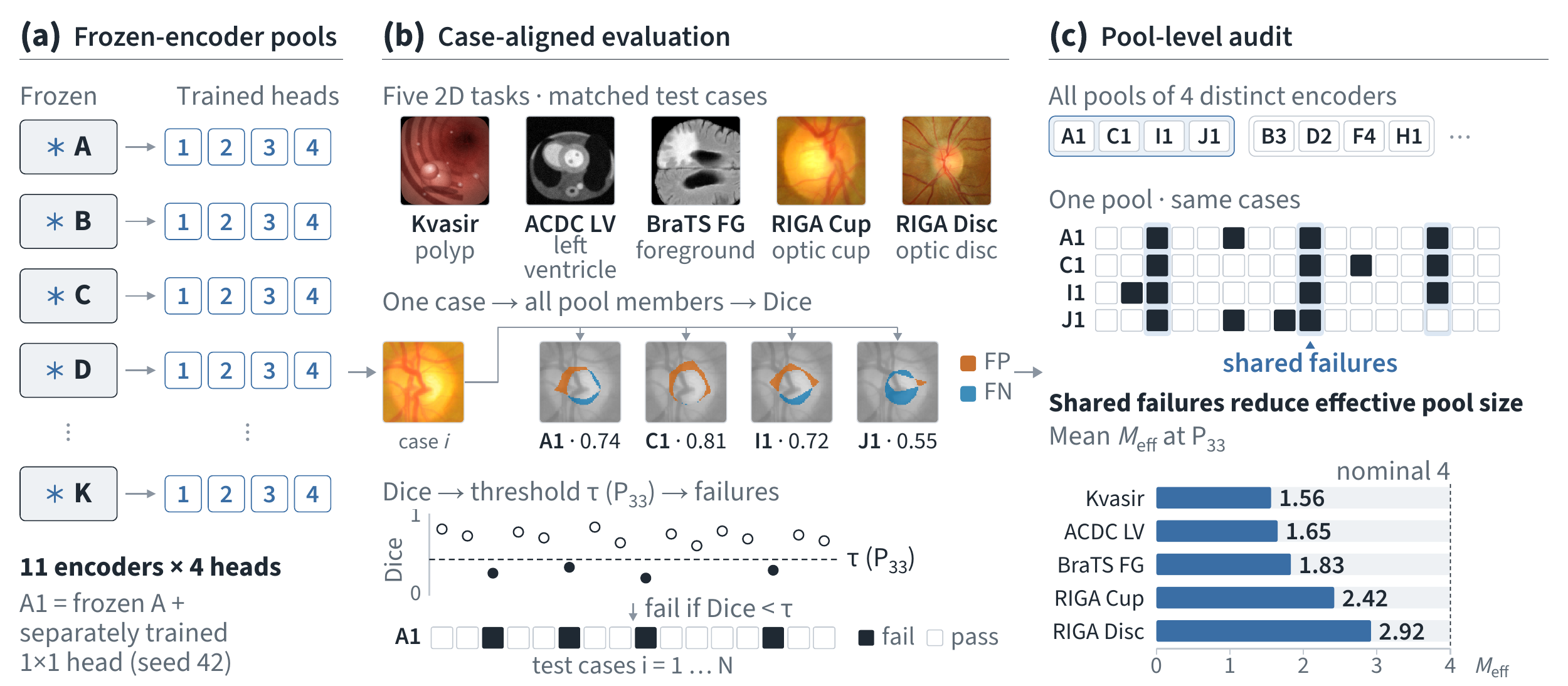}
  \caption{Shared failures reduce effective pool size.
Letters A--K index the 11 frozen encoder checkpoints in the bundle order of Figure~\ref{fig:corr} (A: DINOv2-B, \dots, K: ResNet-18); numbers index head-training seeds 42--45. Panel (b) compares predictions on the same cases (labels give member Dice) and defines failures by thresholding full-mask Dice. Panel (c) enumerates pools of four distinct retained encoders across all seed combinations, with one head per encoder; two example pools are shown. For each pool, positive mean pairwise failure correlation gives an effective size below four. Bars show task-specific mean effective sizes over supported pools at each task's pooled $P_{33}$, on a common 0--4 scale (support counts: Figure~\ref{fig:meff}). Crops show false positives (orange) and false negatives (blue).}
  \label{fig:protocol}
\end{figure}

\paragraph{Bundles and traces.} Figure~\ref{fig:protocol} illustrates the failure-event audit. Let $\mathcal{F}=\{F_1,\dots,F_K\}$ be the pretrained-encoder bundles, each comprising a concrete checkpoint, wrapper, pretraining, input normalisation and feature-extraction path. A pool member combines a bundle with a separately trained decoder head (readout). Trace $(F_k,\sigma)$ denotes the member with decoder seed $\sigma\in\Sigma=\{42,\dots,45\}$ and records Dice $D_{(k,\sigma)}(i)$ on test case $i$ of a task. For distinct traces $a,b$, we compute $r_{a,b}=\mathrm{Pearson}_i[D_a(i),D_b(i)]$ within each task.

\paragraph{Pair groups.} The same-bundle/release group contains pairs of decoders trained with different seeds on the same frozen checkpoint and all DINOv2-B$\times$DINOv2-S pairs. The latter variants share an upstream release, pretraining recipe and ViT-g/14 distillation teacher \citep[Sec.~5]{oquab2024dinov2}. All remaining pairs form the cross-bundle reference. With group means $\rho_{\mathrm{same}}$ and $\rho_{\mathrm{cross}}$ over pairs with defined correlations, the primary gap is $\Delta=\rho_{\mathrm{same}}-\rho_{\mathrm{cross}}$. We apply the mean-Dice floor defined in Section~\ref{sec:design}; $\Delta_f$ denotes the gap at floor $f$, with $\Delta=\Delta_{0.30}$ unless stated. Section~\ref{sec:main} separately analyses shared and distinct checkpoints.

\paragraph{Failure and spatial metrics.} At Dice threshold $\tau$, failure is $Z_a(i;\tau)=\mathbf{1}[D_a(i)<\tau]$; ties are passes. The primary threshold $P_{33}$ is the empirical 33rd percentile of all model-case scores after floor filtering. Failure correlation $\phi$ is the Pearson correlation of the binary indicators; $\kappa$ is their chance-corrected agreement, following the error-consistency convention \citep{geirhos2020errorconsistency}. For $M$ members whose pairwise $\phi$ values have mean $\rho_{\mathrm{fail}}$, the threshold-specific effective size is $M_{\mathrm{eff}}=M/[1+(M-1)\rho_{\mathrm{fail}}]$ (Appendix~\ref{app:meff}); we report it for supported pools, in which every member has at least five failures and five passes. We also measure spatial overlap by averaging pairwise Dice between prediction-error masks over test cases.

\paragraph{Averaging over decoder seeds.} For each pair of distinct bundles, we enumerate all 16 seed combinations (four seeds per bundle). The same-bundle/release group includes six seed pairs per $F_k$ and all 16 DINOv2-B$\times$DINOv2-S pairs. Group means weight all defined trace pairs equally.

\paragraph{Uncertainty.} The primary paired bootstrap ($B{=}2000$) fixes the floor-selected bundle set, resamples aligned test units and recomputes both group means on each draw. We resample images for Kvasir and RIGA, and patients with all their associated slices for ACDC and BraTS. A hierarchical bootstrap sensitivity analysis also resamples bundle/release groups and traces within groups (Appendix~\ref{app:subject-cluster}). For the primary estimates, we report percentile 95\% confidence intervals for the evaluated pools and cohorts. Captions and Appendix~\ref{app:notation} specify undefined-correlation handling and auxiliary resampling units.

\section{Experimental design}
\label{sec:design}

\paragraph{Datasets.} Table~\ref{tab:datasets} summarises the targets, splits and evaluation units for Kvasir-SEG \citep{kvasirseg}, ACDC \citep{acdc}, BraTS 2024 GLI \citep{brats2024gli_challenge} and RIGA+ \citep{docr2022,rigaplus2022,riga,riga_data,messidor2014}. The split files accompany the code, and Appendix~\ref{app:encoders} gives additional data attributions.

Kvasir has no patient/video IDs. All \KvasirOfficialNearDupGroupsWord\ detected near-duplicate groups are in training, and a second group-disjoint split tests partition sensitivity (Appendix~\ref{app:kvasir-official}). ACDC and BraTS use patient-disjoint partitions and enter as 2D slice derivatives (BraTS: FLAIR slices with all annotated foreground, including resection cavities), so the primary study audits a 2D encoder supply chain rather than native-3D segmentation (Section~\ref{sec:controls}). RIGA provides six annotations per image \citep{riga,mrnet2021}; we use their 4-of-6 consensus of unedited rater masks as the Dice reference (Appendix~\ref{app:data-notes}).

\begin{table}[tbp]
  \caption{Datasets, train/test partitions and test units for the five primary tasks.}
  \label{tab:datasets}
  \centering

  \begin{tabularx}{\linewidth}{@{}>{\raggedright\arraybackslash}p{0.18\linewidth}Y>{\raggedright\arraybackslash}p{0.24\linewidth}@{}}
    \toprule
    \bfseries \textbf{Task} & \bfseries \textbf{Target and train/test partition} & \bfseries \textbf{Test units} \\
    \midrule
    Kvasir polyp & Polyp; official \KvasirOfficialTrainN/\KvasirOfficialTestN\ training/validation split, using validation for testing \citep{jha2021kvasirsplit}. & \NTestKvasir\ images \\
    \addlinespace[4pt]
    ACDC LV & LV-positive end-diastolic and end-systolic (ED/ES) slices; official training patients 001--100, test patients 101--150. & \NTestAcdc\ slices; \NAcdcSubjects\ patients \\
    \addlinespace[4pt]
    BraTS foreground (FG) & 2D FLAIR, \texttt{seg>0}; patient-disjoint 80/20 split, with all sessions kept together and up to 10 eligible slices per session. & \NTestBrats\ slices; \BratsTestPatients\ patients, \BratsTestSessions\ sessions \\
    \addlinespace[4pt]
    RIGA Cup/Disc & Optic cup/disc; 4-of-6 rater consensus. Train: BinRushed+MESSIDOR; test: Magrabia. & \NTestCup\ fundus images \\
    \bottomrule
  \end{tabularx}
\end{table}

\paragraph{Encoder bundles.} The 11-bundle primary pool combines foundation-model encoders with conventional pretrained backbones: DINOv2-B/S \citep{oquab2024dinov2}, BiomedCLIP \citep{biomedclip2025}, CLIP ViT-L/14 and ViT-B/16 (CLIP-L/14, CLIP-B/16) \citep{clip2021}, MAE ViT-B/16 (MAE-B) \citep{mae2022}, DeiT ViT-B/16 (DeiT-B) \citep{deit2021}, ConvNeXt-T \citep{convnext2022}, EfficientNet-B0 \citep{efficientnet2019}, and ResNet-50/18 \citep{resnet2016}. Table~\ref{tab:encoders} identifies each wrapper and checkpoint. We evaluate four additional medically pretrained encoders (RETFound \citep{retfound2023}, Phikon \citep{phikon2023}, RAD-DINO \citep{raddino2025} and BiomedParse \citep{biomedparse2025}) under the same frozen protocol on RIGA Cup/Disc and Kvasir, as an encoder-only extension (Section~\ref{sec:controls}).

\paragraph{Protocol.} Images and masks are resized to $224{\times}224$ using bilinear and nearest-neighbour interpolation, respectively; images then receive each encoder's upstream normalisation. We cache FP32 frozen features once per bundle and reuse them across decoder seeds, with TF32 disabled. Each $1{\times}1$ convolutional decoder is trained separately using Adam (learning rate $10^{-3}$), BCE loss, 30 epochs, batch size 16 and seeds 42--45. Evaluation uses the final-epoch weights; pixels are foreground when $\operatorname{sigmoid}(\mathrm{logit})>0.5$. Dice has no smoothing term, and empty-prediction/empty-target pairs score one, following MONAI's \texttt{DiceMetric} with \texttt{ignore\_empty=False} \citep{monai_dicemetric}. All primary targets contain foreground at $224^2$; empty targets occur in the held-out Medical Segmentation Decathlon (MSD) Spleen task \citep{msd2022}, described in Appendix~\ref{app:heldout}. Appendix~\ref{app:encoders} details normalisation, feature taps and mask alignment.

\paragraph{Mean-Dice filtering.} A floor of 0.30 on test mean Dice retains \KKvasir\ bundles on Kvasir, \KAcdc\ on ACDC, \KBrats\ on BraTS and \KCup\ on both RIGA tasks. On ACDC and BraTS, the floor removes only ResNet-18, raising cross-bundle correlation and reducing the gap. The gap stays positive across all tested floors $\{0,0.20,0.25,0.30,0.35,0.40\}$ (Appendix~\ref{app:audit-sensitivity}); Table~\ref{tab:primary-quality} gives each bundle's unfiltered mean Dice.

\section{Results}
\label{sec:main}

\begin{figure}[tbp]
  \centering
  \includegraphics[width=\linewidth]{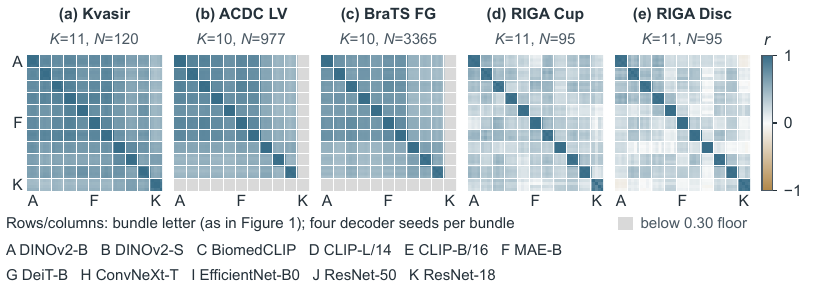}
  \caption{Heads sharing a checkpoint have strongly correlated Dice. Pearson correlations of per-case Dice use a shared scale from $-1$ to $1$, centred at zero. Each bundle (letters A--K, as in Figure~\ref{fig:protocol}) contributes four adjacent rows and columns, one per decoder seed; the key appears below the matrices. Grey denotes bundles excluded by the mean-Dice floor.}
  \label{fig:corr}
\end{figure}

The same-bundle/release group has substantially higher mean per-case Dice correlation than the cross-bundle reference across all five primary tasks (Table~\ref{tab:main}): $\rho_{\mathrm{same}}=\WithinRbarCup$--$\WithinRbarAcdc$ versus $\rho_{\mathrm{cross}}=\CrossRbarDisc$--$\CrossRbarKvasir$, giving $\Delta_{0.30}=\GapMetaMin$--$\GapMetaMax$. Every primary interval excludes zero (case-level for Kvasir and RIGA; patient-clustered for ACDC and BraTS); the narrower slice-level intervals for ACDC and BraTS are sensitivity analyses. Figure~\ref{fig:corr} shows the underlying block structure.

\begin{table}[tbp]
  \caption{Heads sharing a bundle or release are more correlated than cross-bundle pairs on every primary task. $\Delta_{0.30}$: mean per-case Dice correlation of the same-bundle/release group ($\rho_{\mathrm{same}}$) minus that of the cross-bundle reference ($\rho_{\mathrm{cross}}$) after the mean-Dice${\geq}0.30$ floor; $K$: retained bundles (four seeds each); $N$: test units. Case and Cluster CIs: percentile 95\% intervals ($B{=}2000$) from paired case (image or slice) and patient-clustered bootstraps; the latter is primary for ACDC and BraTS, and -- marks image-level tasks. Hier.\ interval: 2.5th--97.5th percentiles of the descriptive case/patient$\times$group$\times$trace resampling sensitivity. Shading marks each primary $\Delta$ and its primary interval.}
  \label{tab:main}
  \centering

  \small\setlength{\tabcolsep}{4pt}
  \begin{tabular}{Trrcccccc}
    \toprule
    \bfseries Task & $K$ & $N$ & $\rho_{\mathrm{same}}$ & $\rho_{\mathrm{cross}}$ & $\Delta_{0.30}$ & \bfseries Case CI & \bfseries Cluster CI & \bfseries Hier.\ interval \\
    \midrule
    Kvasir & 11 & 120 & 0.940 & 0.675 & \cellcolor{TableFocus}0.265 & \cellcolor{TableFocus}[0.217, 0.317] & -- & [0.175, 0.395] \\
\addlinespace[1pt]
ACDC LV & 10 & 977 & 0.955 & 0.638 & \cellcolor{TableFocus}0.317 & [0.297, 0.339] & \cellcolor{TableFocus}[0.289, 0.344] & [0.188, 0.449] \\
\addlinespace[1pt]
BraTS FG & 10 & 3365 & 0.927 & 0.587 & \cellcolor{TableFocus}0.340 & [0.327, 0.353] & \cellcolor{TableFocus}[0.296, 0.388] & [0.234, 0.475] \\
\addlinespace[1pt]
RIGA Cup & 11 & 95 & 0.867 & 0.310 & \cellcolor{TableFocus}0.557 & \cellcolor{TableFocus}[0.500, 0.609] & -- & [0.444, 0.746] \\
\addlinespace[1pt]
RIGA Disc & 11 & 95 & 0.870 & 0.230 & \cellcolor{TableFocus}0.640 & \cellcolor{TableFocus}[0.570, 0.689] & -- & [0.495, 0.838] \\
    \bottomrule
  \end{tabular}
\end{table}

\paragraph{Composition of the same-bundle/release group.} Same-bundle decoder-seed pairs account for \SamePairSharePercentAcdc--\SamePairSharePercentCup\% of this group (Kvasir \SameBundleSeedPairsKvasir/\TotalSamePairsKvasir; ACDC \SameBundleSeedPairsAcdc/\TotalSamePairsAcdc) and share identical frozen encoder weights. The remaining pairs are DINOv2-B$\times$DINOv2-S. Moving them to the cross group changes the gap to \GapCupDinoPairsAsCross\ on RIGA Cup and \GapKvasirDinoPairsAsCross\ on Kvasir (Appendix~\ref{app:cross-checkpoint}, Table~\ref{tab:primary-pool-decomp}).

\begin{figure}[tbp]
  \centering
  \includegraphics[width=\linewidth]{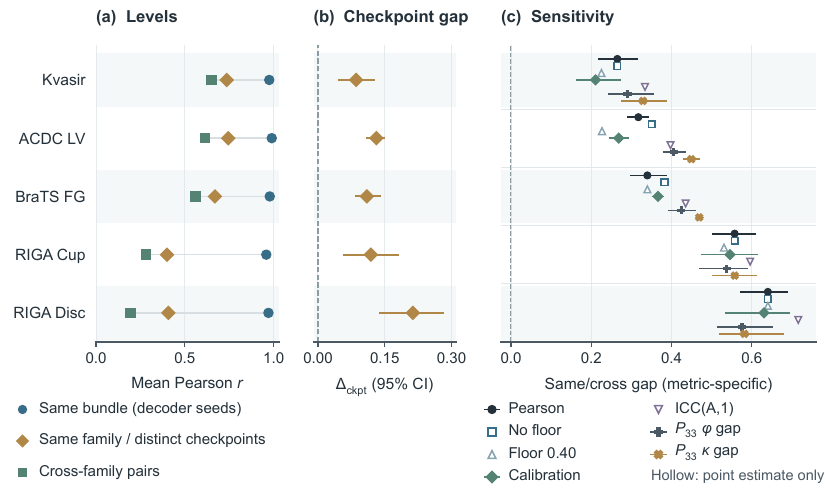}
  \caption{Checkpoint hierarchy and sensitivity to the dependence measure. (a) Correlation levels on the 17-bundle pool (point estimates only). (b) Distinct-checkpoint gaps with 95\% bootstrap CIs. (c) Same/cross gaps under alternative metrics and pool definitions; hollow markers have no CI. Calibration and $\kappa$ intervals resample slices on ACDC/BraTS; their patient-clustered sensitivities are in Appendix~\ref{app:sensitivity}. All gaps are positive, but magnitudes across metrics are not directly comparable.}
  \label{fig:tiers}
\end{figure}

\paragraph{Distinct checkpoints.} We examine distinct checkpoints separately in a 17-bundle pool that adds DINOv2-L, CLIP-B/32, ResNet-34/101 and ConvNeXt-S/B (Table~\ref{tab:distinct-ckpt}, Figure~\ref{fig:tiers}a). Mean same-bundle correlations range from \XckptSameBundleRbarCup\ to \XckptSameBundleRbarAcdc. Same-family pairs with distinct checkpoints have mean correlations of $\rho=\XckptSameFamilyRbarCup$--$\XckptSameFamilyRbarAcdc$, compared with \XckptCrossFamilyRbarDisc--\XckptCrossFamilyRbarKvasir\ across families. Their gaps relative to cross-family pairs, $\Delta_{\mathrm{ckpt}}=\XckptGapKvasir$--$\XckptGapDisc$, are positive but smaller than those of same-bundle pairs (per-task intervals in Table~\ref{tab:distinct-ckpt}). This cross-family reference excludes all same-family pairs, unlike the primary cross-bundle reference, which includes them and uses a smaller pool.

Within the 11-bundle primary pool, $\Delta_{\mathrm{ckpt}}$ ranges from \XckptGapPrimaryPoolCup~\XckptGapPrimaryPoolCupCI\ on RIGA Cup, the only interval that includes zero (the expanded-pool Cup interval excludes it), to \XckptGapPrimaryPoolDisc~\XckptGapPrimaryPoolDiscCI\ on RIGA Disc. The expanded pool follows the checkpoint hierarchy in Figure~\ref{fig:tiers}a.

\paragraph{Sensitivity to metrics and pool selection.} Correlation and agreement gaps remain positive across the examined variants; disagreement gaps are correspondingly negative. In Table~\ref{tab:main-robustness}, the Pearson gap stays positive with no floor and at a 0.40 floor on every task, and the ICC(A,1) \citep{mcgraw1996forming} gap exceeds the Pearson gap by \GapCupIccMinusPearson--\GapBratsIccMinusPearson. Leave-one-bundle-out ranges remain positive (RIGA Cup \GapCupLooRange; Kvasir \GapKvasirLooRange), as do patient-aggregated gaps on the two slice-level tasks (Appendix~\ref{app:subject-cluster}). RIGA Cup illustrates sensitivity to pool membership: ResNet-18 narrowly clears the 0.30 floor (mean Dice \DiceCupResnetEighteen, $K{=}\KCup$), but drops out at 0.35 ($K{=}\KCupFloorThirtyFive$), leaving a positive gap of \GapCupFloorThirtyFive. Failure-event gaps are also positive: $P_{33}$ gaps are \PhiKvasirGapPThirtyThree--\PhiDiscGapPThirtyThree\ for $\phi$ and \KappaGapKvasirPThirtyThree--\KappaGapDiscPThirtyThree\ for error-consistency $\kappa$, with all intervals excluding zero (Table~\ref{tab:failure-events}). Table~\ref{tab:main-robustness} also reports disjoint-case calibration selection (Figure~\ref{fig:tiers}c) and pair-quality adjustment.

\paragraph{Consensus ground truth.} A shared consensus boundary could make cases uniformly hard. Against each individual RIGA rater, the gap stays positive, and with pools fixed by the consensus floor, the consensus gap lies within the single-rater range on Cup and Disc (Table~\ref{tab:per-rater}). The consensus reference alone therefore does not produce the gap, although shared annotation boundaries may still contribute.

\begin{figure}[tbp]
  \centering
  \includegraphics[width=\linewidth]{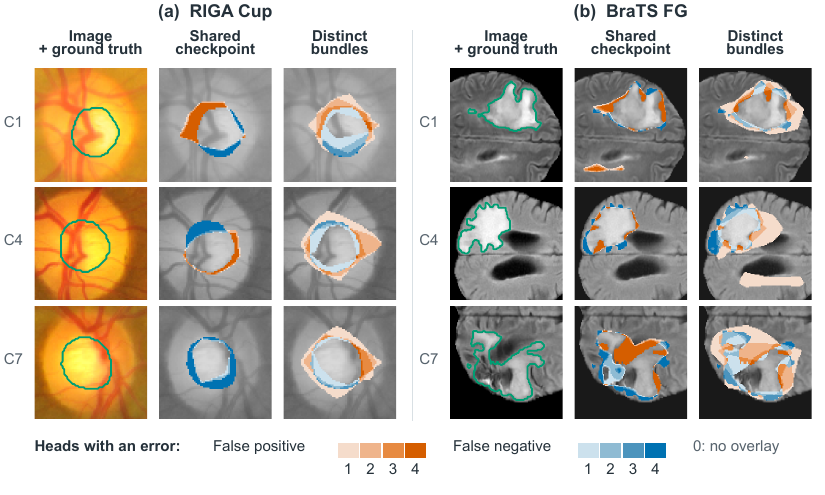}
  \caption{Spatial errors recur across shared-checkpoint heads. Rows show three of the eight area-selected candidate cases (C1, C4, C7; Appendix~\ref{sec:pixerror-app}) for RIGA Cup (a) and BraTS FG (b). Shared: DINOv2-B seeds 42--45; distinct: ResNet-50, EfficientNet-B0, CLIP-L/14 and BiomedCLIP at seed 42. At each pixel, colour intensity counts heads with a false-positive (orange) or false-negative (blue) error; green marks the reference boundary in each panel's first column. Each case uses a common crop. Figure~\ref{fig:spatial-details} and Table~\ref{tab:pixel-error-overlap} give individual examples and cohort-level overlap estimates.}
  \label{fig:hero}
\end{figure}

\paragraph{Spatial overlap.} Shared-checkpoint heads also make spatially overlapping errors (Figure~\ref{fig:hero}). Error masks are the prediction--ground-truth XOR at $224{\times}224$, and overlap is pairwise error-mask Dice averaged over cases. The same-bundle/release-minus-cross gap in error-mask Dice is \PixelGapCup\ on RIGA Cup, \PixelGapDisc\ on RIGA Disc, \PixelGapKvasir\ on official Kvasir and \PixelGapAcdc\ on ACDC LV; all gap intervals exclude zero. BraTS FG appears only in the examples; Appendix~\ref{sec:pixerror-app} details the cohort analysis and example selection.

\paragraph{Item-difficulty sensitivity.} Co-failure statistics depend on the null model \citep{subjectivity2026}. For each of the 70 partitions of the eight upstream families into four reference and four evaluation families, we regress the evaluation traces on the reference families' mean Dice, a common case-difficulty proxy. Averaged over partitions, the residual correlation gap exceeds the unadjusted gap of the same evaluation families on every task (by 0.071--0.417) and is positive in every partition, so this difficulty proxy does not explain the gap (exploratory; Appendix~\ref{app:diffleak}).

\section{Controls and extensions}
\label{sec:controls}
\label{sec:evidence-taxonomy}
\label{sec:arch-vs-pretrain}
\label{sec:samebb-probes}

These controls test how correlation gaps relate to absolute dependence and effective size.

\paragraph{Shared architecture family versus shared checkpoint.} Following representation-similarity work \citep{kornblith2019similarity}, we compare five ViT-B-class bundles (DINOv2-B, BiomedCLIP, MAE-B, DeiT-B and CLIP-B/16) using the same frozen-feature $1{\times}1$ readout architecture. All five pass the mean-Dice floor on every task and share a broad architecture class, while their pretraining differs (Table~\ref{tab:arch-confound}). On RIGA, cross-pretraining correlation is close to the cross-bundle reference and far below the within-bundle seed correlation; on ACDC, Kvasir and BraTS it lies between the two. Training ConvNeXt-T and ResNet-50 from scratch also gives positive gaps between same-architecture and cross-architecture pairs: \RandInitGapAcdc~\RandInitGapAcdcCI\ on ACDC LV and \RandInitGapCup~\RandInitGapCupCI\ on RIGA Cup. With eight seeds per backbone (16 models per task), this control shows dependence without pretraining, but its two backbones and two tasks are not matched to the 11-bundle pool. These contrasts do not causally separate architecture, pretraining corpus, checkpoint lineage, normalisation, optimisation and decoder recipe.

\begin{table}[tbp]
  \caption{ViT-B-class bundles with different pretraining are less correlated than heads sharing a checkpoint. DINOv2-B, BiomedCLIP, MAE-B, DeiT-B and CLIP-B/16 share the frozen $1{\times}1$ readout. Seed correlation: mean cross-seed correlation within each of the five bundles; ViT cross-pretrain: mean correlation between different ViT bundles; cross-bundle ref.: from Table~\ref{tab:main}; $\Delta_{\mathrm{ViT}}$: seed correlation minus ViT cross-pretrain, with case-bootstrap 95\% CI (Table~\ref{tab:arch-confound-3task}: clustered intervals, pair counts).}
  \label{tab:arch-confound}
  \centering

  \small
  \begin{tabular*}{\linewidth}{@{\extracolsep{\fill}}lccccc@{}}
    \toprule
    \bfseries Statistic & \bfseries Kvasir & \bfseries ACDC LV & \bfseries BraTS FG & \bfseries RIGA Cup & \bfseries RIGA Disc \\
    \midrule
    Seed correlation & \VitSeedFloorKvasir & \VitSeedFloorAcdc & \VitSeedFloorBrats & \VitSeedFloorCup & \VitSeedFloorDisc \\
    \addlinespace[2pt]
    ViT cross-pretrain & \VitCrossPretrainKvasir & \VitCrossPretrainAcdc & \VitCrossPretrainBrats & \VitCrossPretrainCup & \VitCrossPretrainDisc \\
    \addlinespace[2pt]
    Cross-bundle ref. & \VitCrossRefKvasir & \VitCrossRefAcdc & \VitCrossRefBrats & \VitCrossRefCup & \VitCrossRefDisc \\
    \addlinespace[2pt]
    \makecell[l]{$\Delta_{\mathrm{ViT}}$\\{}[95\% CI]} & \cellcolor{TableFocus}\makecell[c]{\VitDiffKvasir\\\VitDiffKvasirCaseCI} & \cellcolor{TableFocus}\makecell[c]{\VitDiffAcdc\\\VitDiffAcdcCaseCI} & \cellcolor{TableFocus}\makecell[c]{\VitDiffBrats\\\VitDiffBratsCaseCI} & \cellcolor{TableFocus}\makecell[c]{\VitDiffCup\\\VitDiffCupCaseCI} & \cellcolor{TableFocus}\makecell[c]{\VitDiffDisc\\\VitDiffDiscCaseCI} \\
    \bottomrule
  \end{tabular*}
\end{table}

\begin{figure}[tbp]
  \centering
  \includegraphics[width=\linewidth]{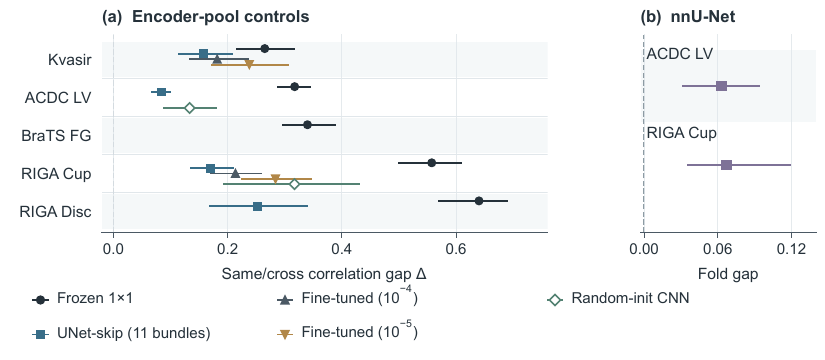}
  \caption{Richer readouts reduce the correlation gap. (a) Frozen, UNet-skip, fine-tuned and random-initialisation controls with 95\% bootstrap CIs. (b) The nnU-Net fold gap $\Delta_{\mathrm{fold}}$ (same-fold minus different-fold pairs) is a different estimand, shown on a separate axis scale. Hollow diamonds mark the random-initialisation CNN (no pretrained weights), whose same/cross groups are same-architecture and cross-architecture pairs; absent markers indicate controls not run; absolute correlation levels are needed to interpret a smaller gap.}
  \label{fig:readout}
\end{figure}

\paragraph{Decoder capacity.} A multi-scale UNet-skip decoder \citep{ronneberger2015unet} narrows the same/cross gap relative to the frozen $1{\times}1$ readout. Across all 11 source bundles, the gaps are \UnetSkipGapKvasir~\UnetSkipGapKvasirCI\ on Kvasir, \UnetSkipGapAcdc~\UnetSkipGapAcdcCI\ on ACDC, \UnetSkipGapCup~\UnetSkipGapCupCI\ on RIGA Cup and \UnetSkipGapDisc~\UnetSkipGapDiscCI\ on RIGA Disc (BraTS not run; Figure~\ref{fig:readout}a; Appendix~\ref{app:unet-skip}). The source bundles match Table~\ref{tab:main}, but UNet-skip retains one additional bundle on ACDC LV. Table~\ref{tab:controls} also reports matched nine-bundle comparisons excluding both OpenAI CLIP bundles. The proportional gap reduction is largest on ACDC and smallest on Kvasir. Richer readouts can thus narrow the gap while raising cross-bundle Dice correlation.

\paragraph{End-to-end fine-tuning.} We unfreeze the same 11 encoders at learning rates $10^{-4}$ and $10^{-5}$, retaining FP32, the readout architecture, epochs, batch size, seeds and splits. The RIGA Cup gap falls to \FtGapLrEMinusFourCup~\FtGapLrEMinusFourCupCI\ and \FtGapLrEMinusFiveCup~\FtGapLrEMinusFiveCupCI; both paired intervals for the gap reduction exclude zero. On official Kvasir, the gaps are \FtGapLrEMinusFourKvasir~\FtGapLrEMinusFourKvasirCI\ and \FtGapLrEMinusFiveKvasir~\FtGapLrEMinusFiveKvasirCI, with only the $10^{-4}$ paired interval for the gap reduction excluding zero (Figure~\ref{fig:readout}a; Appendix~\ref{app:ft}).

The smaller Cup gap accompanies higher cross-bundle dependence. Same/cross correlations are \WithinRbarCup/\CrossRbarCup\ for the frozen readout, \UnetSkipWithinRbarCup/\UnetSkipCrossRbarCup\ for UNet-skip, and \FtWithinRbarLrEMinusFourCup/\FtCrossRbarLrEMinusFourCup\ and \FtWithinRbarLrEMinusFiveCup/\FtCrossRbarLrEMinusFiveCup\ after fine-tuning at $10^{-4}$ and $10^{-5}$. At the pooled $P_{33}$ computed separately for each training configuration, mean four-member effective size on Kvasir is \ScopeKvasirFrozenMeff\ for the frozen readout versus \ScopeKvasirFtFourMeff\ and \ScopeKvasirFtFiveMeff\ after fine-tuning; on Cup it falls from \ScopeCupFrozenMeff\ to \ScopeCupFtFourMeff\ and \ScopeCupFtFiveMeff\ (Table~\ref{tab:adaptation-meff}).

\paragraph{Medically pretrained encoders.} Mean correlation between the four added medical encoders and the primary pool is lower than the primary cross-bundle reference on Disc and Kvasir; the Cup difference interval includes zero (Table~\ref{tab:medenc-per-encoder}). Adding RETFound, Phikon, RAD-DINO and BiomedParse also lowers the extended pool's cross-bundle correlation on Disc (\CrossRbarDisc$\to$\MedEncExtPoolDiscCross) and Kvasir (\CrossRbarKvasir$\to$\MedEncExtPoolKvasirCross), but raises it on Cup (\CrossRbarCup$\to$\MedEncExtPoolCupCross). The gap increases on Disc and Kvasir and remains close to its primary value on Cup (Table~\ref{tab:medenc}): lower cross-bundle dependence can accompany a larger gap. These are comparisons of encoder bundles, which differ in architecture, objective and resolution. RAD-DINO shares DINOv2-B lineage, and BiomedParse uses related optic-disc/cup and polyp training data.

\paragraph{nnU-Net pipeline.} We train nnU-Net v2 \citep{isensee2021nnunet} on the RIGA Cup training sources with its self-configuring 2D pipeline and the primary $224{\times}224$ inputs (\NnunetCupEpochs\ epochs; five folds $\times$ two seeds, \NnunetCupModels\ models). All models use the primary Dice convention on the same \NnunetCupNTest\ Magrabia images, with mean Dice \NnunetCupMeanDice. Mean per-case Dice correlation is \NnunetCupWithinRbar\ within folds and \NnunetCupCrossRbar\ across folds, giving a fold gap $\Delta_{\mathrm{fold}}=\NnunetCupGap$~\NnunetCupGapCI\ (case bootstrap); the five fold models of seed 13 have $M_{\mathrm{eff}}=\NnunetCupMeff$ at their own $P_{33}$. The ACDC 3D counterpart gives correlations of \NnunetAcdcWithinRbar\ within folds and \NnunetAcdcCrossRbar\ across folds, with $\Delta_{\mathrm{fold}}=\NnunetAcdcGap$~\NnunetAcdcGapCI\ (patient-cluster bootstrap). These fold gaps estimate a different contrast from the same/cross gap in Table~\ref{tab:main} (Figure~\ref{fig:readout}b; Appendix~\ref{app:nnunet}).

\paragraph{Volumetric abdominal segmentation.} On \AmosCTN\ CT and \AmosMRIN\ MRI volumes from AMOS22 \citep{ji2022amos}, we compare models from two MRSegmentator folds \citep{haentze2025mrsegmentator} with the modality-specific TotalSegmentator model \citep{wasserthal2023totalsegmentator,akinci2025totalsegmentatormri}. Same-system/cross-system correlations are \AmosCTSameR/\AmosCTCrossR\ on CT and \AmosMRISameR/\AmosMRICrossR\ on MRI. Despite small gaps ($\Delta_{\mathrm{sys}}=\AmosCTGap$~\AmosCTGapCI; \AmosMRIGap~\AmosMRIGapCI), three-member effective sizes at $P_{33}$ are \AmosCTMeff\ and \AmosMRIMeff\ (Appendix~\ref{app:amos}). The comparison covers two released systems within the nnU-Net family, and MRSegmentator training partly uses TotalSegmentator-assisted labels. Failure-event gaps vary by metric, including a negative MRI double-fault (joint-failure) gap. By contrast, in the primary pool and in UNet-skip Cup/Disc and nnU-Net Cup on the same test cases, $\kappa$, Yule's $Q$ and double-fault gaps are positive and disagreement gaps are negative (Tables~\ref{tab:failure-events} and \ref{tab:pipeline-metrics}; definitions in Appendix~\ref{app:notation}).

\section{Effective pool size}
\label{sec:meff-results}

\begin{figure}[tbp]
  \centering
  \includegraphics[width=\linewidth]{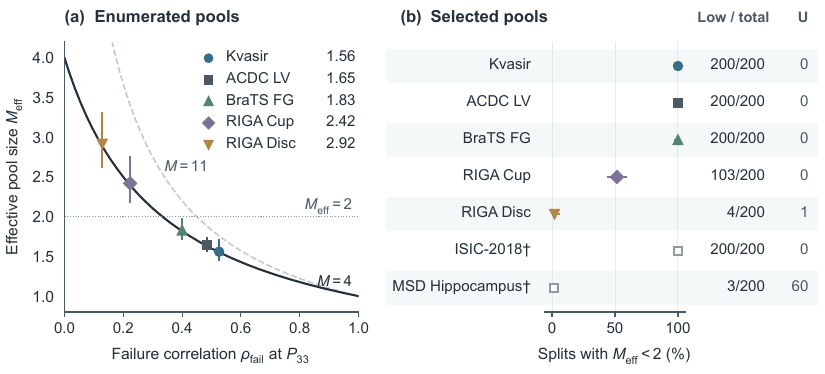}
  \caption{Effective size of evaluated pools. (a) Means over pools of four distinct bundles, with 95\% CIs conditional on observed $P_{33}$ and retained bundles. Curves show the formula for $M{=}4$ and 11; averaging pool-specific effective sizes can place points above the curve. RIGA Disc supports \MeffFourPoolDiscPThirtyThreeNSupported/\MeffFourPoolDiscPThirtyThreeNPools\ pools; other tasks support all pools. (b) Calibration-selected pools: frequency of $M_{\mathrm{eff}}<2$, with 95\% Clopper--Pearson intervals over random splits conditional on the fixed cohort (Table~\ref{tab:poolsel}). Labels give low/total and unsupported (U) counts; unsupported splits enter the total but not the low count. $\dag$: held-out task.}
  \label{fig:meff}
\end{figure}

\paragraph{Enumerated pools.} The equicorrelation effective size $M_{\mathrm{eff}}$ (Section~\ref{sec:formal}; Appendix~\ref{app:meff}) is a variance-equivalent size for equal-variance or standardised failure indicators, not a literal count of independent models. We average its defined values over supported pools of four distinct bundles and one seed per bundle. At pooled $P_{33}$, the task means range from \MeffFourPoolKvasirPThirtyThreeTwoDp\ (Kvasir) to \MeffFourPoolDiscPThirtyThreeTwoDp\ (RIGA Disc), substantially below four (Figure~\ref{fig:meff}a).

\paragraph{Calibration-selected pools.} Substantial failure dependence remains in many pools selected from distinct encoder families. We select four family representatives on calibration data and evaluate them on the evaluation halves of the five primary tasks, ISIC-2018 \citep{isic2019,ham10000,isic_msk2017} and MSD Hippocampus \citep{msd2022,msd2019simpson}, using \PoolSelSplitsKvasir\ random splits per task. Evaluation-half $M_{\mathrm{eff}}<2$ occurs in every split on Kvasir, ACDC, BraTS and ISIC-2018, in \PoolSelFreqCup\% on Cup, in \PoolSelFreqDisc\% on Disc and in \PoolSelFreqHippocampus\% on Hippocampus, where 60 of \PoolSelSplitsKvasir\ splits are unsupported. Overall, \PoolSelOverallLow/\PoolSelOverallSplits\ splits have low effective size (\PoolSelOverallFreq\%, counting \PoolSelOverallUnsupported\ unsupported splits as not low; \PoolSelOverallFreqSupportValid\% among supported splits). Spleen and Heart are excluded because Spleen scores are dominated by empty targets and no Heart bundle passes the floor (Appendix~\ref{app:heldout}).

\paragraph{Selection rule.} Within each calibration split, we retain bundles passing the 0.30 mean-Dice floor, choose each upstream family's best member by calibration mean Dice, and take the top four representatives. The families are DINOv2, OpenAI CLIP, BiomedCLIP, MAE, DeiT, ResNet, ConvNeXt and EfficientNet, so a selected pool can contain several ViTs. The failure threshold is the calibration-best member's $P_{33}$, held fixed on the evaluation half. This differs from the pooled model-case $P_{33}$ used in the primary analysis. ACDC and BraTS are split by patient but retain slice-level failure indicators. ISIC uses image-level splits and has known train/test patient overlap; Hippocampus is split by volume, without established patient independence (Appendix~\ref{app:heldout}).

\section{Discussion}
\label{sec:discussion}

\paragraph{Using the audit.} For retrospective local quality assurance and procurement, select candidates meeting a stated quality criterion on calibration data, then report same-bundle/release and cross-bundle correlation levels on aligned held-out cases, alongside joint-failure rates, threshold-specific $M_{\mathrm{eff}}$ and support counts. A small gap between two correlations near one indicates redundancy; the same gap between correlations near zero indicates little linear association. Whether diverse pools keep mean member Dice at or above $\tau$ more often than same-bundle pools when a member fails depends strongly on $\tau$ (Appendix~\ref{app:tau-sweep}).

\paragraph{Scope and limitations.} The 11 bundles were selected for public checkpoints, compatible dense features and complete four-seed traces; they are not a census of available encoders. Effect sizes and intervals are therefore conditional on these pools, cohorts and readouts; with eight upstream families in the primary pool, the intervals describe case-level uncertainty, not variation over future encoders. Compared bundles also differ in pretraining, normalisation and training recipe (Section~\ref{sec:controls}), and overlap between test images and pretraining corpora is untested. RIGA Cup/Disc share images. Per-task intervals do not constitute a joint significance test. Fine-tuning and nnU-Net use reduced budgets. These retrospective benchmark results do not validate clinical safety or outcomes, and we do not assess demographic fairness.

\paragraph{Code and data availability.}\label{sec:artifact} The \href{https://github.com/eunseobchoi/CoFailMed}{CoFailMed repository} provides aligned per-case Dice scores for the reported analyses, split identifiers, analysis scripts with reference outputs, a SHA-256 manifest, Croissant metadata and the study code. Its guide maps each table and figure to a command and lists the few values that require masks, predictions or run logs. Values derived from Phikon outputs are withheld under Owkin's licence; datasets, masks, predictions, feature caches and checkpoints are not redistributed (Appendix~\ref{app:broader-impacts}).

\paragraph{Conclusion.} In the expanded 17-bundle pool, shared-checkpoint pairs have the highest mean Dice correlation, followed by distinct checkpoints within a family and then cross-family pairs. On the tasks where they were run, richer readouts and fine-tuning narrow the primary gap, yet fine-tuning does not consistently increase effective pool size at recipe-specific pooled $P_{33}$ thresholds. Nominal model diversity therefore does not imply independent failures, and pool audits should measure co-failure directly on held-out cases.

\clearpage
\begin{ack}
This work was supported by Artificial intelligence industrial convergence cluster development project funded by the Ministry of Science and ICT(MSIT, Korea) \& Gwangju Metropolitan City and the IITP AI Semiconductor Program and the ANCHOR Program, funded by the Korean government and the Seoul Metropolitan Government (IITP-2023-RS-2023-00256081; 2026-ANCHOR-01-110).

\paragraph{Competing interests.} The authors declare no competing interests.
\end{ack}

{\small
\bibliographystyle{plainnat}
\bibliography{references}
}

\clearpage
\appendix
\counterwithin{table}{section}\counterwithin{figure}{section}
\pdfbookmark[1]{Appendix contents}{appendix-contents}
\section*{Appendix contents}

\begingroup
\newcommand{\appendixcontentsentry}[2]{%
  \csname l@#1\endcsname
    {\hyperref[#2]{\numberline{\ref*{#2}}\nameref*{#2}}}%
    {\hyperref[#2]{\pageref*{#2}}}%
}
\appendixcontentsentry{section}{app:notation}
\appendixcontentsentry{section}{app:encoders}
\appendixcontentsentry{subsection}{app:data-notes}
\appendixcontentsentry{subsection}{app:encoder-wrappers}
\appendixcontentsentry{subsection}{app:mask-align}
\appendixcontentsentry{subsection}{app:compute}
\appendixcontentsentry{section}{app:primary-extra}
\appendixcontentsentry{subsection}{app:subject-cluster}
\appendixcontentsentry{subsection}{app:per-rater}
\appendixcontentsentry{subsection}{sec:pixerror-app}
\appendixcontentsentry{subsection}{app:kvasir-official}
\appendixcontentsentry{section}{app:sensitivity}
\appendixcontentsentry{subsection}{app:quality-controlled-pairs}
\appendixcontentsentry{subsection}{app:failure-event}
\appendixcontentsentry{subsection}{app:tau-sweep}
\appendixcontentsentry{subsection}{app:diffleak}
\appendixcontentsentry{section}{app:meff}
\appendixcontentsentry{section}{app:arch-diagnostics}
\appendixcontentsentry{subsection}{app:arch-3task}
\appendixcontentsentry{subsection}{app:cross-checkpoint}
\appendixcontentsentry{section}{app:controls}
\appendixcontentsentry{subsection}{app:unet-skip}
\appendixcontentsentry{subsection}{app:ft}
\appendixcontentsentry{subsection}{app:random-medical-controls}
\appendixcontentsentry{subsection}{app:nnunet}
\appendixcontentsentry{subsection}{app:pipeline-metrics}
\appendixcontentsentry{subsection}{app:amos}
\appendixcontentsentry{section}{app:medsam-note}
\appendixcontentsentry{section}{app:heldout}
\appendixcontentsentry{subsection}{app:heldout-tasks}
\appendixcontentsentry{subsection}{app:pool-selection}
\appendixcontentsentry{section}{app:broader-impacts}
\endgroup

\clearpage
\section{Notation and conventions}
\label{app:notation}

\begin{description}
\item[Tasks] Kvasir (Kvasir-SEG polyp, official split), ACDC LV (left ventricle, official split, slice level), BraTS FG (BraTS 2024 GLI 2D foreground, slice level), RIGA Cup and RIGA Disc (optic cup and disc, Magrabia test source), listed in this order in tables and figures. The held-out tasks are ISIC-2018 and the Hippocampus, Spleen and Heart derivatives of the Medical Segmentation Decathlon (MSD). ``(grouped)'' marks use of the content-group-disjoint Kvasir split in place of the official split.
\item[Bundle] A frozen encoder checkpoint with its wrapper, pretraining, normalisation and feature path (Section~\ref{sec:formal}); the primary pool has four decoder seeds per bundle.
\item[Decoder] The separately trained segmentation head on encoder features, also called the readout or head; the primary decoder is a $1{\times}1$ convolution, a linear probe.
\item[{\boldmath$K$}] The number of bundles retained by the mean-Dice${\geq}0.30$ floor ($4K$ traces in the primary pool). We call a bundle ``functional'' when it passes this floor. ``$K$/avail.'' gives retained / available bundles; ``$K$/members'' gives retained bundles / retained heads.
\item[{\boldmath$\rho_{\mathrm{same}}$, $\rho_{\mathrm{cross}}$}] Mean pairwise Pearson correlation of per-case Dice within the same-bundle/release group (same-bundle seed pairs plus all DINOv2-B$\times$DINOv2-S seed pairs) and within the cross-bundle reference (all remaining pairs); the pixel-overlap table summarises the same strata by error-mask Dice.
\item[{\boldmath$\Delta$}] $\rho_{\mathrm{same}}-\rho_{\mathrm{cross}}$, the primary gap; $\Delta_f$ is the gap at floor $f$, and $\Delta=\Delta_{0.30}$ unless stated. $\Delta_{\mathrm{ckpt}}$ is the same-family/distinct-checkpoint mean minus the cross-family mean (Table~\ref{tab:distinct-ckpt}); $\Delta_{\mathrm{med}}$ is the mean correlation of medical encoders with primary bundles minus $\rho_{\mathrm{cross}}$ (Tables~\ref{tab:medenc} and \ref{tab:medenc-per-encoder}; Table~\ref{tab:medenc}(a) excludes BiomedCLIP from both primary sets); neither is a same/cross gap. $\Delta_{\mathrm{ViT}}$ (Tables~\ref{tab:arch-confound} and~\ref{tab:arch-confound-3task}) is within-bundle seed correlation minus ViT cross-pretraining correlation. $\Delta_{\mathrm{fold}}$ (nnU-Net, Table~\ref{tab:nnunet}) and $\Delta_{\mathrm{sys}}$ (AMOS22, Table~\ref{tab:amos}) are same-fold minus cross-fold and same-system minus cross-system gaps, which estimate different contrasts from $\Delta$.
\item[Intervals] Unless stated otherwise, paired 95\% percentile intervals use $B{=}2000$ bootstrap draws, recomputing both strata on each resample. The resampling unit is the test image for Kvasir and RIGA and the patient (cluster bootstrap) for ACDC LV and BraTS FG; ``slice-level'' or ``case'' resampling uses slices there instead (cases are images for Kvasir and RIGA). ``hier.'' denotes the descriptive hierarchical case/patient$\times$group$\times$trace bootstrap (Appendix~\ref{app:subject-cluster}). Cells show estimates above or beside bracketed intervals; ``--'' means not applicable.
\item[{\boldmath$P_{33}$}] The empirical 33rd percentile of model-case Dice after floor filtering, used as the failure threshold in the primary analyses; pool selection instead uses a calibration-best-model threshold (Appendix~\ref{app:pool-selection}).
\item[{\boldmath$\phi$}] The Pearson correlation of two traces' binary failure indicators.
\item[{\boldmath$\kappa$}] Geirhos error consistency \citep{geirhos2020errorconsistency}, with \KappaNeverFailConvention\ (\href{https://github.com/wichmann-lab/error-consistency/commit/eb53bdb}{\texttt{wichmann-lab/error-consistency}, commit \texttt{eb53bdb}}).
\item[{\boldmath$Q$}, disagreement, double fault] For two traces on $n$ cases, let $n_{11}$ and $n_{00}$ count cases on which both pass and both fail, respectively, and $n_{10},n_{01}$ cases on which exactly one fails. Yule's $Q=(n_{11}n_{00}-n_{10}n_{01})/(n_{11}n_{00}+n_{10}n_{01})$ is undefined when its denominator is zero; the disagreement rate is $(n_{10}+n_{01})/n$ and the double-fault rate is $n_{00}/n$ \citep{kuncheva2003diversity}.
\item[{\boldmath$M_{\mathrm{eff}}$}] $M/[1+(M-1)\rho_{\mathrm{fail}}]$, the equicorrelation effective pool size at the stated threshold, defined for supported pools with a positive, numerically nonsingular denominator.
\item[Supported] A pool in which every member has at least five failures and five passes at the stated threshold. A $\phi$ pair is defined when both members have nonconstant failure indicators; this is a different criterion.
\item[Joint-fail ratio] The ratio of pairwise joint-failure probability in the same-bundle/release group to that in the cross-bundle reference.
\item[Seed correlation] The mean same-bundle cross-seed correlation of a bundle set.
\item[17-bundle pool] The primary pool plus DINOv2-L, CLIP-B/32, ResNet-34/101 and ConvNeXt-S/B. Its cross-family stratum excludes same-family/distinct-checkpoint pairs and therefore differs from the cross-bundle reference.
\item[Shorthand] ``Same/release'' and ``cross'' abbreviate the same-bundle/release group and the cross-bundle reference, so the ``same/cross gap'' is $\Delta$; ``DINOv2-B$\times$S'' denotes the DINOv2-B$\times$DINOv2-S seed pairs.
\end{description}\par Pearson correlations are undefined when either trace has standard deviation below $10^{-8}$; stratum means use finite pairs. Gap intervals omit bootstrap draws without a finite mean in both strata. All five primary gap intervals retain $2000/2000$ draws; the degenerate MSD Spleen gap retains $1987/2000$.

\section{Implementation details}
\label{app:encoders}

\subsection{Datasets, attributions and preprocessing}
\label{app:data-notes}

Both RIGA tasks use the 4-of-6 rater consensus as ground truth: cup is raw label 128 and disc is raw label ${\geq}128$. Rater masks are used unedited. The training set holds the 649 BinRushed and MESSIDOR images; the 95 Magrabia images form the test set. In one of the 570 Magrabia test rater masks (95 images $\times$ six raters), the disc rim (label 255) does not fully enclose the cup; because the targets are defined by label values, this mask needs no repair. Appendix~\ref{app:per-rater} re-scores the same predictions against each individual rater. The RIGA+ training images from MESSIDOR \citep{messidor2014} are used under the Messidor terms of use: ``Kindly provided by the Messidor program partners (see \url{https://www.adcis.net/en/third-party/messidor/}).''

The BraTS FG task is a BraTS 2024 GLI 2D FLAIR derivative with \BratsTestSlices\ test slices from \BratsTestSessions\ sessions of \BratsTestPatients\ patients under a patient-disjoint split (585 training and \BratsTestPatients\ test patients; \BratsTrainPatients\ training patients have eligible slices). Each session contributes up to 10 axial slices with the largest native-grid foreground areas ($\mathrm{seg}>0$), each requiring at least 64 foreground pixels before resizing. The BraTS data-use terms require citing MedPerf \citep{karargyris2023medperf} and the following attribution: ``Data used in this publication were obtained as part of the Brain Tumor Segmentation (BraTS) Challenge project through Synapse ID: syn53708249.'' The portal provides \extlink{https://www.synapse.org/Synapse:syn53708249/wiki/627759}{data access} and \extlink{https://www.synapse.org/Synapse:syn53708249/wiki/627500}{GLI label definitions}.

ACDC LV uses the official challenge split of \citet{acdc}: heads are trained on patients 001--100 and evaluated on the official test patients 101--150, restricted to the LV-positive end-diastolic and end-systolic (ED/ES) slices (\NTestAcdc\ slices of \NAcdcSubjects\ patients). The held-out ISIC-2018 task likewise uses the official Task~1 training/test split \citep{isic2019,ham10000,isic_msk2017} (\HeldoutNTestIsic\ test images); Appendix~\ref{app:heldout} lists the split units of all held-out tasks.

For ACDC and BraTS, NIfTI arrays are read with NiBabel and sliced along their native third axis, without affine reorientation, cropping, padding or intensity windowing. Each selected slice is converted to float32, min--max scaled over all pixels (including background) to $[0,255]$ and truncated to uint8; constant slices become zero. The slice is replicated into three channels, bilinearly resized to $224{\times}224$ with \texttt{align\_corners=False} and no antialiasing, then clamped and truncated to uint8. Encoder input conversion divides by 255 before applying the bundle-specific mean and standard deviation. The release implements these loaders and the feature-extraction path in \path{code/src/fmpool/} (Section~\ref{sec:artifact}).

\subsection{Encoder wrappers and head conventions}
\label{app:encoder-wrappers}

Table~\ref{tab:encoders} lists checkpoints and feature tensors for DINOv2 \citep{oquab2024dinov2}, CLIP \citep{clip2021}, BiomedCLIP \citep{biomedclip2025}, MAE \citep{mae2022}, DeiT \citep{deit2021}, ResNet \citep{resnet2016}, ConvNeXt \citep{convnext2022}, EfficientNet \citep{efficientnet2019}, RETFound \citep{retfound2023}, Phikon \citep{phikon2023}, RAD-DINO \citep{raddino2025} and BiomedParse \citep{biomedparse2025}. The implementations use open\_clip \citep{openclip2021}, timm \citep{timm2019} and torchvision \citep{torchvision2016}, with the shared normalisation and resizing conventions described below.

All bundles use the same $224{\times}224$ inputs, with bilinear image downsampling without antialiasing and nearest-neighbour mask resizing. Official model presets use antialiasing (torchvision) or bicubic PIL resampling (open\_clip, timm, Hugging Face processors), so absolute Dice may differ from the vendor pipelines. The three OpenAI CLIP wrappers read the last transformer block's patch tokens before CLIP's final LayerNorm (\texttt{ln\_post}); all other ViT wrappers read LayerNorm-normalised tokens. Because the $1{\times}1$ head has no BatchNorm, CLIP features enter it under a different scaling convention.

The $1{\times}1$ head is a simplified DINOv2-style linear segmentation probe (no BatchNorm, no augmentation, $224{\times}224$ inputs, Adam with a constant learning rate). Linear-probe rankings can differ from fine-tuning rankings \citep{kerssies2024benchmark}. The convolution produces one-channel logits on the native feature grid. These are bilinearly upsampled to $224{\times}224$ with \texttt{align\_corners=False} before BCE-with-logits training and sigmoid thresholding at evaluation. Features are cached once per bundle and reused by every seed. Head weights are trained separately for each bundle and seed, using a common head architecture and training recipe. The Dice convention of Section~\ref{sec:design} follows \texttt{DiceMetric} in MONAI~1.6.0 \citep{monai_dicemetric}, implemented in \href{https://github.com/Project-MONAI/MONAI/blob/1.6.0/monai/metrics/meandice.py}{\texttt{monai/metrics/meandice.py}}.

RAD-DINO (518\,px) and BiomedParse (1024\,px) have larger native input sizes but, like all bundles, receive $224{\times}224$ inputs. Bundles use ImageNet normalisation except CLIP/BiomedCLIP (OpenAI CLIP normalisation) and RAD-DINO (its processor).

\begin{table}[htbp]
  \caption{Encoder wrappers. Checkpoint identities and decoder inputs. LN denotes final LayerNorm.}
  \label{tab:encoders}
  \centering

    \par\noindent\textbf{(a) Checkpoint sources}\par\smallskip
  \begin{tabularx}{\linewidth}{@{}>{\raggedright\arraybackslash}p{0.20\linewidth}Y@{}}
    \toprule
    \bfseries \textbf{Bundle} & \bfseries \textbf{Checkpoint / identifier} \\
    \midrule
    DINOv2-S/B/L & torch.hub \path{facebookresearch/dinov2}; pinned revision \path{7764ea0f} \\
    \addlinespace[3pt]
    CLIP-B/16, B/32, L/14 & open\_clip \path{ViT-B-16-quickgelu}, \path{ViT-B-32-quickgelu}, \path{ViT-L-14-quickgelu}; \path{openai} weights \\
    \addlinespace[3pt]
    BiomedCLIP & open\_clip hf-hub \path{microsoft/BiomedCLIP-PubMedBERT_256-vit_base_patch16_224} \\
    \addlinespace[3pt]
    MAE-B & timm \path{vit_base_patch16_224.mae} \\
    \addlinespace[3pt]
    DeiT-B & timm \path{deit_base_patch16_224.fb_in1k} \\
    \addlinespace[3pt]
    ResNet, ConvNeXt, EfficientNet & torchvision \path{IMAGENET1K_V1} enums: ResNet-18/34/50/101, ConvNeXt-T/S/B, EfficientNet-B0 \\
    \addlinespace[3pt]
    RETFound & Hugging Face \path{YukunZhou/RETFound_mae_natureCFP} \\
    \addlinespace[3pt]
    Phikon & Hugging Face \path{owkin/phikon} \\
    \addlinespace[3pt]
    RAD-DINO & Hugging Face \path{microsoft/rad-dino} (DINOv2 ViT-B/14) \\
    \addlinespace[3pt]
    BiomedParse v1 & Hugging Face \path{microsoft/BiomedParse}, \path{biomedparse_v1.pt}; backbone keys only \\
    \bottomrule
  \end{tabularx}
  \par\medskip\noindent\textbf{(b) Features passed to the decoder}\par\smallskip
  \begin{tabularx}{\linewidth}{@{}>{\raggedright\arraybackslash}p{0.20\linewidth}Yc@{}}
    \toprule
    \bfseries \textbf{Bundle} & \bfseries \textbf{Feature tensor} & \bfseries \textbf{Grid} \\
    \midrule
    DINOv2-S/B/L & Final-block patch tokens after LN (\path{x_norm_patchtokens}) & $16{\times}16$ \\
    CLIP-B/16 & Last-block patch tokens before \path{ln_post}; no LN & $14{\times}14$ \\
    CLIP-B/32 & Last-block patch tokens before \path{ln_post}; no LN & $7{\times}7$ \\
    CLIP-L/14 & Last-block patch tokens before \path{ln_post}; no LN & $16{\times}16$ \\
    BiomedCLIP & Final-norm patch tokens, LN & $14{\times}14$ \\
    MAE-B & Final-norm patch tokens, LN & $14{\times}14$ \\
    DeiT-B & Final-norm patch tokens, LN; CLS dropped & $14{\times}14$ \\
    ResNet & Last stage, \path{layer4} & $7{\times}7$ \\
    ConvNeXt & \path{features}, without classifier LayerNorm & $7{\times}7$ \\
    EfficientNet & \path{features}, including head convolution & $7{\times}7$ \\
    RETFound & Final encoder norm, CLS dropped, LN & $14{\times}14$ \\
    Phikon & Final-norm patch tokens, LN with official $\epsilon{=}10^{-6}$ & $14{\times}14$ \\
    RAD-DINO & Final-block patch tokens, LN & $16{\times}16$ \\
    BiomedParse v1 & FocalNet \path{res5}, after stage LayerNorm & $7{\times}7$ \\
    \bottomrule
  \end{tabularx}
\end{table}

\subsection{Mask resize convention}
\label{app:mask-align}

Table~\ref{tab:mask-align} compares default and exact mask alignment on four tasks. The gaps change by at most \MaskAlignMaxAbsDelta, $K$ is unchanged, and all same/cross gap intervals exclude zero.

Images use bilinear resizing (\texttt{align\_corners=False}); masks use PyTorch's default \texttt{nearest} mode, which floors the scaled destination index rather than aligning pixel centres and shifts foreground centroids by about \MaskAlignOffsetPx\ output pixels relative to \texttt{nearest-exact} (task range \MaskAlignOffsetPxRange; absolute mean shift per axis, averaged over masks with foreground under both resizes, pooling training and test masks for Kvasir and RIGA and the 1808 official training slices for ACDC). The sensitivity retrains all $1{\times}1$ heads for RIGA Cup, RIGA Disc, ACDC LV (official 100/50-patient split) and official-split Kvasir with \texttt{nearest-exact} masks under the same protocol; BraTS is not evaluated. All bundles, seeds and partitions share the resizing rule, cached features are independent of the masks, and results use the default \texttt{nearest} convention unless labelled \texttt{nearest-exact}.

\begin{table}[htbp]
  \caption{Mask resize sensitivity. $\Delta$ under default \texttt{nearest} and \texttt{nearest-exact} mask resizing, with their paired difference (exact minus nearest) computed on shared bootstrap draws.}
  \label{tab:mask-align}
  \centering
  \setlength{\tabcolsep}{5pt}

  \begin{tabular}{@{}TcEEE@{}}
    \toprule
    \bfseries Task & \bfseries \makecell[c]{$K$\\(\texttt{nearest}/exact)} & \multicolumn{1}{c}{\bfseries \makecell[c]{$\Delta$ [95\% CI]\\\texttt{nearest}}} & \multicolumn{1}{c}{\bfseries \makecell[c]{$\Delta$ [95\% CI]\\\texttt{nearest-exact}}} & \multicolumn{1}{c}{\bfseries \makecell[c]{Difference [95\% CI]\\(exact $-$ nearest)}} \\
    \midrule
    Kvasir & 11/11 & 0.265 [0.217, 0.317] & 0.267 [0.219, 0.318] & 0.0018 [0.0003, 0.0033] \\
\addlinespace[1pt]
ACDC LV & 10/10 & 0.317 [0.289, 0.344] & 0.322 [0.292, 0.349] & 0.0043 [0.0019, 0.0068] \\
\addlinespace[1pt]
RIGA Cup & 11/11 & 0.557 [0.500, 0.609] & 0.554 [0.498, 0.606] & \mn0.0029 [\mn0.0075, 0.0021] \\
\addlinespace[1pt]
RIGA Disc & 11/11 & 0.640 [0.570, 0.689] & 0.639 [0.569, 0.689] & \mn0.0008 [\mn0.0048, 0.0038] \\
    \bottomrule
  \end{tabular}
\end{table}

\subsection{Compute}
\label{app:compute}

The released scalar analyses can be recomputed on CPU; the documented reproduction commands complete in about seven minutes in total on a desktop CPU. Feature extraction and decoder training ran on \GpuCount\ \GpuModel\ GPUs of a shared academic cluster. Table~\ref{tab:compute} lists the recorded GPU occupancy by experiment family, measured as the wall-clock union of run intervals per GPU (elapsed run times where per-interval timing was not recorded).

\begin{table}[htbp]
  \caption{Recorded GPU occupancy (GPU-h) by experiment family. The official-split Kvasir family covers the primary, distinct-checkpoint and medical-encoder pools; the official-split ACDC family covers the primary, distinct-checkpoint, UNet-skip and random-initialisation pools. Family totals also include the Kvasir content-group-split sensitivity runs and unreported runs on an earlier internal ACDC split that the official-split runs replaced, including 10 nnU-Net training runs. Feature extraction is included; nnU-Net preprocessing and inference, native-3D AMOS inference and other preliminary validation runs are excluded. The sum is computed before rounding; distinct occupancy counts overlapping intervals once because inference runs co-ran with training.}
  \label{tab:compute}
  \centering
  \begin{tabular*}{\linewidth}{@{\extracolsep{\fill}}lrlr@{}}
    \toprule
    \bfseries Experiment family & \bfseries GPU-h & \bfseries Experiment family & \bfseries GPU-h \\
    \midrule
    Primary $1{\times}1$ grid & \GpuHoursPrimary & Per-rater inference & \GpuHoursPerRater \\
    Distinct-checkpoint tier & \GpuHoursXckpt & Pixel-overlap inference & \GpuHoursPixelReplay \\
    Medical encoders & \GpuHoursMedEnc & Held-out tasks & \GpuHoursHeldout \\
    UNet-skip readouts & \GpuHoursUnetSkip & Mask-alignment sensitivity & \GpuHoursMaskAlign \\
    End-to-end fine-tuning & \GpuHoursFt & Official-split Kvasir & \GpuHoursKvasirOfficial \\
    nnU-Net training & \GpuHoursNnunet & Official-split ACDC and ISIC-2018 & \GpuHoursOfficialSplits \\
    Random-initialisation controls & \GpuHoursRandInit & & \\
    \midrule
    Sum over families & \GpuHoursTotal & Distinct occupancy & \GpuHoursDistinct \\
    \bottomrule
  \end{tabular*}
\end{table}

For the official splits, recorded times per frozen-feature encoder cell (feature extraction and four trained heads) were 2.4--10.7 min on Kvasir, 3.9--9.1 min on ACDC LV and 15.0--32.8 min on ISIC-2018 (11 cells per task; cells co-ran on shared GPUs). The 100-epoch nnU-Net training runs took 33.16--34.98 min each on RIGA Cup and 53.73--56.08 min each on official ACDC (10 runs per task). These shared-cluster elapsed times are not inference latency measurements.

\FloatBarrier
\section{Additional results on the primary tasks}
\label{app:primary-extra}

Table~\ref{tab:primary-quality} reports mean Dice for all 11 primary bundles before mean-Dice filtering. Each value averages per-case Dice across all four decoder seeds under the primary protocol and gives the quality basis for pool selection.

\begin{table}[htbp]
  \caption{Primary-bundle segmentation quality before the mean-Dice floor. Mean Dice over test cases and four decoder seeds; ACDC LV and BraTS FG are slice-level means. All 11 bundles are shown, including those below the 0.30 floor. ResNet-18 falls below the floor on ACDC LV and BraTS FG.}
  \label{tab:primary-quality}
  \centering
  \begin{tabular*}{\linewidth}{@{\extracolsep{\fill}}lccccc@{}}
    \toprule
    \bfseries Bundle & \bfseries Kvasir & \bfseries ACDC LV & \bfseries BraTS FG & \bfseries RIGA Cup & \bfseries RIGA Disc \\
    \midrule
    DINOv2-B & 0.729 & 0.705 & 0.736 & 0.698 & 0.920 \\
\addlinespace[1pt]
DINOv2-S & 0.729 & 0.686 & 0.711 & 0.592 & 0.911 \\
\addlinespace[1pt]
BiomedCLIP & 0.642 & 0.698 & 0.693 & 0.700 & 0.933 \\
\addlinespace[1pt]
CLIP-L/14 & 0.682 & 0.662 & 0.690 & 0.646 & 0.899 \\
\addlinespace[1pt]
CLIP-B/16 & 0.649 & 0.575 & 0.645 & 0.499 & 0.884 \\
\addlinespace[1pt]
MAE-B & 0.653 & 0.672 & 0.730 & 0.609 & 0.869 \\
\addlinespace[1pt]
DeiT-B & 0.681 & 0.663 & 0.686 & 0.713 & 0.905 \\
\addlinespace[1pt]
ConvNeXt-T & 0.568 & 0.450 & 0.489 & 0.459 & 0.783 \\
\addlinespace[1pt]
EfficientNet-B0 & 0.538 & 0.382 & 0.420 & 0.566 & 0.840 \\
\addlinespace[1pt]
ResNet-50 & 0.496 & 0.367 & 0.442 & 0.391 & 0.789 \\
\addlinespace[1pt]
ResNet-18 & 0.356 & 0.263 & 0.296 & 0.304 & 0.729 \\
    \bottomrule
  \end{tabular*}
\end{table}

\subsection{Bootstrap intervals and patient clustering}
\label{sec:bootstrap}
\label{app:subject-cluster}
\label{app:brats-bootstrap-caveat}

The intervals for the patient-clustered, slice-level, hierarchical and patient-aggregated estimates below all exclude zero.

The case interval resamples test units with replacement and recomputes both strata on each draw. For ACDC LV and BraTS FG, the primary interval resamples patients, retaining all slices of each sampled patient. Slice-level intervals are narrower; Table~\ref{tab:main} reports both.

The hierarchical interval resamples cases (patients for ACDC LV and BraTS FG), bundle/release groups and traces within groups, all with replacement. DINOv2-B/S form one eight-trace release group; each other retained bundle forms a separate four-trace group. Same-group pairs come from one sampled occurrence; cross-group pairs have different source groups, and pairs between separate occurrences of the same source group are excluded. Duplicate trace draws within an occurrence retain their self-correlations; excluding them lowers the lower endpoints by at most \VerifHierSelfPairMaxDrop\ (RIGA Cup \VerifHierLoWithSelfCup$\to$\VerifHierLoNoSelfCup, RIGA Disc \VerifHierLoWithSelfDisc$\to$\VerifHierLoNoSelfDisc).

ACDC LV evaluates \NTestAcdc\ LV-positive slices from \NAcdcSubjects\ patients, identified by the released split files. Averaging each head's per-slice Dice within patient before computing Pearson gives a different estimand from the clustered slice-level statistic: ACDC $\Delta{=}\GapAcdcSubject$ \GapAcdcSubjectCI\ (\NAcdcSubjects\ patients) and BraTS $\Delta{=}\GapBratsSubject$ \GapBratsSubjectCI\ (\NBratsSubjects\ patients). Patient aggregation in the distinct-checkpoint tier of Table~\ref{tab:distinct-ckpt} gives ACDC \XckptSubjectGapAcdc\ \XckptSubjectGapAcdcCI\ and BraTS \XckptSubjectGapBrats\ \XckptSubjectGapBratsCI.

\subsection{Per-rater RIGA re-scoring}
\label{app:per-rater}

The correlation gap remains positive against every individual RIGA rater (Table~\ref{tab:per-rater}).

We score the same primary predictions for both RIGA+ \citep{docr2022,rigaplus2022} tasks against each of the six rater masks and the 4-of-6 consensus used in Table~\ref{tab:main}. The GT-specific pool applies the Dice${\geq}0.30$ floor separately for each ground truth, so $K$ can vary by rater. The fixed pool retains the bundles selected using the consensus.

Section~\ref{sec:main} compares the consensus gaps with the single-rater ranges. On Cup, the consensus gap exceeds the GT-specific single-rater range by \PerRaterCupGtConsensusAboveSingleMax; the consensus pool retains ResNet-18, whereas four rater-specific pools exclude it. These comparisons therefore reflect both annotation and pool-composition differences.

\begin{table}[htbp]
  \caption{Per-rater re-scoring of the RIGA heads. GT-specific pool: floor applied per ground truth; fixed pool: bundles retained on the consensus; $K$/members: retained bundles / retained heads. On RIGA Disc every rater-specific floor retains all 11 bundles, so the two pools coincide and one block is shown.}
  \label{tab:per-rater}
  \centering

  \par\noindent\textbf{(a) RIGA Cup}\par\smallskip
  \begin{tabular*}{\linewidth}{@{\extracolsep{\fill}}lc c E c c E@{}}
    \toprule
    & \multicolumn{3}{c}{\bfseries GT-specific pool} & \multicolumn{3}{c}{\bfseries Fixed consensus pool} \\
    \cmidrule(lr){2-4}\cmidrule(lr){5-7}
    \bfseries \textbf{GT} & \bfseries $K$/members & \bfseries \makecell[c]{$\rho_{\mathrm{same}}$\\$\rho_{\mathrm{cross}}$} & \multicolumn{1}{c}{\bfseries $\Delta$ [95\% CI]} & \bfseries $K$/members & \bfseries \makecell[c]{$\rho_{\mathrm{same}}$\\$\rho_{\mathrm{cross}}$} & \multicolumn{1}{c}{\bfseries $\Delta$ [95\% CI]} \\
    \midrule
    Rater 1 & 10/40 & \makecell[t]{0.876\\0.331} & 0.545 [0.487, 0.598] & 11/44 & \makecell[t]{0.876\\0.313} & 0.563 [0.506, 0.614] \\
\addlinespace[1pt]
Rater 2 & 10/40 & \makecell[t]{0.879\\0.370} & 0.509 [0.447, 0.567] & 11/44 & \makecell[t]{0.878\\0.342} & 0.536 [0.481, 0.590] \\
\addlinespace[1pt]
Rater 3 & 10/40 & \makecell[t]{0.882\\0.342} & 0.539 [0.479, 0.593] & 11/44 & \makecell[t]{0.880\\0.330} & 0.550 [0.490, 0.605] \\
\addlinespace[1pt]
Rater 4 & 10/40 & \makecell[t]{0.883\\0.372} & 0.511 [0.451, 0.570] & 11/44 & \makecell[t]{0.882\\0.344} & 0.537 [0.480, 0.592] \\
\addlinespace[1pt]
Rater 5 & 11/44 & \makecell[t]{0.870\\0.344} & 0.526 [0.472, 0.577] & 11/44 & \makecell[t]{0.870\\0.344} & 0.526 [0.472, 0.577] \\
\addlinespace[1pt]
Rater 6 & 11/44 & \makecell[t]{0.879\\0.364} & 0.516 [0.454, 0.574] & 11/44 & \makecell[t]{0.879\\0.364} & 0.516 [0.454, 0.574] \\
\addlinespace[1pt]
Consensus & 11/44 & \makecell[t]{0.867\\0.310} & 0.557 [0.500, 0.609] & 11/44 & \makecell[t]{0.867\\0.310} & 0.557 [0.500, 0.609] \\
    \bottomrule
  \end{tabular*}

  \par\medskip
  \par\noindent\textbf{(b) RIGA Disc}\par\smallskip
  \begin{tabular*}{\linewidth}{@{\extracolsep{\fill}}lccE@{}}
    \toprule
    & \multicolumn{3}{c}{\bfseries GT-specific pool (identical to the fixed consensus pool)} \\
    \cmidrule(lr){2-4}
    \bfseries \textbf{GT} & \bfseries $K$/members & \bfseries \makecell[c]{$\rho_{\mathrm{same}}$\\$\rho_{\mathrm{cross}}$} & \multicolumn{1}{c}{\bfseries $\Delta$ [95\% CI]} \\
    \midrule
    Rater 1 & 11/44 & \makecell[t]{0.863\\0.216} & 0.647 [0.582, 0.695] \\
\addlinespace[1pt]
Rater 2 & 11/44 & \makecell[t]{0.902\\0.275} & 0.627 [0.526, 0.704] \\
\addlinespace[1pt]
Rater 3 & 11/44 & \makecell[t]{0.902\\0.278} & 0.623 [0.553, 0.685] \\
\addlinespace[1pt]
Rater 4 & 11/44 & \makecell[t]{0.880\\0.230} & 0.651 [0.580, 0.702] \\
\addlinespace[1pt]
Rater 5 & 11/44 & \makecell[t]{0.906\\0.279} & 0.627 [0.553, 0.682] \\
\addlinespace[1pt]
Rater 6 & 11/44 & \makecell[t]{0.888\\0.267} & 0.621 [0.564, 0.665] \\
\addlinespace[1pt]
Consensus & 11/44 & \makecell[t]{0.870\\0.230} & 0.640 [0.570, 0.689] \\
    \bottomrule
  \end{tabular*}

\end{table}

\subsection{Pixel-level error overlap, example cases and per-case scatter}
\label{sec:pixerror-app}
\label{sec:confidence-app}

\begin{figure}[htbp]
  \centering
  \includegraphics[width=\linewidth]{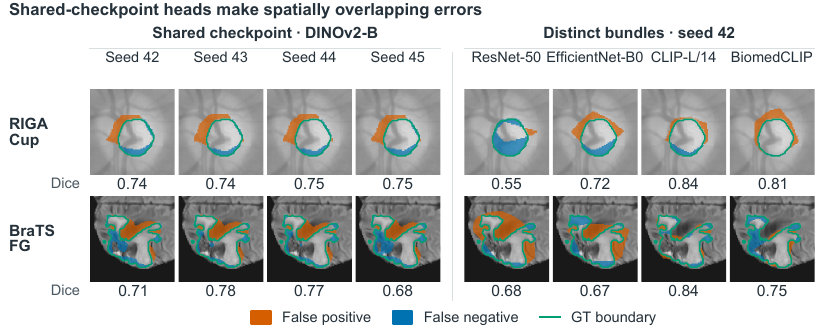}
  \caption{Individual-head errors on two example cases. RIGA Cup C1 and BraTS FG C7 from Figure~\ref{fig:hero}, with false positives, false negatives and ground-truth boundaries; labels give per-head Dice. The left group shares DINOv2-B across decoder seeds 42--45; the right uses ResNet-50, EfficientNet-B0, CLIP-L/14 and BiomedCLIP at seed 42. Cohort-level overlap estimates are given in Table~\ref{tab:pixel-error-overlap}.}
  \label{fig:spatial-details}
\end{figure}

\begin{figure}[htbp]
  \centering
  \includegraphics[width=\linewidth]{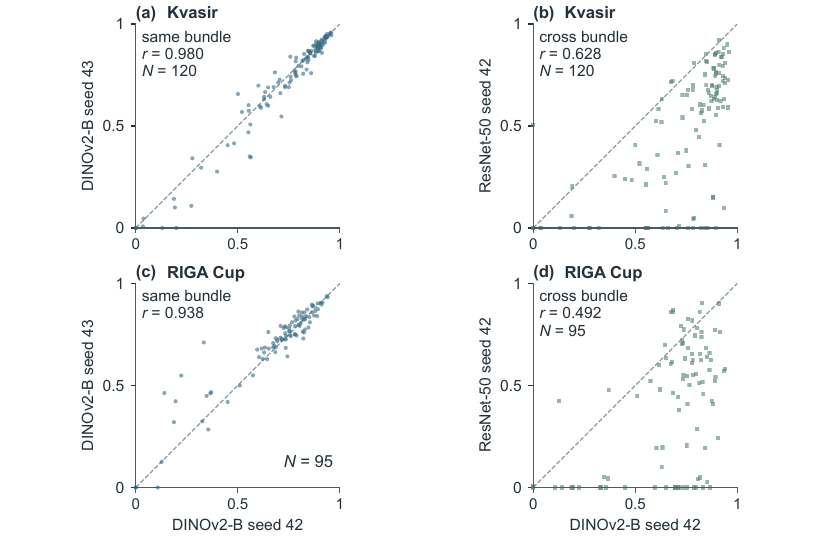}
  \caption{Per-case Dice correlation within and across bundles. DINOv2-B seed 42 is compared with seed 43 of the same bundle or ResNet-50 seed 42 on Kvasir and RIGA Cup. Panels report Pearson $r$ and case count; dashed lines indicate identical Dice.}
  \label{fig:scatter}
\end{figure}

Table~\ref{tab:pixel-error-overlap} reports the error-mask Dice strata and error-type breakdown; the same-bundle/release group has greater overlap for both false positives and false negatives. Figure~\ref{fig:spatial-details} shows individual-head errors on two example cases, and Figure~\ref{fig:scatter} illustrates the stronger per-case Dice correlation of a same-bundle pair relative to a cross-bundle pair.

\paragraph{Cohort overlap.} For primary heads from bundles retained by the floor, error masks are the prediction XOR the ground truth at $224{\times}224$. We compute Dice between each pair of heads' error masks and average over test cases. Strata follow Table~\ref{tab:main}; the DINOv2-B$\times$S column isolates the only distinct-checkpoint pairs in the same-bundle/release group. BraTS FG appears only in the fixed-case examples; the cohort-level overlap covers the other four primary tasks. The patient-resampled ACDC LV gap interval is \PixelGapAcdcClusterCI. Kvasir uses the official-split heads; on the content-group split, the gap is \PixelGapKvasirGroupedSens~\PixelGapKvasirGroupedSensCI. The gap persists for both false-positive-only and false-negative-only masks, so spatial co-failure is not driven by a single error type.

\paragraph{Empty error masks.} When both error masks are empty, the unconditional Dice is one. No observed pair has two empty full error masks (false positives $\cup$ false negatives) on the same case; this can occur for FP-only and FN-only masks. Table~\ref{tab:pixel-empty} reports these counts and recomputes the mean after excluding comparisons with both masks empty. Within each stratum, we average defined pairs per case, then average these case means, omitting cases with no defined comparisons. All conditional gap intervals remain positive; they resample cases (slices for ACDC), not patients.

\paragraph{Example selection.} Figure~\ref{fig:spatial-details} shows, for each task, the test case whose mean Dice over the eight shown heads is closest to the mean of the other seven panel cases (RIGA Cup: MagrabiFemale image 5; BraTS FG: BraTS-GLI-02193-104, slice 112). Crops are centred on the union of ground truth and the eight predictions. Selection is confined to these eight panel candidates (the middle eight test cases with positive ground-truth area, sorted by area and case identifier; C1--C8 follow this stored order and denote different cases across tasks), which are not a random cohort sample. Figure~\ref{fig:hero} displays fixed candidate indices C1, C4 and C7 per task, including those two examples; each case uses the same union-based crop for both head groups.

\paragraph{Error-count maps.} For a displayed four-head group $G$ with binary predictions $p_h$ and ground truth $Y$, Figure~\ref{fig:hero} counts false positives and false negatives at each pixel:
\[
\begin{aligned}
n_{\mathrm{FP}}(x)&=\sum_{h\in G}\mathbf{1}[p_h(x)=1,Y(x)=0],\\
n_{\mathrm{FN}}(x)&=\sum_{h\in G}\mathbf{1}[p_h(x)=0,Y(x)=1].
\end{aligned}
\]
The common discrete colour scales run from one to four; pixels with zero errors retain the greyscale image. These fixed-case counts describe error recurrence; the cohort estimates in Table~\ref{tab:pixel-error-overlap} use pairwise error-mask Dice.

\begin{table}[htbp]
  \caption{Pixel-level error-mask Dice. Panel (a) gives mean overlap per primary stratum and the same-bundle/release-minus-cross-bundle gap. Panel (b) separates pair types and false-positive/false-negative errors. Its final column compares same-bundle pairs with all distinct-bundle pairs, with the DINOv2-B$\times$S pairs in the cross reference. Intervals are case-level (slice-level for ACDC LV, whose patient-clustered gap interval is given in the text).}
  \label{tab:pixel-error-overlap}
  \label{tab:pixel-fpfn}
  \centering
  \setlength{\tabcolsep}{5pt}

  \par\noindent\textbf{(a) Primary strata and gap}\par\smallskip
  \begin{tabular*}{\linewidth}{@{\extracolsep{\fill}}TccEEE@{}}
    \toprule
    \bfseries \textbf{Task} & \bfseries $N$ & \bfseries $K$ & \multicolumn{1}{c}{\bfseries \makecell[c]{Same/release\\group}} & \multicolumn{1}{c}{\bfseries \makecell[c]{Cross-bundle\\reference}} & \multicolumn{1}{c}{\bfseries $\Delta$ [95\% CI]} \\
    \midrule
    Kvasir & 120 & 11 & 0.842 [0.831, 0.854] & 0.499 [0.467, 0.534] & 0.343 [0.319, 0.366] \\
\addlinespace[1pt]
ACDC LV & 977 & 10 & 0.860 [0.856, 0.863] & 0.465 [0.455, 0.476] & 0.394 [0.386, 0.403] \\
\addlinespace[1pt]
RIGA Cup & 95 & 11 & 0.842 [0.830, 0.853] & 0.486 [0.458, 0.514] & 0.355 [0.334, 0.378] \\
\addlinespace[1pt]
RIGA Disc & 95 & 11 & 0.840 [0.830, 0.848] & 0.397 [0.377, 0.417] & 0.443 [0.428, 0.457] \\
    \bottomrule
  \end{tabular*}

  \par\medskip
  \par\noindent\textbf{(b) Pair types and error types}\par\smallskip
  \begin{tabular*}{\linewidth}{@{\extracolsep{\fill}}TEEEEE@{}}
    \toprule
    \bfseries \textbf{Task} & \multicolumn{1}{c}{\bfseries \makecell[c]{Same-bundle\\pairs}} & \multicolumn{1}{c}{\bfseries DINOv2-B$\times$S} & \multicolumn{1}{c}{\bfseries FP-only $\Delta$} & \multicolumn{1}{c}{\bfseries FN-only $\Delta$} & \multicolumn{1}{c}{\bfseries \makecell[c]{Same minus\\distinct bundle\\$\Delta$}} \\
    \midrule
    Kvasir & 0.896 [0.889, 0.904] & 0.619 [0.586, 0.653] & 0.467 [0.450, 0.484] & 0.319 [0.292, 0.345] & 0.395 [0.368, 0.420] \\
\addlinespace[1pt]
ACDC LV & 0.925 [0.924, 0.927] & 0.614 [0.600, 0.627] & 0.503 [0.495, 0.511] & 0.365 [0.356, 0.375] & 0.457 [0.448, 0.466] \\
\addlinespace[1pt]
RIGA Cup & 0.904 [0.898, 0.910] & 0.584 [0.541, 0.625] & 0.447 [0.424, 0.471] & 0.340 [0.317, 0.363] & 0.416 [0.393, 0.440] \\
\addlinespace[1pt]
RIGA Disc & 0.907 [0.903, 0.911] & 0.562 [0.523, 0.596] & 0.514 [0.497, 0.530] & 0.459 [0.438, 0.481] & 0.507 [0.489, 0.525] \\
    \bottomrule
  \end{tabular*}

\end{table}

\begin{table}[htbp]
  \caption{Empty-error-mask sensitivity. Counts of head-pair comparisons on individual cases in which both error masks are empty, and the same/release-minus-cross gap after excluding those comparisons.}
  \label{tab:pixel-empty}
  \centering
  \begin{tabular*}{\linewidth}{@{\extracolsep{\fill}}llccE@{}}
    \toprule
    \bfseries Task & \bfseries Error & \bfseries \makecell{Both-empty/total\\same/release} & \bfseries \makecell{Both-empty/total\\cross-bundle} & \multicolumn{1}{c}{\bfseries \makecell{Conditional $\Delta$\\{}[95\% CI]}} \\
    \midrule
    Kvasir & FP & 905/9840 & 4524/103{,}680 & 0.465 [0.450, 0.480] \\
\addlinespace[1pt]
 & FN & 92/9840 & 160/103{,}680 & 0.317 [0.291, 0.343] \\
\addlinespace[4pt]
ACDC LV & FP & 12{,}356/74{,}252 & 59{,}994/687{,}808 & 0.519 [0.513, 0.526] \\
\addlinespace[1pt]
 & FN & 2110/74{,}252 & 4563/687{,}808 & 0.363 [0.354, 0.372] \\
\addlinespace[4pt]
RIGA Cup & FP & 2194/7790 & 12{,}688/82{,}080 & 0.466 [0.444, 0.489] \\
\addlinespace[1pt]
 & FN & 372/7790 & 1004/82{,}080 & 0.334 [0.313, 0.358] \\
\addlinespace[4pt]
RIGA Disc & FP & 708/7790 & 1807/82{,}080 & 0.504 [0.487, 0.519] \\
\addlinespace[1pt]
 & FN & 424/7790 & 1176/82{,}080 & 0.453 [0.433, 0.473] \\
    \bottomrule
  \end{tabular*}
\end{table}

\subsection{Kvasir official split and content-group sensitivity}
\label{app:kvasir-official}

Table~\ref{tab:kvasir-official} compares the official split with the content-group-disjoint split; both gap intervals exclude zero.

The primary Kvasir task uses the official Kvasir-SEG training/validation split of \citet{jha2021kvasirsplit} (\KvasirOfficialTrainN/\KvasirOfficialTestN), with validation used for testing. All \KvasirOfficialNearDupGroupsWord\ detected near-duplicate groups lie in its training set. The sensitivity uses a second content-group-disjoint split (\KvasirSplitTrainN/\KvasirSplitTestN), which also places every detected near-duplicate group in training; it therefore tests sensitivity to the test partition rather than correcting leakage. The test sets share \KvasirOfficialOverlap\ of \KvasirOfficialTestN\ images and are not fully case-matched. The cross-split bootstrap resamples them separately; it is descriptive because it omits dependence from shared test images. Content groups combine exact image-file hash matches, one audited near-duplicate pair, and pairs with greyscale-image correlation ${\geq}0.995$ and mask Dice ${\geq}0.95$ at $64{\times}64$. Masks are nearest-resized and thresholded at 127; overlapping groups merge transitively. The grouped partition uses SHA-256 ordering of group IDs and deterministic allocation of 120 test images. The release records the audited pair, group memberships and final partitions in \path{code/data/splits/kvasir_grouped_split.json} and \path{kvasir_official_split.json}; \path{code/src/fmpool/datasets/kvasir.py} implements the grouping rule.

ResNet-18 passes the mean-Dice floor on the official split (mean Dice \KvasirOfficialDicePrimaryResnetEighteen). At a 0.40 floor, it is excluded ($K{=}\KKvasirFloorForty$), giving a gap of \GapKvasirFloorForty. The Kvasir pixel-overlap values in Appendix~\ref{sec:pixerror-app} also use the official-split heads.

\begin{table}[htbp]
  \caption{Kvasir on the official split and on the content-group split; cells give estimates with 95\% CIs. The two test sets share \KvasirOfficialOverlap\ of \KvasirOfficialTestN\ images, so the columns are not fully paired. The distinct-checkpoint tier uses the 17-bundle pool; the pooled medical-encoder $\Delta_{\mathrm{med}}$ uses the full primary reference (Table~\ref{tab:medenc-per-encoder}); the content-group UNet-skip pool retains $K{=}\UnetSkipKKvasirGroupedSens$.}
  \label{tab:kvasir-official}
  \centering

  \begin{tabularx}{\linewidth}{@{}Ycc@{}}
    \toprule
    \bfseries Analysis & \bfseries Official split & \bfseries Content-group split \\
    \midrule
    Primary $1{\times}1$: $K$ & \KvasirOfficialK & \KKvasirGroupedSens \\
    \addlinespace[2pt]
    Primary $1{\times}1$: $\rho_{\mathrm{same}}$ / $\rho_{\mathrm{cross}}$ & \KvasirOfficialWithinRbar\ / \KvasirOfficialCrossRbar & \WithinRbarKvasirGroupedSens\ / \CrossRbarKvasirGroupedSens \\
    \addlinespace[2pt]
    Primary $1{\times}1$: $\Delta$ & \KvasirOfficialGap~\KvasirOfficialGapPrimaryCI & \GapKvasirGroupedSens~\GapKvasirGroupedSensCI \\
    \addlinespace[2pt]
    Distinct-checkpoint tier: $\Delta_{\mathrm{ckpt}}$ & \KvasirOfficialXckptGap~\KvasirOfficialXckptGapCI & \XckptGapKvasirGroupedSens~\XckptGapKvasirGroupedSensCI \\
    \addlinespace[2pt]
    Medical encoders: pooled $\Delta_{\mathrm{med}}$ & \KvasirOfficialMedEncPooledDiff~\KvasirOfficialMedEncPooledDiffCI & \MedEncPooledKvasirDiffGroupedSens~\MedEncPooledKvasirDiffGroupedSensCI \\
    \addlinespace[2pt]
    Medical encoders: 15-bundle $\Delta$ & \KvasirOfficialMedEncExtPoolGap~\KvasirOfficialMedEncExtPoolGapCI & \MedEncExtPoolKvasirGapGroupedSens~\MedEncExtPoolKvasirGapGroupedSensCI \\
    \addlinespace[2pt]
    UNet-skip: $\Delta$ & \KvasirOfficialUnetSkipGap~\KvasirOfficialUnetSkipGapCI & \UnetSkipGapKvasirGroupedSens~\UnetSkipGapKvasirGroupedSensCI \\
    \bottomrule
  \end{tabularx}
\end{table}

\FloatBarrier
\section{Sensitivity to metrics and analysis choices}
\label{app:sensitivity}
\label{app:audit-sensitivity}

The following diagnostics use the same traces to examine different mean-Dice floors, correlation statistics, pool definitions, pair-quality adjustments, failure thresholds and item-difficulty nulls (Tables~\ref{tab:main-robustness}--\ref{tab:diffleak}). Some change the metric or comparison population rather than estimating the same $\Delta$. Correlation and chance-corrected agreement gaps remain positive; disagreement-rate gaps are negative (Table~\ref{tab:failure-events}). Magnitudes are metric-specific. All analyses in this appendix are descriptive sensitivity analyses of the evaluated pools: regression coefficients are associational, and permutation tail fractions are not calibrated $p$-values.

\subsection{Floor, correlation statistic, pool definition and pair-quality controls}
\label{app:quality-controlled-pairs}

Table~\ref{tab:main-robustness} reports $\Delta$ with no floor and at floors from 0.20 to 0.40, with ICC(A,1), with each bundle or task left out, and on bundles selected using disjoint calibration data. It also reports the same-bundle/release coefficient $\beta_{\mathrm{same}}$ from a pair-quality regression and the fit's $R^2$; its caption gives the pair and refit counts. Every variant remains positive, and every reported interval excludes zero.

The absolute-agreement, single-measurement intraclass correlation ICC(A,1) \citep{mcgraw1996forming} gives a larger gap than Pearson on every task. The leave-one-bundle-out range covers omission of each retained bundle.

The five-task mean Pearson gap at the 0.30 floor is \GapMetaMean\ (median \GapMetaMedian, range \GapMetaMin--\GapMetaMax). A bundle-label permutation diagnostic, which reapplies the release grouping after each shuffle, compares the observed gap with gaps under randomly reassigned labels (Table~\ref{tab:diffleak}; Appendix~\ref{app:diffleak}).

The calibration split assigns 35\% of each task's aligned units (images for Kvasir and RIGA, patients for ACDC LV and BraTS FG) to calibration, taking units in SHA-256 order of their task-prefixed identifiers; the remaining units form the evaluation part. Bundles are retained when their calibration mean Dice (averaged over slices on ACDC LV and BraTS FG) reaches the 0.30 floor, and $\Delta$ is evaluated on the evaluation part with case-bootstrap intervals. This single fixed split differs from the 200 random half-splits used for pool selection (Appendix~\ref{app:pool-selection}). It is a leakage-sensitivity analysis, not a prospective validation protocol.

The pair-quality regression is fitted by ordinary least squares over all pairs of traces retained in Table~\ref{tab:main} that have a defined correlation:
\begin{equation}
\label{eq:pair-quality}
r_{a,b}=\beta_0+\beta_{\mathrm{same}}I_{\mathrm{same}}(a,b)+\beta_1\bar D_{ab}+\beta_2|\bar D_a-\bar D_b|+\beta_3\bar V_{ab}+\beta_4|V_a-V_b|+\varepsilon_{a,b},
\end{equation}
where $\bar D_a=n^{-1}\sum_iD_a(i)$ and $V_a=n^{-1}\sum_i[D_a(i)-\bar D_a]^2$ are the mean and variance over $n$ cases, and $\bar D_{ab}=(\bar D_a+\bar D_b)/2$ and $\bar V_{ab}=(V_a+V_b)/2$ are their pair averages. The absolute differences control for within-pair quality and variance imbalance. The indicator $I_{\mathrm{same}}$ follows the primary grouping rule (same-bundle pairs plus all DINOv2-B$\times$S seed pairs). The positive coefficient $\beta_{\mathrm{same}}$ indicates that the association persists after these controls. Its interval uses $B{=}1000$ case-bootstrap refits (Table~\ref{tab:main-robustness}(c)).

For ACDC LV and BraTS FG, the calibration-split and $\beta_{\mathrm{same}}$ columns resample slices. The corresponding patient-resampled intervals are \VerifCalibClusterCIAcdc\ and \VerifBetaClusterCIAcdc\ on ACDC LV, and \VerifCalibClusterCIBrats\ and \VerifBetaClusterCIBrats\ on BraTS FG. These are \VerifClusterWidthRatioMin--\VerifClusterWidthRatioMax$\times$ wider than the slice-level intervals; all exclude zero.

\begin{table}[htbp]
  \caption{Primary-gap sensitivity to metrics and pool selection. $\Delta$ at the 0.30 floor with its primary interval, with no floor and at other floors (retained $K$ in parentheses), with ICC(A,1), over leave-one-bundle-out pools (LOO range), and as the mean over the other four tasks when each task is left out. Calibration-selected bundles (calibration/evaluation units and selected count) and the pair-quality coefficient $\beta_{\mathrm{same}}$ appear in panel (c), which uses image/slice-bootstrap intervals, including slice resampling for ACDC and BraTS; $\beta_{\mathrm{same}}$ (Eq.~\eqref{eq:pair-quality}) is fitted over all retained trace pairs of Table~\ref{tab:main} (82 same and 864 cross pairs with 11 bundles; 76 and 704 with 10) with $B{=}1000$ refits, and $R^2$ is that fit's coefficient of determination. Patient-clustered sensitivities are given in the text.}
  \label{tab:main-robustness}
  \label{tab:estimator-variants}
  \label{tab:audit-sensitivity}
  \label{tab:floor-sweep}
  \label{tab:calibration-split}
  \label{tab:quality-controlled-pairs}
  \centering

  \par\noindent\textbf{(a) Mean-Dice floor sensitivity}\par\smallskip
  \begin{tabular*}{\linewidth}{@{\extracolsep{\fill}}TEcPPPP@{}}
    \toprule
    \bfseries \textbf{Task} & \multicolumn{1}{c}{\bfseries \makecell[c]{$\Delta_{0.30}$\\{}[95\% CI]}} & \bfseries \makecell[c]{No floor\\($K{=}11$)} & \multicolumn{1}{c}{\bfseries \makecell[c]{0.20\\($K$)}} & \multicolumn{1}{c}{\bfseries \makecell[c]{0.25\\($K$)}} & \multicolumn{1}{c}{\bfseries \makecell[c]{0.35\\($K$)}} & \multicolumn{1}{c}{\bfseries \makecell[c]{0.40\\($K$)}} \\
    \midrule
    Kvasir & 0.265 [0.217, 0.317] & 0.265 & 0.265 (11) & 0.265 (11) & 0.265 (11) & 0.225 (10) \\
\addlinespace[1pt]
ACDC LV & 0.317 [0.289, 0.344] & 0.351 & 0.351 (11) & 0.351 (11) & 0.317 (10) & 0.227 (8) \\
\addlinespace[1pt]
BraTS FG & 0.340 [0.296, 0.388] & 0.382 & 0.382 (11) & 0.382 (11) & 0.340 (10) & 0.340 (10) \\
\addlinespace[1pt]
RIGA Cup & 0.557 [0.500, 0.609] & 0.557 & 0.557 (11) & 0.557 (11) & 0.541 (10) & 0.531 (9) \\
\addlinespace[1pt]
RIGA Disc & 0.640 [0.570, 0.689] & 0.640 & 0.640 (11) & 0.640 (11) & 0.640 (11) & 0.640 (11) \\
    \bottomrule
  \end{tabular*}

  \par\medskip
  \par\noindent\textbf{(b) Correlation and leave-out analyses}\par\smallskip
  \begin{tabular*}{\linewidth}{@{\extracolsep{\fill}}Tccc@{}}
    \toprule
    \bfseries Task & \bfseries ICC(A,1) $\Delta$ & \bfseries LOO range & \bfseries Leave-task-out mean $\Delta$ \\
    \midrule
    Kvasir & 0.334 & 0.225--0.308 & 0.464 \\
\addlinespace[1pt]
ACDC LV & 0.397 & 0.278--0.361 & 0.451 \\
\addlinespace[1pt]
BraTS FG & 0.435 & 0.314--0.399 & 0.445 \\
\addlinespace[1pt]
RIGA Cup & 0.596 & 0.535--0.653 & 0.391 \\
\addlinespace[1pt]
RIGA Disc & 0.716 & 0.602--0.750 & 0.370 \\
    \bottomrule
  \end{tabular*}

  \par\medskip
  \par\noindent\textbf{(c) Calibration selection and pair-quality adjustment}\par\smallskip
  \begin{tabular*}{\linewidth}{@{\extracolsep{\fill}}TccEEc@{}}
    \toprule
    \bfseries Task & \bfseries \makecell[c]{Cal./eval.\\units} & \bfseries Selected $K$ & \multicolumn{1}{c}{\bfseries \makecell[c]{Cal.-split $\Delta$\\{}[95\% CI]}} & \multicolumn{1}{c}{\bfseries $\beta_{\mathrm{same}}$ [95\% CI]} & \bfseries $R^2$ \\
    \midrule
    Kvasir & 42/78 & 10 & 0.210 [0.161, 0.273] & 0.153 [0.115, 0.201] & 0.836 \\
\addlinespace[1pt]
ACDC LV & 18/32 & 9 & 0.269 [0.243, 0.294] & 0.221 [0.200, 0.241] & 0.917 \\
\addlinespace[1pt]
BraTS FG & 51/95 & 11 & 0.366 [0.352, 0.382] & 0.233 [0.220, 0.247] & 0.901 \\
\addlinespace[1pt]
RIGA Cup & 33/62 & 10 & 0.546 [0.473, 0.616] & 0.544 [0.476, 0.608] & 0.713 \\
\addlinespace[1pt]
RIGA Disc & 33/62 & 11 & 0.630 [0.534, 0.695] & 0.572 [0.493, 0.631] & 0.687 \\
    \bottomrule
  \end{tabular*}
\end{table}

\subsection{Failure thresholds and failure-event statistics}
\label{app:failure-event}

\begin{table}[htbp]
  \caption{Failure-event gaps and effective pool size at $P_{33}$. (a) Same/release minus cross-bundle gaps in $\phi$, $\kappa$, Yule's $Q$, disagreement rate and double-fault rate. (b) Joint-fail ratio (same/release over cross-bundle pairwise joint-failure probability) and equicorrelation $M_{\mathrm{eff}}$ of nominal four-member pools drawn from distinct bundles (mean over supported pools with a defined ratio, with supported/total counts), also at the fixed $\tau{=}0.85$, whose support is lower on Kvasir and RIGA Cup (Table~\ref{tab:meff}(a)). Threshold handling and resampling units are given in the text below.}
  \label{tab:alt-diversity}
  \label{tab:failure-events}
  \centering
  \setlength{\tabcolsep}{5pt}

  \par\noindent\textbf{(a) Same/release minus cross-bundle failure-event gaps}\par\smallskip
  \begin{tabular*}{\linewidth}{@{\extracolsep{\fill}}TEEEEE@{}}
    \toprule
    \bfseries \textbf{Task} & \multicolumn{1}{c}{\bfseries $\phi$} & \multicolumn{1}{c}{\bfseries $\kappa$} & \multicolumn{1}{c}{\bfseries Yule's $Q$} & \multicolumn{1}{c}{\bfseries Disagreement} & \multicolumn{1}{c}{\bfseries Double fault} \\
    \midrule
    Kvasir & 0.290 [0.243, 0.356] & 0.329 [0.273, 0.388] & 0.072 [0.039, 0.113] & \mn0.176 [\mn0.209, \mn0.147] & 0.055 [0.043, 0.067] \\
\addlinespace[1pt]
ACDC LV & 0.405 [0.378, 0.436] & 0.449 [0.428, 0.471] & 0.128 [0.107, 0.151] & \mn0.233 [\mn0.244, \mn0.221] & 0.084 [0.079, 0.088] \\
\addlinespace[1pt]
BraTS FG & 0.425 [0.391, 0.460] & 0.469 [0.458, 0.480] & 0.169 [0.154, 0.183] & \mn0.252 [\mn0.258, \mn0.246] & 0.088 [0.086, 0.091] \\
\addlinespace[1pt]
RIGA Cup & 0.537 [0.469, 0.590] & 0.558 [0.502, 0.613] & 0.438 [0.341, 0.559] & \mn0.282 [\mn0.313, \mn0.252] & 0.117 [0.103, 0.131] \\
\addlinespace[1pt]
RIGA Disc & 0.576 [0.512, 0.652] & 0.582 [0.517, 0.680] & 0.547 [0.291, 0.658] & \mn0.362 [\mn0.388, \mn0.336] & 0.128 [0.120, 0.136] \\
    \bottomrule
  \end{tabular*}

  \par\medskip
  \par\noindent\textbf{(b) Joint failures and effective pool size}\par\smallskip
  \begin{tabular*}{\linewidth}{@{\extracolsep{\fill}}TE Ecc@{}}
    \toprule
    \bfseries \textbf{Task} & \multicolumn{1}{c}{\bfseries \makecell[c]{Joint-fail ratio\\{}[95\% CI]}} & \multicolumn{1}{c}{\bfseries \makecell[c]{$M_{\mathrm{eff}}$, $P_{33}$\\{}[95\% CI]}} & \bfseries \makecell[c]{Supported/total\\pools at $P_{33}$} & \bfseries \makecell[c]{$M_{\mathrm{eff}}$\\$\tau{=}0.85$} \\
    \midrule
    Kvasir & 1.259 [1.190, 1.353] & 1.559 [1.454, 1.704] & 84{,}480/84{,}480 & 1.865 \\
\addlinespace[1pt]
ACDC LV & 1.427 [1.373, 1.491] & 1.645 [1.573, 1.728] & 53{,}760/53{,}760 & 2.360 \\
\addlinespace[1pt]
BraTS FG & 1.498 [1.421, 1.583] & 1.828 [1.714, 1.966] & 53{,}760/53{,}760 & 2.154 \\
\addlinespace[1pt]
RIGA Cup & 1.800 [1.651, 1.944] & 2.420 [2.182, 2.749] & 84{,}480/84{,}480 & 2.309 \\
\addlinespace[1pt]
RIGA Disc & 2.071 [1.968, 2.212] & 2.918 [2.626, 3.308] & 48{,}384/84{,}480 & 3.001 \\
    \bottomrule
  \end{tabular*}

\end{table}

Table~\ref{tab:failure-events} reports the $P_{33}$ gaps in $\phi$, $\kappa$, Yule's $Q$, disagreement and double-fault rate, together with the joint-fail ratio and the effective size of nominal four-member pools. The $\phi$, $\kappa$, $Q$ and double-fault gaps are positive with intervals excluding zero, the disagreement gap is negative, and all reported pool-mean $M_{\mathrm{eff}}$ values lie below four. Table~\ref{tab:failure-event-summary} scores the traces at the empirical 25th, 33rd and 50th percentiles of model-case Dice after floor filtering.

These empirical thresholds define failure relative to each task's score distribution. Both $\phi$ and the joint-fail ratio use $B{=}1000$ draws with the percentile threshold re-estimated in each draw. The $P_{33}$ threshold and the retained bundles are held fixed for $\kappa$, $Q$, disagreement, double fault and $M_{\mathrm{eff}}$. Intervals for $\kappa$, $Q$, disagreement and double fault resample cases on every task; the $\phi$, joint-fail-ratio and $M_{\mathrm{eff}}$ intervals are patient-clustered for ACDC LV and BraTS FG. For ACDC LV and BraTS FG, the $\kappa$ gap resamples slices; its patient-resampled intervals \VerifKappaClusterCIAcdc\ and \VerifKappaClusterCIBrats\ exclude zero.

On RIGA Disc, \VerifNConstantMembersDisc\ of the 44 heads (BiomedCLIP seeds 43 and 45) never fail at $P_{33}$. A pair with exactly one never-failing head has $\kappa=0$ by construction because observed agreement equals chance agreement; these pairs are retained. The single pair in which both heads never fail uses the convention in Appendix~\ref{app:notation}.

The Disc $P_{33}$ $\kappa$ gap is \KappaGapDiscPThirtyThree\ under this convention and remains positive when the both-never-failing pair is excluded. Excluding every pair involving a never-failing head gives \VerifKappaGapDiscExclConstant. Both $\phi$ and $Q$ exclude pairs with a never-failing head. The five-failure/five-pass rule excludes five heads: all four BiomedCLIP heads (1, 0, 1 and 0 failures for seeds 42--45) and DINOv2-B seed 45 (4 failures). Thus, \MeffFourPoolDiscPThirtyThreeNSupported\ of \MeffFourPoolDiscPThirtyThreeNPools\ four-member Disc pools have a defined $M_{\mathrm{eff}}$; its mean is conditional on these supported pools.

\begin{table}[htbp]
  \caption{Binary failure-event diagnostic across thresholds. $\phi$ per stratum, the $\phi$ gap and the joint-fail ratio at each percentile threshold $\tau$; defined/total: pairs with a finite $\phi$; joint-failure rates average all pairs.}
  \label{tab:failure-event-summary}
  \centering
  \setlength{\tabcolsep}{5pt}

  \begin{tabular}{@{}TcccEEc@{}}
    \toprule
    \bfseries \textbf{Task} & \bfseries Percentile & \bfseries $\tau$ & \bfseries \makecell[c]{$\phi_{\mathrm{same}}$\\$\phi_{\mathrm{cross}}$} & \multicolumn{1}{c}{\bfseries $\phi$ gap [95\% CI]} & \multicolumn{1}{c}{\bfseries \makecell[c]{Joint-fail ratio\\{}[95\% CI]}} & \bfseries \makecell[c]{Defined/total\\pairs\\(same, cross)} \\
    \midrule
    Kvasir & $P_{25}$ & 0.461 & \makecell[t]{0.833\\0.526} & 0.307 [0.250, 0.376] & 1.343 [1.246, 1.471] & 82/82, 864/864 \\
\addlinespace[1pt]
 & $P_{33}$ & 0.569 & \makecell[t]{0.816\\0.525} & 0.290 [0.243, 0.356] & 1.259 [1.190, 1.353] & 82/82, 864/864 \\
\addlinespace[1pt]
 & $P_{50}$ & 0.715 & \makecell[t]{0.852\\0.520} & 0.333 [0.283, 0.392] & 1.174 [1.135, 1.223] & 82/82, 864/864 \\
\addlinespace[4pt]
ACDC LV & $P_{25}$ & 0.390 & \makecell[t]{0.881\\0.508} & 0.373 [0.347, 0.399] & 1.465 [1.398, 1.540] & 76/76, 704/704 \\
\addlinespace[1pt]
 & $P_{33}$ & 0.547 & \makecell[t]{0.885\\0.480} & 0.405 [0.378, 0.436] & 1.427 [1.373, 1.491] & 76/76, 704/704 \\
\addlinespace[1pt]
 & $P_{50}$ & 0.715 & \makecell[t]{0.849\\0.397} & 0.452 [0.420, 0.481] & 1.295 [1.264, 1.328] & 76/76, 704/704 \\
\addlinespace[4pt]
BraTS FG & $P_{25}$ & 0.540 & \makecell[t]{0.821\\0.399} & 0.422 [0.384, 0.463] & 1.645 [1.531, 1.790] & 76/76, 704/704 \\
\addlinespace[1pt]
 & $P_{33}$ & 0.616 & \makecell[t]{0.821\\0.396} & 0.425 [0.391, 0.460] & 1.498 [1.421, 1.583] & 76/76, 704/704 \\
\addlinespace[1pt]
 & $P_{50}$ & 0.713 & \makecell[t]{0.826\\0.387} & 0.439 [0.408, 0.467] & 1.286 [1.254, 1.315] & 76/76, 704/704 \\
\addlinespace[4pt]
RIGA Cup & $P_{25}$ & 0.388 & \makecell[t]{0.742\\0.182} & 0.560 [0.505, 0.609] & 2.178 [1.977, 2.386] & 82/82, 864/864 \\
\addlinespace[1pt]
 & $P_{33}$ & 0.521 & \makecell[t]{0.756\\0.219} & 0.537 [0.469, 0.590] & 1.800 [1.651, 1.944] & 82/82, 864/864 \\
\addlinespace[1pt]
 & $P_{50}$ & 0.655 & \makecell[t]{0.700\\0.203} & 0.497 [0.450, 0.546] & 1.371 [1.317, 1.431] & 82/82, 864/864 \\
\addlinespace[4pt]
RIGA Disc & $P_{25}$ & 0.835 & \makecell[t]{0.717\\0.105} & 0.612 [0.532, 0.713] & 2.676 [2.478, 2.874] & 76/82, 704/864 \\
\addlinespace[1pt]
 & $P_{33}$ & 0.858 & \makecell[t]{0.698\\0.123} & 0.576 [0.512, 0.652] & 2.071 [1.968, 2.212] & 77/82, 784/864 \\
\addlinespace[1pt]
 & $P_{50}$ & 0.892 & \makecell[t]{0.724\\0.111} & 0.613 [0.542, 0.662] & 1.497 [1.440, 1.557] & 82/82, 864/864 \\
    \bottomrule
  \end{tabular}
\end{table}

\FloatBarrier
\subsection{Conditional mean-score threshold crossing}
\label{app:tau-sweep}

Table~\ref{tab:tau-sweep} varies the threshold $\tau$. The recovery advantage $A_{\mathrm{rec}}$, the excess rate at which diverse rather than same-bundle pools keep mean member Dice at or above $\tau$ when a member fails, is largest at the lowest examined threshold ($\tau{=}0.70$) on four tasks and is strongly threshold-dependent.

For each pool, we compute $P[\text{pool-mean Dice}\geq\tau\mid\text{any member Dice}<\tau]$, then average across pools. Same-bundle pools contain four seeds of one functional bundle; diverse pools enumerate all four-bundle functional compositions and seed tuples. The difference $A_{\mathrm{rec}}$ is the diverse-pool rate minus the same-bundle rate. At $\tau{=}0.90$, the difference is small and negative on Kvasir, ACDC LV, BraTS FG and RIGA Cup; RIGA Disc retains a positive difference, with its peak at $\tau{=}0.75$ (Table~\ref{tab:tau-sweep}). We condition separately within each pool. Pools with no member failures have undefined conditional rates and are omitted from the pool-mean summary.

\begin{table}[htbp]
  \caption{Conditional mean-score threshold crossing. Same-bundle and diverse rates and their difference $A_{\mathrm{rec}}$ (diverse minus same-bundle) at each $\tau$; a member-score diagnostic, not the Dice of an aggregated ensemble prediction.}
  \label{tab:tau-sweep}
  \centering

  \begin{tabular*}{\linewidth}{@{\extracolsep{\fill}}Tlccccc@{}}
    \toprule
    & & \multicolumn{5}{c}{\bfseries $\tau$} \\
    \cmidrule(lr){3-7}
    \bfseries Task & \bfseries Rate & \bfseries 0.70 & \bfseries 0.75 & \bfseries 0.80 & \bfseries 0.85 & \bfseries 0.90 \\
    \midrule
    Kvasir & Same-bundle & 0.075 & 0.081 & 0.067 & 0.043 & 0.026 \\
\addlinespace[1pt]
 & Diverse & 0.278 & 0.242 & 0.182 & 0.100 & 0.023 \\
\addlinespace[1pt]
 & $A_{\mathrm{rec}}$ & 0.203 & 0.162 & 0.114 & 0.057 & \mn0.003 \\
\addlinespace[4pt]
ACDC LV & Same-bundle & 0.068 & 0.061 & 0.056 & 0.039 & 0.019 \\
\addlinespace[1pt]
 & Diverse & 0.310 & 0.241 & 0.157 & 0.060 & 0.006 \\
\addlinespace[1pt]
 & $A_{\mathrm{rec}}$ & 0.242 & 0.180 & 0.101 & 0.021 & \mn0.014 \\
\addlinespace[4pt]
BraTS FG & Same-bundle & 0.087 & 0.069 & 0.042 & 0.020 & 0.003 \\
\addlinespace[1pt]
 & Diverse & 0.289 & 0.192 & 0.106 & 0.034 & 0.001 \\
\addlinespace[1pt]
 & $A_{\mathrm{rec}}$ & 0.203 & 0.123 & 0.065 & 0.014 & \mn0.002 \\
\addlinespace[4pt]
RIGA Cup & Same-bundle & 0.087 & 0.083 & 0.054 & 0.028 & 0.005 \\
\addlinespace[1pt]
 & Diverse & 0.223 & 0.132 & 0.056 & 0.015 & 0.001 \\
\addlinespace[1pt]
 & $A_{\mathrm{rec}}$ & 0.136 & 0.049 & 0.002 & \mn0.013 & \mn0.004 \\
\addlinespace[4pt]
RIGA Disc & Same-bundle & 0.426 & 0.162 & 0.196 & 0.264 & 0.110 \\
\addlinespace[1pt]
 & Diverse & 0.945 & 0.886 & 0.781 & 0.565 & 0.225 \\
\addlinespace[1pt]
 & $A_{\mathrm{rec}}$ & 0.519 & 0.725 & 0.585 & 0.302 & 0.114 \\
    \bottomrule
  \end{tabular*}
\end{table}

\subsection{Item-difficulty sensitivity with a common reference panel}
\label{app:diffleak}

For each four-family reference subset $R$ of the eight upstream families, define the common difficulty regressor
\[
  d_R(i)=\frac{1}{|R|}\sum_{g\in R}\frac{1}{n_g}\sum_{a\in g}D_a(i),
\]
where $n_g$ counts the retained traces in family $g$. Thus traces are averaged within each family before the family means receive equal weight. The complementary families form the evaluation panel on the same cases. For each evaluation trace, fit $D_a(i)=\alpha_a+\gamma_a d_R(i)+\epsilon_a(i)$ by OLS with an intercept and compute the gap on the residuals. The regressor is common to all evaluation traces within a panel; the gaps from all $\binom{8}{4}$ reference subsets receive equal weight. Each paired case/patient bootstrap draw reconstructs the family means and common regressor and refits every OLS model. The retained bundles and panel definitions remain fixed.

The residual gap exceeds the raw gap on every task and is positive on every eligible panel. The matched raw gap is the equally weighted mean over the \ItemPanelsAcdc\ complementary evaluation panels (e.g.\ ACDC LV \ItemRawGapAcdc\ vs.\ \GapAcdcPrimary\ in Table~\ref{tab:main}). This panel weighting differs from the weighting of the full-pool gap in Table~\ref{tab:main}.

Table~\ref{tab:diffleak} gives the raw Pearson gap's bundle-label permutation null, with tail fraction $f_{\mathrm{perm}}$. The raw-gap null shuffles the multiset of bundle labels across individual complete traces, preserving label counts but breaking the original checkpoint/seed clusters, then reapplies the same-bundle/release grouping. Its upper-tail fraction is $(1+\#\{\Delta_{\mathrm{perm}}\geq\Delta_{\mathrm{obs}}\})/(1+n_{\mathrm{perm}})$, where $n_{\mathrm{perm}}$ counts valid permutations; because bundle labels are not exchangeable sampling units, this fraction is descriptive. This is not a null for the residualised gap. A separate residualised-gap null (\VerifResidNullNPerm\ bundle-label permutations, with the family reference and evaluation panels rebuilt from the permuted labels) has task means between $-0.004$ and $-0.002$ and a maximum of 0.169, below every observed residualised gap.

\begin{table}[htbp]
  \caption{Common-panel item-difficulty sensitivity. Raw and residualised $\Delta$, their difference and residual-to-raw ratio; eligible/total reference panels (positive residual-gap count in parentheses); and the range and SD of residual gaps across panels. The raw-gap permutation null gives 2.5th/97.5th percentiles; over 1000 permutations, every task has tail fraction $f_{\mathrm{perm}}{=}0.001$, which is not a calibrated $p$-value. For slice-level tasks, we also report the residual-minus-raw gap after patient aggregation.}
  \label{tab:diffleak}
  \label{tab:diffleak-null}
  \centering
  \setlength{\tabcolsep}{5pt}

  \par\noindent\textbf{(a) Raw and residualised gaps}\par\smallskip
  \begin{tabular*}{\linewidth}{@{\extracolsep{\fill}}TEEEE@{}}
    \toprule
    \bfseries \textbf{Task} & \multicolumn{1}{c}{\bfseries \makecell[c]{Raw $\Delta$\\{}[95\% CI]}} & \multicolumn{1}{c}{\bfseries \makecell[c]{Residual $\Delta$\\{}[95\% CI]}} & \multicolumn{1}{c}{\bfseries \makecell[c]{Residual $-$ raw\\{}[95\% CI]}} & \multicolumn{1}{c}{\bfseries \makecell[c]{Residual/raw\\{}[95\% CI]}} \\
    \midrule
    Kvasir & 0.266 [0.219, 0.317] & 0.683 [0.645, 0.715] & 0.417 [0.364, 0.467] & 2.568 [2.163, 3.078] \\
\addlinespace[1pt]
ACDC LV & 0.328 [0.298, 0.356] & 0.698 [0.682, 0.713] & 0.370 [0.339, 0.403] & 2.127 [1.956, 2.341] \\
\addlinespace[1pt]
BraTS FG & 0.351 [0.305, 0.399] & 0.686 [0.671, 0.700] & 0.335 [0.292, 0.375] & 1.955 [1.738, 2.217] \\
\addlinespace[1pt]
RIGA Cup & 0.570 [0.514, 0.624] & 0.725 [0.696, 0.751] & 0.155 [0.107, 0.205] & 1.271 [1.175, 1.395] \\
\addlinespace[1pt]
RIGA Disc & 0.655 [0.592, 0.701] & 0.725 [0.672, 0.761] & 0.071 [0.041, 0.106] & 1.108 [1.061, 1.172] \\
    \bottomrule
  \end{tabular*}

  \par\medskip
  \par\noindent\textbf{(b) Panel coverage, variation and permutation diagnostic}\par\smallskip
  \begin{tabular*}{\linewidth}{@{\extracolsep{\fill}}Tccccc@{}}
    \toprule
    \bfseries \textbf{Task} & \bfseries \makecell[c]{Eligible/total\\(positive)} & \bfseries \makecell[c]{Panel\\range} & \bfseries \makecell[c]{Panel\\SD} & \multicolumn{1}{c}{\bfseries \makecell[c]{Permutation null\\{}[2.5th, 97.5th]}} & \bfseries \makecell[c]{Patient-level\\residual $-$ raw} \\
    \midrule
    Kvasir & 70/70 (70) & 0.333--0.864 & 0.129 & [\mn0.025, 0.031] & -- \\
\addlinespace[1pt]
ACDC LV & 70/70 (70) & 0.299--0.895 & 0.140 & [\mn0.030, 0.040] & 0.235 \\
\addlinespace[1pt]
BraTS FG & 70/70 (70) & 0.327--0.897 & 0.134 & [\mn0.029, 0.039] & 0.484 \\
\addlinespace[1pt]
RIGA Cup & 70/70 (70) & 0.518--0.921 & 0.111 & [\mn0.040, 0.048] & -- \\
\addlinespace[1pt]
RIGA Disc & 70/70 (70) & 0.529--0.940 & 0.115 & [\mn0.048, 0.055] & -- \\
    \bottomrule
  \end{tabular*}

\end{table}

\FloatBarrier
\section{Effective pool size}
\label{app:meff}

Table~\ref{tab:meff} gives the equicorrelation effective size of nominal four-member pools formed from distinct bundles, with one seed per bundle. At $P_{33}$, task-level means range from \MeffFourPoolKvasirPThirtyThree\ (Kvasir) to \MeffFourPoolDiscPThirtyThree\ (RIGA Disc).

Consider a pool of $M$ members whose per-case failure indicators $Z_1,\dots,Z_M$ at threshold $\tau$ share a common marginal variance $\sigma^2_Z$ and have mean pairwise Pearson correlation $\rho_{\mathrm{fail}}$. The variance of the per-case mean indicator is then $\sigma^2_Z[1+(M-1)\rho_{\mathrm{fail}}]/M$. Equating this with the variance $\sigma^2_Z/M_{\mathrm{eff}}$ of a hypothetical independent pool gives
\begin{equation}
\label{eq:effective-pool-size}
  M_{\mathrm{eff}} = \frac{M}{1 + (M-1)\,\rho_{\mathrm{fail}}}.
\end{equation}
Equation~\eqref{eq:effective-pool-size} is therefore exact under equal marginal variances and does not require equal pairwise correlations. With unequal variances, it summarises the variance of the mean after standardising each failure indicator. It is a threshold-specific variance summary. A nonpositive or numerically singular denominator makes this statistic undefined.

For the empirical covariance matrix $S$ of a pool's binary failure indicators, the exact-covariance variant is
\begin{equation}
\label{eq:effective-pool-covariance}
  M_{\mathrm{eff,cov}} = \frac{M\,\operatorname{tr}(S)}{\mathbf{1}^{\mathsf T}S\mathbf{1}}.
\end{equation}
Equation~\eqref{eq:effective-pool-covariance} compares the variance of the pool mean with that of independent members having the observed average marginal variance. Table~\ref{tab:meff} averages this ratio over supported pools at $\tau{=}0.85$ and $P_{33}$, applying the same five-failure/five-pass rule and excluding zero or numerically singular mean variance. At $P_{33}$, both variants are defined on the same supported pools. The covariance normalisation cancels in the ratio.

Table~\ref{tab:meff} uses binary failure indicators. Unsupported pools are counted rather than imputed; each mean includes only supported pools with a defined ratio. At $\tau{=}0.85$, task-level mean effective sizes range from \MeffFourPoolKvasirFixedEightyFive\ (Kvasir) to \MeffFourPoolDiscFixedEightyFive\ (RIGA Disc). Many RIGA Cup compositions fail the support rule (\MeffFourPoolCupFixedEightyFiveNSupported\ of \MeffFourPoolCupFixedEightyFiveNPools\ supported), so $P_{33}$ is the primary threshold.

\begin{table}[htbp]
  \caption{Effective size of nominal four-member diverse pools. $M_{\mathrm{eff}}$ at the fixed $\tau{=}0.85$ and empirical $P_{33}$ thresholds (mean over supported pools with a defined ratio, with a 95\% bootstrap CI at $P_{33}$), the exact-covariance variant, supported pools over enumerated bundle-and-seed compositions, and unsupported pools; exact-covariance values are point estimates at both thresholds. The $P_{33}$ CI fixes the observed threshold and retained bundles, while rechecking support and ratio eligibility on each draw.}
  \label{tab:meff}
  \centering

  \par\noindent\textbf{(a) Fixed failure threshold $\tau{=}0.85$}\par\smallskip
  \begin{tabular*}{\linewidth}{@{\extracolsep{\fill}}Tcccc@{}}
    \toprule
    \bfseries \textbf{Task} & \bfseries $M_{\mathrm{eff}}$ & \bfseries \makecell[c]{Exact\\covariance} & \bfseries \makecell[c]{Supported/total\\pools} & \bfseries \textbf{Unsupported} \\
    \midrule
    Kvasir & 1.865 & 1.864 & 53{,}760/84{,}480 & 30{,}720 \\
\addlinespace[1pt]
ACDC LV & 2.360 & 2.313 & 53{,}760/53{,}760 & 0 \\
\addlinespace[1pt]
BraTS FG & 2.154 & 2.178 & 53{,}760/53{,}760 & 0 \\
\addlinespace[1pt]
RIGA Cup & 2.309 & 2.317 & 13{,}440/84{,}480 & 71{,}040 \\
\addlinespace[1pt]
RIGA Disc & 3.001 & 3.055 & 48{,}384/84{,}480 & 36{,}096 \\
    \bottomrule
  \end{tabular*}

  \par\medskip
  \par\noindent\textbf{(b) Empirical failure threshold $P_{33}$}\par\smallskip
  \begin{tabular*}{\linewidth}{@{\extracolsep{\fill}}TEccc@{}}
    \toprule
    \bfseries \textbf{Task} & \multicolumn{1}{c}{\bfseries $M_{\mathrm{eff}}$ [95\% CI]} & \bfseries \makecell[c]{Exact\\covariance} & \bfseries \makecell[c]{Supported/total\\pools} & \bfseries \textbf{Unsupported} \\
    \midrule
    Kvasir & 1.559 [1.454, 1.704] & 1.572 & 84{,}480/84{,}480 & 0 \\
\addlinespace[1pt]
ACDC LV & 1.645 [1.573, 1.728] & 1.672 & 53{,}760/53{,}760 & 0 \\
\addlinespace[1pt]
BraTS FG & 1.828 [1.714, 1.966] & 1.857 & 53{,}760/53{,}760 & 0 \\
\addlinespace[1pt]
RIGA Cup & 2.420 [2.182, 2.749] & 2.460 & 84{,}480/84{,}480 & 0 \\
\addlinespace[1pt]
RIGA Disc & 2.918 [2.626, 3.308] & 2.956 & 48{,}384/84{,}480 & 36{,}096 \\
    \bottomrule
  \end{tabular*}

\end{table}

\FloatBarrier
\section{Architecture and checkpoint diagnostics}
\label{app:arch-diagnostics}

\subsection{Shared ViT-family diagnostic across pretraining}
\label{app:arch-3task}

Table~\ref{tab:arch-confound-3task} adds patient-clustered intervals and pair counts to Table~\ref{tab:arch-confound}. On RIGA, cross-pretraining correlation is close to the cross-bundle reference. On ACDC LV, Kvasir and BraTS, it lies between that reference and within-bundle seed correlation.

All five ViT-B-class bundles (DINOv2-B, BiomedCLIP, MAE-B, DeiT-B, CLIP-B/16) pass the mean-Dice floor on every task. They share a broad architecture class but differ in pretraining, feature extraction and normalisation. Within-bundle seed correlation is computed within this five-bundle subset. We report $\Delta_{\mathrm{ViT}}$ with the case interval and, for slice-level tasks, the patient-clustered interval.

\begin{table}[htbp]
  \caption{Shared ViT-family diagnostic across pretraining. Seed correlation and ViT cross-pretraining correlation with case-bootstrap intervals. $\Delta_{\mathrm{ViT}}$ is within-bundle seed correlation minus cross-pretraining correlation, with a case interval and, for slice-level tasks, a patient-clustered interval below it. Pair counts are seed pairs / cross-pretraining pairs.}
  \label{tab:arch-confound-3task}
  \centering

  \begin{tabular}{lcccccc}
    \toprule
    \bfseries Task & \bfseries \makecell[c]{$K$/\\avail.} & \bfseries Seed correlation & \bfseries \makecell[c]{ViT cross-\\pretrain} & \bfseries $\rho_{\mathrm{cross}}$ & \bfseries $\Delta_{\mathrm{ViT}}$ [95\% CI] & \bfseries Pairs \\
    \midrule
    Kvasir & \VitFunctionalKvasir & \makecell[c]{\VitSeedFloorKvasir\\\VitSeedFloorKvasirCI} & \makecell[c]{\VitCrossPretrainKvasir\\\VitCrossPretrainKvasirCI} & \VitCrossRefKvasir & \makecell[c]{\VitDiffKvasir\\\VitDiffKvasirCaseCI} & \VitSeedPairsKvasir\,/\,\VitCrossPretrainPairsKvasir \\
    \addlinespace[3pt]
    ACDC LV & \VitFunctionalAcdc & \makecell[c]{\VitSeedFloorAcdc\\\VitSeedFloorAcdcCI} & \makecell[c]{\VitCrossPretrainAcdc\\\VitCrossPretrainAcdcCI} & \VitCrossRefAcdc & \makecell[c]{\VitDiffAcdc\\\VitDiffAcdcCaseCI\\\VitDiffAcdcClusterCI} & \VitSeedPairsAcdc\,/\,\VitCrossPretrainPairsAcdc \\
    \addlinespace[3pt]
    BraTS FG & \VitFunctionalBrats & \makecell[c]{\VitSeedFloorBrats\\\VitSeedFloorBratsCI} & \makecell[c]{\VitCrossPretrainBrats\\\VitCrossPretrainBratsCI} & \VitCrossRefBrats & \makecell[c]{\VitDiffBrats\\\VitDiffBratsCaseCI\\\VitDiffBratsClusterCI} & \VitSeedPairsBrats\,/\,\VitCrossPretrainPairsBrats \\
    \addlinespace[3pt]
    RIGA Cup & \VitFunctionalCup & \makecell[c]{\VitSeedFloorCup\\\VitSeedFloorCupCI} & \makecell[c]{\VitCrossPretrainCup\\\VitCrossPretrainCupCI} & \VitCrossRefCup & \makecell[c]{\VitDiffCup\\\VitDiffCupCaseCI} & \VitSeedPairsCup\,/\,\VitCrossPretrainPairsCup \\
    \addlinespace[3pt]
    RIGA Disc & \VitFunctionalDisc & \makecell[c]{\VitSeedFloorDisc\\\VitSeedFloorDiscCI} & \makecell[c]{\VitCrossPretrainDisc\\\VitCrossPretrainDiscCI} & \VitCrossRefDisc & \makecell[c]{\VitDiffDisc\\\VitDiffDiscCaseCI} & \VitSeedPairsDisc\,/\,\VitCrossPretrainPairsDisc \\
    
    \bottomrule
  \end{tabular}

\end{table}

\subsection{Distinct-checkpoint tier and primary-pool decomposition}
\label{app:cross-checkpoint}

Table~\ref{tab:distinct-ckpt} separates same-bundle seed pairs, same-family/distinct-checkpoint pairs and cross-family pairs. In the 17-bundle expanded pool, mean per-case Dice correlation is highest for same-bundle seed pairs, followed by distinct checkpoints within a family and then cross-family pairs.

The same-bundle/release group is dominated by same-bundle, different-seed pairs. The 17-bundle pool groups distinct checkpoints by broad upstream family (DINOv2-S/B/L, OpenAI CLIP B/16, B/32 and L/14, ConvNeXt-T/S/B and ResNet-18/34/50/101 where functional). BiomedCLIP remains separate from OpenAI CLIP because its pretraining corpus, text encoder and training configuration differ. The cross-family stratum differs from the cross-bundle reference as stated in Appendix~\ref{app:notation}. Table~\ref{tab:distinct-ckpt}(b) reports the corresponding statistics within the 11-bundle primary pool.

\begin{table}[htbp]
  \caption{Distinct-checkpoint tier and grouping sensitivity. (a) 17-bundle pool after the floor: mean pairwise correlation (pair counts in parentheses) of same-bundle seed pairs, same-family/distinct-checkpoint pairs and cross-family pairs; $\Delta_{\mathrm{ckpt}}$ is same-family minus cross-family, also restricted to the primary pool. (b) 11-bundle primary pool: $\Delta$ with DINOv2-B$\times$S pairs dropped from the same-bundle/release group or moved to the cross-bundle reference; distinct-checkpoint $\rho$ (pairs); and, against a cross reference that excludes all same-family pairs, $\Delta_{\mathrm{ckpt}}$ and the same-bundle gap $\Delta_{\mathrm{same}}$.}
  \label{tab:distinct-ckpt}
  \label{tab:primary-pool-decomp}
  \centering
  \setlength{\tabcolsep}{5pt}

  \par\noindent\textbf{(a) Distinct-checkpoint tier}\par\smallskip
  \begin{tabular*}{\linewidth}{@{\extracolsep{\fill}}TcPPPEE@{}}
    \toprule
    \bfseries \textbf{Task} & \bfseries \makecell[c]{$K$/\\avail.} & \multicolumn{1}{c}{\bfseries \makecell[c]{Same\\bundle}} & \multicolumn{1}{c}{\bfseries \makecell[c]{Same family,\\distinct ckpt}} & \multicolumn{1}{c}{\bfseries \makecell[c]{Cross\\family}} & \multicolumn{1}{c}{\bfseries \makecell[c]{$\Delta_{\mathrm{ckpt}}$\\{}[95\% CI]}} & \multicolumn{1}{c}{\bfseries \makecell[c]{$\Delta_{\mathrm{ckpt}}$, 11-bundle\\{}[95\% CI]}} \\
    \midrule
    Kvasir & 17/17 & 0.975 (102) & 0.736 (240) & 0.650 (1936) & 0.086 [0.047, 0.126] & 0.061 [0.003, 0.114] \\
\addlinespace[1pt]
ACDC LV & 15/17 & 0.989 (90) & 0.745 (160) & 0.613 (1520) & 0.131 [0.111, 0.148] & 0.150 [0.122, 0.176] \\
\addlinespace[1pt]
BraTS FG & 16/17 & 0.978 (96) & 0.670 (192) & 0.560 (1728) & 0.110 [0.085, 0.139] & 0.120 [0.086, 0.158] \\
\addlinespace[1pt]
RIGA Cup & 16/17 & 0.959 (96) & 0.400 (192) & 0.281 (1728) & 0.119 [0.058, 0.179] & 0.045 [\mn0.044, 0.126] \\
\addlinespace[1pt]
RIGA Disc & 17/17 & 0.970 (102) & 0.407 (240) & 0.194 (1936) & 0.213 [0.139, 0.280] & 0.165 [0.016, 0.281] \\
    \bottomrule
  \end{tabular*}

  \par\medskip
  \par\noindent\textbf{(b) Grouping sensitivity within the primary pool}\par\smallskip
  \begin{tabular*}{\linewidth}{@{\extracolsep{\fill}}TccPccc@{}}
    \toprule
    \bfseries \textbf{Task} & \bfseries \makecell[c]{$\Delta$, same-\\bundle only} & \bfseries \makecell[c]{$\Delta$, DINOv2\\as cross} & \multicolumn{1}{c}{\bfseries \makecell[c]{Distinct-ckpt\\$\rho$ (pairs)}} & \bfseries \makecell[c]{Cross ref.,\\excl. family} & \bfseries \makecell[c]{$\Delta_{\mathrm{ckpt}}$ vs.\\cross excl.} & \bfseries \makecell[c]{$\Delta_{\mathrm{same}}$ vs.\\cross excl.} \\
    \midrule
    Kvasir & 0.302 & 0.300 & 0.735 (48) & 0.674 & 0.061 & 0.303 \\
\addlinespace[1pt]
ACDC LV & 0.350 & 0.346 & 0.786 (32) & 0.636 & 0.150 & 0.352 \\
\addlinespace[1pt]
BraTS FG & 0.394 & 0.391 & 0.705 (32) & 0.585 & 0.120 & 0.396 \\
\addlinespace[1pt]
RIGA Cup & 0.643 & 0.639 & 0.356 (48) & 0.311 & 0.045 & 0.642 \\
\addlinespace[1pt]
RIGA Disc & 0.736 & 0.731 & 0.390 (48) & 0.225 & 0.165 & 0.740 \\
    \bottomrule
  \end{tabular*}

\end{table}

\section{Adaptation and pipeline controls}
\label{app:controls}

These controls separate the same/cross Dice-correlation gap from absolute dependence.

\subsection{Decoder-capacity control}
\label{app:unet-skip}

Table~\ref{tab:controls} compares the UNet-skip and frozen $1{\times}1$ readouts on the same source bundles. The proportional gap reduction is largest on ACDC LV and smallest on Kvasir; the gap remains positive on every measured task.

The UNet-skip decoder replaces the $1{\times}1$ readout on all 11 bundles, using frozen multi-scale features computed online, FP32 and the primary seeds, epochs and batch size. CLIP-B/16 and CLIP-L/14 use block taps from their QuickGELU wrappers. The nine-bundle rows omit these two separately wrapped encoders, and the gap reduction is similar without them; the frozen $1{\times}1$ gap is recomputed on the same nine source bundles, applying the mean-Dice floor separately to each readout. For ViT bundles, the four levels are transformer-block taps (ViT-B blocks 3, 6, 9, 11, 0-indexed as in the \extlink{https://github.com/Project-MONAI/MONAI/blob/1.6.0/monai/networks/nets/unetr.py}{MONAI implementation} of UNETR \citep{unetr2022}; CLIP-L/14 blocks 5, 11, 17, 23). These retain a single token resolution rather than forming a pyramid. Tap normalisation follows each wrapper (DINOv2: all taps; timm: last tap only; CLIP: none). The CNNs supply a true stride pyramid. Each of the four feature levels is projected to 128 channels by a biased $1{\times}1$ convolution. Three top-down stages bilinearly resize the deeper map to the next skip resolution, add the projected skip, and apply two $3{\times}3$ bias-free convolutions, each followed by BatchNorm and ReLU. After resizing to $224{\times}224$, one further Conv--BatchNorm--ReLU block and a biased $1{\times}1$ classifier produce the logits. All interpolations use \texttt{align\_corners=False}. Decoder parameter counts range from 1,157,505 to 1,558,913 across bundles; the per-bundle input channels are listed in the released decoder code (\path{code/src/fmpool/heads_unet.py} and \path{code/src/fmpool/multiscale.py}).

\begin{table}[htbp]
  \caption{Decoder-capacity controls. UNet-skip (11/9) denotes the source-bundle count; the nine-bundle pool excludes the two OpenAI CLIP bundles. The frozen reference uses the same source-bundle set and split, with the mean-Dice floor applied separately, so retained $K$ may differ on ACDC LV. Kvasir denotes the official split.}
  \label{tab:controls}
    \label{tab:app-controls}
  \centering

  \begin{tabular*}{\linewidth}{@{\extracolsep{\fill}}>{\raggedright\arraybackslash\collectcell\taskshort}p{1.65cm}<{\endcollectcell}>{\raggedright\arraybackslash\collectcell\ctllabel}p{1.8cm}<{\endcollectcell}cccEK@{}}
    \toprule
    \bfseries \textbf{Task} & \bfseries \textbf{Readout} & \bfseries $K$ & \bfseries $\rho_{\mathrm{same}}$ & \bfseries $\rho_{\mathrm{cross}}$ & \multicolumn{1}{c}{\bfseries $\Delta$ [95\% CI]} & \multicolumn{1}{c}{\bfseries \makecell[c]{Frozen $1{\times}1$\\$\Delta$ [95\% CI] ($K$)}} \\
    \midrule
    Kvasir (official) & UNet-skip readout, 11 bundles & 11 & 0.851 & 0.693 & 0.158 [0.115, 0.208] & \GapKvasirPrimary{} \GapKvasirPrimaryCI ($K{=}\KKvasir$) \\
\addlinespace[1pt]
ACDC LV & UNet-skip readout, 11 bundles & 11 & 0.908 & 0.824 & 0.084 [0.068, 0.100] & \GapAcdcPrimary{} \GapAcdcPrimaryCI ($K{=}\KAcdc$) \\
\addlinespace[1pt]
RIGA Cup & UNet-skip readout, 11 bundles & 11 & 0.806 & 0.637 & 0.169 [0.136, 0.209] & \GapCupPrimary{} \GapCupPrimaryCI ($K{=}\KCup$) \\
\addlinespace[1pt]
RIGA Disc & UNet-skip readout, 11 bundles & 11 & 0.811 & 0.559 & 0.252 [0.169, 0.339] & \GapDiscPrimary{} \GapDiscPrimaryCI ($K{=}\KDisc$) \\
\addlinespace[4pt]
Kvasir (official) & UNet-skip readout, 9 bundles (CLIP excluded) & 9 & 0.845 & 0.672 & 0.173 [0.125, 0.229] & \PrimaryNineBundleGapKvasir{} \PrimaryNineBundleGapKvasirCI ($K{=}\PrimaryNineBundleKKvasir$) \\
\addlinespace[1pt]
ACDC LV & UNet-skip readout, 9 bundles (CLIP excluded) & 9 & 0.902 & 0.816 & 0.086 [0.068, 0.103] & \PrimaryNineBundleGapAcdc{} \PrimaryNineBundleGapAcdcCI ($K{=}\PrimaryNineBundleKAcdc$) \\
\addlinespace[1pt]
RIGA Cup & UNet-skip readout, 9 bundles (CLIP excluded) & 9 & 0.802 & 0.631 & 0.171 [0.131, 0.219] & \PrimaryNineBundleGapCup{} \PrimaryNineBundleGapCupCI ($K{=}\PrimaryNineBundleKCup$) \\
\addlinespace[1pt]
RIGA Disc & UNet-skip readout, 9 bundles (CLIP excluded) & 9 & 0.804 & 0.540 & 0.264 [0.174, 0.352] & \PrimaryNineBundleGapDisc{} \PrimaryNineBundleGapDiscCI ($K{=}\PrimaryNineBundleKDisc$) \\
    \bottomrule
  \end{tabular*}
\end{table}

\subsection{End-to-end fine-tuning control}
\label{app:ft}

Training uses encoder learning rates of $10^{-4}$ or $10^{-5}$, a head learning rate of $10^{-3}$, FP32, 30 epochs, batch size 16 and seeds 42--45. Evaluation uses the final-epoch checkpoint with the primary splits and per-case Dice convention. Pair grouping follows the starting encoder bundle/release; encoder weights can diverge across seeds during fine-tuning. Encoders run in training mode: BatchNorm statistics are re-estimated for ResNet-18/50 and EfficientNet-B0, and stochastic depth is active for ConvNeXt-T and EfficientNet-B0, while training mode adds no stochastic depth to the ViT bundles. Optimisation uses Adam without weight decay and BCE loss, with a constant learning rate, no warm-up and no augmentation.

One fine-tuned head on the content-group Kvasir split (CLIP-L/14, encoder $10^{-4}$, seed 42) collapsed to near-zero Dice and remains in the pool. Excluding it gives \FtGapExclCollapsedLrEMinusFourKvasirGroupedSens~\FtGapExclCollapsedLrEMinusFourKvasirGroupedSensCI\ on that split. No head collapsed on the official split at either learning rate. The paired fine-tuned-minus-frozen intervals (last column of Table~\ref{tab:ft}) exclude zero on RIGA Cup and the content-group split at both learning rates. On the official split, they exclude zero at $10^{-4}$ but include it at $10^{-5}$.

Fine-tuning does not uniformly raise per-bundle Dice. At an encoder learning rate of $10^{-4}$, Dice falls below the frozen-readout value for DINOv2-S and DINOv2-B on the official Kvasir split and for DINOv2-S, DINOv2-B and CLIP-L/14 on the content-group split. On RIGA Cup, Dice is lower than the frozen-readout value at an encoder learning rate of $10^{-5}$ for EfficientNet-B0 (\FtMeanDiceLrEMinusFiveCupEfficientnetBZero\ vs.\ \DiceCupEfficientnetBZero\ frozen).

\paragraph{Failure dependence under adaptation.} Table~\ref{tab:adaptation-meff} applies the effective-size definition and four-distinct-bundle enumeration of Appendix~\ref{app:meff} to frozen and fine-tuned traces. Each pool uses one seed per bundle. The $P_{33}$ threshold is computed from each recipe's pooled model-case scores and held fixed during the paired case bootstrap ($B{=}2000$, seed 0). Every member must have at least five failures and five passes; unsupported pools are omitted within each draw. Recipes therefore share a threshold rule, not a numerical Dice threshold. The Kvasir difference intervals include zero at both learning rates. Cup effective size decreases at both rates despite a smaller Pearson gap.

\begin{table}[htbp]
  \caption{Fine-tuning and other pool controls. FT, $10^{-k}$ denotes end-to-end fine-tuning at encoder learning rate $10^{-k}$; for the random-initialisation CNN, same/cross compares same-architecture with cross-architecture pairs. Rows show the retained bundles, correlation levels and same/cross gap. Kvasir denotes the official split unless marked grouped; paired fine-tuned-minus-frozen comparisons appear in Table~\ref{tab:ft}.}
  \label{tab:controls-other}
  \centering
  \begin{tabular}{>{\raggedright\arraybackslash\collectcell\taskshort}p{1.8cm}<{\endcollectcell}>{\raggedright\arraybackslash\collectcell\ctllabel}p{2.4cm}<{\endcollectcell}cccE}
    \toprule
    \bfseries Task & \bfseries Control & \bfseries $K$ & \bfseries $\rho_{\mathrm{same}}$ & \bfseries $\rho_{\mathrm{cross}}$ & \multicolumn{1}{c}{\bfseries $\Delta$ [95\% CI]} \\
    \midrule
    Kvasir (official) & FT, $10^{-4}$ & 11 & 0.796 & 0.615 & 0.182 [0.133, 0.237] \\
\addlinespace[1pt]
Kvasir (official) & FT, $10^{-5}$ & 11 & 0.890 & 0.652 & 0.238 [0.172, 0.306] \\
\addlinespace[1pt]
Kvasir (content-group split, sensitivity) & FT, $10^{-4}$ & 11 & 0.776 & 0.601 & 0.175 [0.114, 0.250] \\
\addlinespace[1pt]
Kvasir (content-group split, sensitivity) & FT, $10^{-5}$ & 11 & 0.884 & 0.657 & 0.227 [0.142, 0.328] \\
\addlinespace[1pt]
RIGA Cup & FT, $10^{-4}$ & 11 & 0.871 & 0.657 & 0.214 [0.171, 0.259] \\
\addlinespace[1pt]
RIGA Cup & FT, $10^{-5}$ & 11 & 0.864 & 0.581 & 0.284 [0.225, 0.347] \\
\addlinespace[1pt]
ACDC LV & random-initialisation CNN & 2 & 0.909 & 0.776 & 0.133 [0.089, 0.179] \\
\addlinespace[1pt]
RIGA Cup & random-initialisation CNN & 2 & 0.840 & 0.523 & 0.317 [0.193, 0.430] \\
    \bottomrule
  \end{tabular}
\end{table}

\begin{table}[htbp]
  \caption{Matched fine-tuning comparisons. Frozen $1{\times}1$ gaps use the same source-bundle set and split; the final column gives the paired fine-tuned-minus-frozen gap with its 95\% interval. FT estimates are in Table~\ref{tab:controls-other}.}
  \label{tab:ft}
  \centering
  \begin{tabular}{>{\raggedright\arraybackslash\collectcell\taskshort}p{1.8cm}<{\endcollectcell}>{\raggedright\arraybackslash\collectcell\ctllabel}p{2.4cm}<{\endcollectcell}KE}
    \toprule
    \bfseries Task & \bfseries Recipe & \multicolumn{1}{c}{\bfseries \makecell[c]{Frozen $\Delta$\\{}[95\% CI] ($K$)}} & \multicolumn{1}{c}{\bfseries \makecell[c]{FT $-$ frozen $\Delta$\\(paired) [95\% CI]}} \\
    \midrule
    Kvasir (official) & FT, $10^{-4}$ & 0.265 [0.217, 0.317] ($K{=}11$) & \mn0.083 [\mn0.147, \mn0.018] \\
\addlinespace[1pt]
Kvasir (official) & FT, $10^{-5}$ & 0.265 [0.217, 0.317] ($K{=}11$) & \mn0.027 [\mn0.102, 0.048] \\
\addlinespace[1pt]
Kvasir (content-group split, sensitivity) & FT, $10^{-4}$ & 0.323 [0.257, 0.403] ($K{=}11$) & \mn0.148 [\mn0.223, \mn0.071] \\
\addlinespace[1pt]
Kvasir (content-group split, sensitivity) & FT, $10^{-5}$ & 0.323 [0.257, 0.403] ($K{=}11$) & \mn0.096 [\mn0.175, \mn0.006] \\
\addlinespace[1pt]
RIGA Cup & FT, $10^{-4}$ & 0.557 [0.500, 0.609] ($K{=}11$) & \mn0.344 [\mn0.391, \mn0.288] \\
\addlinespace[1pt]
RIGA Cup & FT, $10^{-5}$ & 0.557 [0.500, 0.609] ($K{=}11$) & \mn0.274 [\mn0.335, \mn0.208] \\
    \bottomrule
  \end{tabular}
\par\bigskip
  \caption{Effective size under fine-tuning. Nominal four-member pools at each recipe's pooled $P_{33}$; paired FT-minus-frozen intervals hold both observed thresholds fixed. Frozen marginal intervals are recomputed using these paired draws, so the RIGA Cup frozen interval differs slightly from Table~\ref{tab:meff}(b). Support counts refer to the observed cohort. Pale blue marks the paired changes.}
  \label{tab:adaptation-meff}
  \centering
  \begin{tabular*}{\linewidth}{@{\extracolsep{\fill}}llcEEc@{}}
    \toprule
    \bfseries Task & \bfseries Recipe & \bfseries $P_{33}$ & \multicolumn{1}{c}{\bfseries $M_{\mathrm{eff}}$ [95\% CI]} & \multicolumn{1}{c}{\bfseries \makecell{FT $-$ frozen\\{}[95\% CI]}} & \bfseries \makecell{Supported/\\total pools} \\
    \midrule
    Kvasir & Frozen & 0.569 & 1.559 [1.454, 1.704] & -- & 84{,}480/84{,}480 \\
\addlinespace[1pt]
Kvasir & FT, $10^{-4}$ & 0.760 & 1.630 [1.497, 1.801] & \multicolumn{1}{>{\columncolor{TableFocus}}c}{\estci{0.071 [\mn0.082, 0.224]}} & 84{,}480/84{,}480 \\
\addlinespace[1pt]
Kvasir & FT, $10^{-5}$ & 0.799 & 1.552 [1.437, 1.685] & \multicolumn{1}{>{\columncolor{TableFocus}}c}{\estci{\mn0.008 [\mn0.164, 0.141]}} & 84{,}480/84{,}480 \\
\addlinespace[1pt]
RIGA Cup & Frozen & 0.521 & 2.420 [2.182, 2.737] & -- & 84{,}480/84{,}480 \\
\addlinespace[1pt]
RIGA Cup & FT, $10^{-4}$ & 0.781 & 1.649 [1.513, 1.853] & \multicolumn{1}{>{\columncolor{TableFocus}}c}{\estci{\mn0.771 [\mn1.090, \mn0.508]}} & 84{,}480/84{,}480 \\
\addlinespace[1pt]
RIGA Cup & FT, $10^{-5}$ & 0.764 & 1.747 [1.605, 1.939] & \multicolumn{1}{>{\columncolor{TableFocus}}c}{\estci{\mn0.673 [\mn0.986, \mn0.411]}} & 84{,}480/84{,}480 \\
    \bottomrule
  \end{tabular*}
\end{table}

\subsection{Random-initialisation and medical-encoder controls}
\label{app:random-medical-controls}

The random-initialisation CNN control jointly trains randomly initialised ConvNeXt-T or ResNet-50 encoders and their $1{\times}1$ heads on ACDC LV and RIGA Cup. Training uses FP32 Adam (learning rate $10^{-3}$ for both encoder and head), BCE loss, 30 epochs and batch size 16, with eight seeds per architecture (seeds 42--49; 16 models per task). Here, same/cross compares pairs of runs using the same backbone architecture with pairs using different architectures. Its positive gap shows that this structure also occurs without pretrained weights (Table~\ref{tab:controls-other}). The 15-bundle frozen medical-encoder pool (Appendix~\ref{app:medsam-note}, Table~\ref{tab:medenc}(b)) has a gap close to the primary gap on RIGA Cup and larger on RIGA Disc and Kvasir. Both controls differ from the primary analysis in pool composition and size.

\subsection{nnU-Net evaluated on the same test cases}
\label{app:nnunet}

The nnU-Net v2 control in Table~\ref{tab:nnunet} uses five folds and two seeds, scoring every model on identical test cases. Same-fold pairs have higher mean correlation than different-fold pairs; this fold gap is $\Delta_{\mathrm{fold}}=\rho_{\text{same-fold}}-\rho_{\text{cross-fold}}$.

For RIGA Cup, nnU-Net v2 uses its self-configuring 2D configuration on the same $224{\times}224$ resized images and consensus cup masks (raw label 128) as the primary protocol. Training runs for \NnunetCupEpochs\ epochs, fewer than the default 1000 because of the compute budget, with five folds $\times$ seeds $\{13, 37\}$ (\NnunetCupModels\ models). Each seed fixes only network initialisation; augmentation randomness is not fixed. The seeds therefore label replicate runs, which are not bitwise reproducible.

The ACDC counterpart uses the \texttt{3d\_fullres} configuration with 3D frame volumes as cases and nnU-Net's \texttt{SimpleITKIO} geometry convention. Its five folds partition official training patients 001--100 at the patient level. Every model is scored on the same \NnunetAcdcNTest\ frame volumes: the ED and ES frames of each of the \NAcdcSubjects\ official test patients. Both tasks use \NnunetAcdcEpochs\ epochs and identical test cases across models. Test predictions (the \NnunetCupNTest\ Magrabia images for RIGA Cup) use the primary per-case Dice convention: empty prediction and empty target $=1$, with no smoothing. nnU-Net's own evaluator instead returns NaN for empty/empty cases and likewise uses no smoothing term. Same-fold pairs are the two seeds of one fold; all other pairs are different-fold pairs.

\begin{table}[htbp]
  \caption{nnU-Net evaluated on the same test cases. Both tasks use \NnunetAcdcEpochs\ epochs and \NnunetAcdcModels\ models (five folds, two seeds). $N$ counts test frame volumes for ACDC LV and test images for RIGA Cup. $\rho_{\text{same-fold}}$ and $\rho_{\text{cross-fold}}$ are mean correlations within and between folds, and $\Delta_{\mathrm{fold}}$ is their difference. $M_{\mathrm{eff}}$ uses one model per fold at training seed 13, with $P_{33}$ pooled over their case scores.}
  \label{tab:nnunet}
  \centering
  \begin{tabular}{Tccccc}
    \toprule
    \bfseries Task & \bfseries $N$ & \bfseries Mean Dice & \bfseries \makecell[c]{$\rho_{\text{same-fold}}$\\$\rho_{\text{cross-fold}}$} & \bfseries $\Delta_{\mathrm{fold}}$ [95\% CI] & \bfseries $M_{\mathrm{eff}}$ [95\% CI] \\
    \midrule
    ACDC LV & \NnunetAcdcNTest & \NnunetAcdcMeanDice & \makecell[c]{\NnunetAcdcWithinRbar\\\NnunetAcdcCrossRbar} & \makecell[c]{\NnunetAcdcGap\\\NnunetAcdcGapCI} & \makecell[c]{\NnunetAcdcMeff\\\NnunetAcdcMeffCI} \\
    RIGA Cup & \NnunetCupNTest & \NnunetCupMeanDice & \makecell[c]{\NnunetCupWithinRbar\\\NnunetCupCrossRbar} & \makecell[c]{\NnunetCupGap\\\NnunetCupGapCI} & \makecell[c]{\NnunetCupMeff\\\NnunetCupMeffCI} \\
    \bottomrule
  \end{tabular}
\end{table}

\subsection{Alternative failure-event metrics for stronger pipelines}
\label{app:pipeline-metrics}

Table~\ref{tab:pipeline-metrics} applies $\kappa$, $Q$, disagreement and double-fault rates to UNet-skip Cup/Disc traces and nnU-Net Cup traces evaluated on the same test cases. This diagnostic covers only the RIGA tasks (UNet-skip on Cup and Disc; nnU-Net on Cup, its only 2D task). The mean-Dice floor is applied by bundle. The pooled model-case $P_{33}$ is held fixed during the paired case bootstrap ($B{=}2000$, seed 0); the $\kappa$ convention follows Appendix~\ref{app:failure-event}. UNet-skip uses the primary same-bundle/release grouping; nnU-Net compares same-fold with different-fold pairs. The $\kappa$ and $Q$ gaps support the same ordering as the Pearson gaps. At the stated threshold, same-group pairs also have more joint failures and fewer disagreements, but these two rates additionally depend on the members' marginal failure probabilities. All reported gap intervals exclude zero. Every member must have at least five failures and five passes on the observed cohort and each resample. All resamples meet this requirement, and all pairs have finite values here.

\begin{table}[htbp]
  \caption{Failure-event gaps for stronger pipelines at pooled $P_{33}$. Each cell gives the gap and its paired case-bootstrap 95\% interval: same-bundle/release minus cross-bundle for UNet-skip, same-fold minus cross-fold for nnU-Net.}
  \label{tab:pipeline-metrics}
  \centering
  \begin{tabular*}{\linewidth}{@{\extracolsep{\fill}}llcccc@{}}
    \toprule
    \bfseries Task & \bfseries Pipeline & \bfseries $\kappa$ & \bfseries $Q$ & \bfseries Disagreement & \bfseries Double fault \\
    \midrule
    RIGA Cup & UNet-skip & \makecell[c]{\PipelineUnetCupKappaGap\\\PipelineUnetCupKappaGapCI} & \makecell[c]{\PipelineUnetCupQGap\\\PipelineUnetCupQGapCI} & \makecell[c]{\PipelineUnetCupDisagreementGap\\\PipelineUnetCupDisagreementGapCI} & \makecell[c]{\PipelineUnetCupDoubleFaultGap\\\PipelineUnetCupDoubleFaultGapCI} \\
    RIGA Disc & UNet-skip & \makecell[c]{\PipelineUnetDiscKappaGap\\\PipelineUnetDiscKappaGapCI} & \makecell[c]{\PipelineUnetDiscQGap\\\PipelineUnetDiscQGapCI} & \makecell[c]{\PipelineUnetDiscDisagreementGap\\\PipelineUnetDiscDisagreementGapCI} & \makecell[c]{\PipelineUnetDiscDoubleFaultGap\\\PipelineUnetDiscDoubleFaultGapCI} \\
    RIGA Cup & nnU-Net & \makecell[c]{\PipelineNnunetCupKappaGap\\\PipelineNnunetCupKappaGapCI} & \makecell[c]{\PipelineNnunetCupQGap\\\PipelineNnunetCupQGapCI} & \makecell[c]{\PipelineNnunetCupDisagreementGap\\\PipelineNnunetCupDisagreementGapCI} & \makecell[c]{\PipelineNnunetCupDoubleFaultGap\\\PipelineNnunetCupDoubleFaultGapCI} \\
    \bottomrule
  \end{tabular*}
\end{table}

\FloatBarrier
\subsection{Native-3D abdominal comparison}
\label{app:amos}

We evaluate released MRSegmentator folds 0 and 1 (release tag \texttt{v1.2.0}) and TotalSegmentator \texttt{2.4.0} (\texttt{total} for CT, \texttt{total\_mr} for MRI) on AMOS22. The convenience sample comprises the first \AmosCTN\ labelled CT cases in sorted case-ID order and all \AmosMRIN\ labelled MRI cases. Inference uses each system's native 3D pipeline. Prediction and ground-truth labels are resampled with nearest-neighbour interpolation to a $1.5{\times}1.5{\times}3$\,mm grid for volumetric scoring, so evaluation is not on the original acquisition grid.

\begin{table}[htbp]
  \caption{Native-3D AMOS22 comparison. Two MRSegmentator folds and one modality-specific TotalSegmentator model; nominal pool size $M{=}3$, with $M_{\mathrm{eff}}$ evaluated at pooled $P_{33}$; $\Delta_{\mathrm{sys}}$ is same-system minus cross-system $r$.}
  \label{tab:amos}
  \centering
  \begin{tabular*}{\linewidth}{@{\extracolsep{\fill}}lcEEEE@{}}
    \toprule
    \bfseries Modality & \bfseries $N$ & \multicolumn{1}{c}{\bfseries \makecell{Same-system\\$r$ [95\% CI]}} & \multicolumn{1}{c}{\bfseries \makecell{Cross-system\\$r$ [95\% CI]}} & \multicolumn{1}{c}{\bfseries $\Delta_{\mathrm{sys}}$ [95\% CI]} & \multicolumn{1}{c}{\bfseries $M_{\mathrm{eff}}$ [95\% CI]} \\
    \midrule
    CT & 60 & 0.988 [0.980, 0.994] & 0.976 [0.960, 0.987] & 0.012 [0.001, 0.029] & 1.071 [1.016, 1.154] \\
\addlinespace[1pt]
MRI & 60 & 0.975 [0.953, 0.987] & 0.918 [0.861, 0.951] & 0.057 [0.026, 0.105] & 1.268 [1.144, 1.466] \\
    \bottomrule
  \end{tabular*}
\end{table}

\begin{table}[htbp]
  \caption{AMOS22 failure-event gaps at pooled $P_{33}$. Each cell gives the same-system-minus-cross-system gap and its paired volume-bootstrap 95\% interval.}
  \label{tab:amos-alternatives}
  \centering
  \begin{tabular*}{\linewidth}{@{\extracolsep{\fill}}lcccc@{}}
    \toprule
    \bfseries Modality & \bfseries $\kappa$ & \bfseries $Q$ & \bfseries Disagreement & \bfseries Double fault \\
    \midrule
    CT & \makecell[c]{\AmosCTKappaGap\\\AmosCTKappaGapCI} & \makecell[c]{\AmosCTQGap\\\AmosCTQGapCI} & \makecell[c]{\AmosCTDisagreementGap\\\AmosCTDisagreementGapCI} & \makecell[c]{\AmosCTDoubleFaultGap\\\AmosCTDoubleFaultGapCI} \\
    MRI & \makecell[c]{\AmosMRIKappaGap\\\AmosMRIKappaGapCI} & \makecell[c]{\AmosMRIQGap\\\AmosMRIQGapCI} & \makecell[c]{\AmosMRIDisagreementGap\\\AmosMRIDisagreementGapCI} & \makecell[c]{\AmosMRIDoubleFaultGap\\\AmosMRIDoubleFaultGapCI} \\
    \bottomrule
  \end{tabular*}
\end{table}

The \AmosOrgans\ common targets are spleen, right/left kidney, gallbladder, esophagus, liver, stomach, aorta, inferior vena cava, pancreas, right/left adrenal gland and duodenum. MONAI 1.5.0 Dice with \texttt{ignore\_empty=True} assigns undefined scores to empty ground-truth organs. Within each modality the three models share the same missingness pattern (8 undefined case-organ cells per model on CT and 6 on MRI), leaving \AmosMinOrgans--\AmosMaxOrgans\ finite organ scores per case. The case score averages only those finite organ scores. Every case mean is finite.

The same-system statistic correlates the two MRSegmentator score traces; the cross-system statistic averages each trace's correlation with the TotalSegmentator trace. Pearson intervals jointly resample volumes ($B{=}2000$). Failure metrics use the pooled $P_{33}$ (CT \AmosCTThreshold, MRI \AmosMRIThreshold), held fixed during bootstrapping. Equation~\eqref{eq:effective-pool-size} applies to the three fixed members, each requiring at least five failures and passes; support is rechecked on each resample. Table~\ref{tab:amos} reports the absolute levels, gap and effective size.

MRSegmentator training partly uses TotalSegmentator-assisted labels, so the systems also share upstream lineage. The comparison illustrates that a small gap can coexist with high absolute dependence; it does not identify the cause or generalise to diverse 3D model families. At $P_{33}$, all CT alternative-metric gap intervals include zero. MRI $\kappa$, $Q$ and disagreement intervals also include zero, while the double-fault gap is negative (Table~\ref{tab:amos-alternatives}). Thus, the Pearson ordering does not hold across all failure metrics. Assessing broader 3D scope first requires a larger cohort and a system outside the nnU-Net family.

\FloatBarrier
\section{Medical-domain encoders}
\label{app:medsam-note}

Table~\ref{tab:medenc} summarises RETFound, Phikon, RAD-DINO and BiomedParse as a pool, Table~\ref{tab:medenc-per-encoder} compares each with the 11 primary bundles, and Table~\ref{tab:medenc}(b) reports the 15-bundle extended pool. Mean mutual correlations among the medical encoders range from \MedEncMutualDisc\ to \MedEncMutualKvasir. Their pooled mean correlation with primary bundles is near the cross-bundle reference on RIGA Cup and slightly below it on RIGA Disc and Kvasir. Against the full cross-bundle reference, pooled $\Delta_{\mathrm{med}}$ is \MedEncPooledCupDiff\ on RIGA Cup, \MedEncPooledDiscDiff\ on RIGA Disc and \MedEncPooledKvasirDiff\ on Kvasir (Table~\ref{tab:medenc-per-encoder}). The extended-pool gap remains close to the primary value on RIGA Cup and increases on RIGA Disc and Kvasir. Values that involve Phikon cannot be recomputed from the release, because Owkin's licence restricts sharing its outputs (Appendix~\ref{app:broader-impacts}); the RETFound, RAD-DINO and BiomedParse rows of Table~\ref{tab:medenc-per-encoder} can.

\begin{table}[!htbp]
  \caption{Medical-encoder pooled summary. (a) Mutual $\rho$: mean per-case Dice correlation over seed pairs from distinct retained medical bundles. Pooled $\Delta_{\mathrm{med}}$: mean medical-to-primary correlation minus the primary cross-bundle mean (not a same/cross gap), with BiomedCLIP, the primary pool's medically pretrained encoder, excluded from both primary sets; Table~\ref{tab:medenc-per-encoder} gives the full-reference difference. (b) 15-bundle extended pool ($K{=}15$ after the floor) and 11-bundle primary gap.}
  \label{tab:medenc}
  \centering

  \par\noindent\textbf{(a) Dependence among medical encoders and with the primary pool}\par\smallskip
  \begin{tabular*}{\linewidth}{@{\extracolsep{\fill}}TEE@{}}
    \toprule
    \bfseries Task & \multicolumn{1}{c}{\bfseries Mutual $\rho$ [95\% CI]} & \multicolumn{1}{c}{\bfseries \makecell[c]{Pooled $\Delta_{\mathrm{med}}$ [95\% CI]\\(excl.\ BiomedCLIP)}} \\
    \midrule
    Kvasir & 0.647 [0.568, 0.720] & \mn0.059 [\mn0.113, \mn0.010] \\
\addlinespace[1pt]
RIGA Cup & 0.449 [0.366, 0.535] & 0.044 [\mn0.012, 0.096] \\
\addlinespace[1pt]
RIGA Disc & 0.136 [0.021, 0.209] & \mn0.065 [\mn0.112, \mn0.007] \\
    \bottomrule
  \end{tabular*}

  \par\medskip
  \par\noindent\textbf{(b) Extended and primary pools}\par\smallskip
  \begin{tabular*}{\linewidth}{@{\extracolsep{\fill}}TccEE@{}}
    \toprule
    \bfseries Task & \bfseries \makecell[c]{15-bundle\\$\rho_{\mathrm{same}}$} & \bfseries \makecell[c]{15-bundle\\$\rho_{\mathrm{cross}}$} & \multicolumn{1}{c}{\bfseries \makecell[c]{15-bundle $\Delta$\\{}[95\% CI]}} & \multicolumn{1}{c}{\bfseries \makecell[c]{11-bundle $\Delta$\\{}[95\% CI]}} \\
    \midrule
    Kvasir & 0.950 & 0.649 & 0.302 [0.254, 0.357] & 0.265 [0.217, 0.317] \\
\addlinespace[1pt]
RIGA Cup & 0.889 & 0.338 & 0.551 [0.498, 0.602] & 0.557 [0.500, 0.609] \\
\addlinespace[1pt]
RIGA Disc & 0.896 & 0.197 & 0.698 [0.635, 0.741] & 0.640 [0.570, 0.689] \\
    \bottomrule
  \end{tabular*}
\end{table}

The four encoders use the same frozen FP32 $1{\times}1$ protocol, with four decoder seeds and the same epochs, batch size and optimiser. RETFound is a ViT-L/16 MAE pretrained on colour fundus photographs; its CLS token is dropped and its patch tokens are reshaped to a spatial map. Phikon is pretrained on histopathology and is used with its official LayerNorm $\epsilon$. RAD-DINO is pretrained on chest radiographs. BiomedParse uses a FocalNet backbone \citep{focalnet2022}, whose final-stage features are used. Each added encoder is scored against the 11 primary bundles under the primary grouping. This encoder-only probe differs from the official fine-tuning or prompting pipelines.

RAD-DINO is initialised from DINOv2-B, a primary-pool member, so its pairs with DINOv2-B share lineage. DINOv2-B is RAD-DINO's most correlated primary bundle on RIGA Cup, but not on RIGA Disc or Kvasir. BiomedParse's training corpus includes optic disc/cup and polyp segmentation datasets, as well as ACDC, MSD and ISIC-2018. Its RIGA and Kvasir tasks may therefore involve training exposure to related images. BiomedParse is not run on the ACDC, MSD or ISIC tasks, and overlap between its corpus and the test images is untested.

\begin{table}[htbp]
  \caption{Per-encoder medical probe. $\rho$ to primary bundles: mean per-case Dice correlation between the encoder's four heads and all retained primary-bundle heads; $\Delta_{\mathrm{med}}$: that value minus $\rho_{\mathrm{cross}}$, not a same/cross gap.}
  \label{tab:medenc-per-encoder}
  \centering

  \begin{tabular}{TlcEcE}
    \toprule
    \bfseries Task & \bfseries Encoder & \bfseries Mean Dice & \multicolumn{1}{c}{\bfseries \makecell[c]{$\rho$ to primary\\bundles [95\% CI]}} & \bfseries $\rho_{\mathrm{cross}}$ & \multicolumn{1}{c}{\bfseries $\Delta_{\mathrm{med}}$ [95\% CI]} \\
    \midrule
    Kvasir & RETFound & 0.515 & 0.597 [0.505, 0.677] & 0.675 & \mn0.078 [\mn0.164, \mn0.002] \\
\addlinespace[1pt]
Kvasir & Phikon & 0.616 & 0.685 [0.591, 0.759] & 0.675 & 0.010 [\mn0.050, 0.059] \\
\addlinespace[1pt]
Kvasir & RAD-DINO & 0.571 & 0.593 [0.488, 0.681] & 0.675 & \mn0.082 [\mn0.167, \mn0.007] \\
\addlinespace[1pt]
Kvasir & BiomedParse & 0.715 & 0.592 [0.487, 0.689] & 0.675 & \mn0.083 [\mn0.176, 0.002] \\
\addlinespace[1pt]
Kvasir & pooled medical & -- & 0.616 [0.541, 0.681] & 0.675 & \mn0.058 [\mn0.106, \mn0.012] \\
\addlinespace[4pt]
RIGA Cup & RETFound & 0.740 & 0.357 [0.249, 0.459] & 0.310 & 0.048 [\mn0.047, 0.127] \\
\addlinespace[1pt]
RIGA Cup & Phikon & 0.673 & 0.410 [0.316, 0.497] & 0.310 & 0.100 [0.017, 0.168] \\
\addlinespace[1pt]
RIGA Cup & RAD-DINO & 0.525 & 0.339 [0.231, 0.436] & 0.310 & 0.029 [\mn0.061, 0.106] \\
\addlinespace[1pt]
RIGA Cup & BiomedParse & 0.743 & 0.323 [0.238, 0.408] & 0.310 & 0.013 [\mn0.068, 0.090] \\
\addlinespace[1pt]
RIGA Cup & pooled medical & -- & 0.357 [0.284, 0.427] & 0.310 & 0.047 [\mn0.005, 0.096] \\
\addlinespace[4pt]
RIGA Disc & RETFound & 0.936 & 0.040 [\mn0.056, 0.139] & 0.230 & \mn0.190 [\mn0.302, \mn0.087] \\
\addlinespace[1pt]
RIGA Disc & Phikon & 0.925 & 0.280 [0.206, 0.377] & 0.230 & 0.051 [\mn0.010, 0.126] \\
\addlinespace[1pt]
RIGA Disc & RAD-DINO & 0.895 & 0.212 [0.159, 0.336] & 0.230 & \mn0.018 [\mn0.076, 0.096] \\
\addlinespace[1pt]
RIGA Disc & BiomedParse & 0.900 & 0.131 [0.019, 0.241] & 0.230 & \mn0.099 [\mn0.214, 0.008] \\
\addlinespace[1pt]
RIGA Disc & pooled medical & -- & 0.166 [0.122, 0.229] & 0.230 & \mn0.064 [\mn0.112, \mn0.007] \\
    \bottomrule
  \end{tabular}
\end{table}

\section{Held-out tasks and pool selection}
\label{app:heldout}

\subsection{Held-out tasks}
\label{app:heldout-tasks}
Table~\ref{tab:heldout}(a) reports four held-out tasks using the 11 bundles and four decoder seeds under the same protocol. ISIC-2018 and MSD Hippocampus yield functional pools with positive gaps; MSD Spleen is degenerate, and MSD Heart has no functional pool. The held-out tasks are 2D binary derivatives evaluated under the primary protocol, rather than official challenge evaluations. ISIC-2018 uses the official Task~1 training/test split; the MSD tasks use task-specific splits.

Each MSD volume contributes one central slice at $z=\lfloor Z/2\rfloor$ along its third spatial axis. Intensities are min--max scaled within that slice to uint8, replicated into three channels and resized to $224{\times}224$. Masks use nearest-neighbour resizing, with $\mathrm{seg}>0$ as the target. Hippocampus combines its head/body labels, Heart targets the left atrium, and Spleen targets spleen foreground. Thus, $N$ counts single central slices (one per volume), not whole-volume Dice evaluations. The slice-selection rule can yield empty targets even when the volume contains foreground. The split units are:
\begin{itemize}
  \item ISIC-2018 \citep{isic2019,ham10000,isic_msk2017}: \HeldoutSplitUnitIsic.
  \item MSD Hippocampus \citep{msd2022,msd2019simpson}: \HeldoutSplitUnitHippocampus.
  \item MSD Spleen: \splitunit{\HeldoutSplitUnitSpleen}.
  \item MSD Heart: \splitunit{\HeldoutSplitUnitHeart}.
\end{itemize}
The official ISIC Archive metadata (accessed 25 October 2026) match all 1000 test-image IDs in the released traces. Patient IDs are available for 890 test images from 711 patients and for 390 training images from 350 patients. The partitions share 48 known patient IDs, involving 53 test images; 47 of these 53 test images also share a known lesion ID with training images.\footnote{Official training and test metadata: \href{https://api.isic-archive.com/collections/63/metadata/}{\nolinkurl{api.isic-archive.com/collections/63/metadata}} and \href{https://api.isic-archive.com/collections/64/metadata/}{\nolinkurl{api.isic-archive.com/collections/64/metadata}}.} The reported ISIC results retain the official image benchmark: bootstrap draws and calibration/evaluation splits use images. These analyses do not establish patient-disjoint generalisation, and unknown patient IDs prevent a complete overlap assessment.\par The Hippocampus case bootstrap treats the \HeldoutNTestHippocampus\ test volumes as independent. However, MSD releases left/right crops of the same subject without a mapping, so crops from one subject may appear in both training and test partitions or together in the test set. Patient independence is not established, and the case-bootstrap interval may understate uncertainty.

MSD Spleen passes the floor only through empty targets: \SpleenGtEmptyCases\ of its \SpleenTestCases\ test targets are empty. Of its \SpleenPredHeads\ heads, \SpleenPredHeadsAllEmpty\ predict no foreground on any test slice (\SpleenPredNonemptyCaseHeads\ of \SpleenPredCaseHeads\ case-head predictions are non-empty). Per-case Dice is 1 for \SpleenEmptyGtDiceOnePairs\ model--case pairs with empty targets and 0 for \SpleenNonemptyGtDiceZeroPairs\ model--case pairs with non-empty targets. Its near-zero gap is therefore degenerate (the gap interval retains $1987/2000$ bootstrap draws), and Spleen is excluded from pool selection.

MSD Heart gives no functional pool (\HeldoutKHeart\ of \HeldoutKAvailHeart\ bundles pass the mean-Dice${\geq}0.30$ floor) and is excluded from selection.

\begin{table}[p]
  \caption{Held-out tasks and encoder-family pool selection. (a) Frozen $1{\times}1$ analysis on the five primary tasks and four held-out task derivatives. (b) Over \PoolSelSplitsKvasir\ calibration/evaluation half-splits per task: the count and share of splits whose four-member pool, selected by encoder family and calibration Dice, has evaluation-half $M_{\mathrm{eff}}<2$ (95\% Clopper--Pearson intervals over random splits, conditional on the fixed cohort); unsupported-split counts; and the share among supported splits. MSD Spleen (degenerate) and MSD Heart (no functional pool) do not enter pool selection; Appendix~\ref{app:heldout-tasks} gives the reasons and the image-level resampling used for ISIC.}
  \label{tab:heldout}
  \label{tab:poolsel}
  \centering

  \par\noindent\textbf{(a) Frozen $1{\times}1$ readout}\par\smallskip
  \begin{tabular*}{\linewidth}{@{\extracolsep{\fill}}lcccEc@{}}
    \toprule
    \bfseries \textbf{Task} & \bfseries $K$/avail. & \bfseries $\rho_{\mathrm{same}}$ & \bfseries $\rho_{\mathrm{cross}}$ & \multicolumn{1}{c}{\bfseries $\Delta$ [95\% CI]} & \bfseries $N$ \\
    \midrule
    Kvasir & 11/11 & 0.940 & 0.675 & 0.265 [0.217, 0.317] & 120 \\
\addlinespace[1pt]
ACDC LV & 10/11 & 0.955 & 0.638 & 0.317 [0.289, 0.344] & 977 \\
\addlinespace[1pt]
BraTS FG & 10/11 & 0.927 & 0.587 & 0.340 [0.296, 0.388] & 3365 \\
\addlinespace[1pt]
RIGA Cup & 11/11 & 0.867 & 0.310 & 0.557 [0.500, 0.609] & 95 \\
\addlinespace[1pt]
RIGA Disc & 11/11 & 0.870 & 0.230 & 0.640 [0.570, 0.689] & 95 \\
\addlinespace[1pt]
ISIC-2018 & 11/11 & 0.930 & 0.679 & 0.250 [0.226, 0.276] & 1000 \\
\addlinespace[1pt]
MSD Hippocampus & 11/11 & 0.908 & 0.441 & 0.467 [0.326, 0.648] & 52 \\
\addlinespace[1pt]
MSD Spleen & 11/11 & 0.983 & 0.979 & 0.003 [0.000, 0.007] & 8 \\
\addlinespace[1pt]
MSD Heart & 0/11 & -- & -- & -- & 4 \\
    \bottomrule
  \end{tabular*}

  \par\medskip
  \par\noindent\textbf{(b) Encoder-family pool selection}\par\smallskip
  \begin{tabular*}{\linewidth}{@{\extracolsep{\fill}}l c E c c@{}}
    \toprule
    \bfseries \textbf{Task} & \bfseries \makecell[c]{Low/\\splits} & \multicolumn{1}{c}{\bfseries \makecell[c]{Share (\%)\\{}[95\% CI]}} & \bfseries \textbf{Unsupported} & \bfseries \makecell[c]{Supported-only\\share (\%)} \\
    \midrule
    Kvasir & 200/200 & 100.0 [98.2, 100.0] & 0 & 100.0 \\
\addlinespace[1pt]
ACDC LV & 200/200 & 100.0 [98.2, 100.0] & 0 & 100.0 \\
\addlinespace[1pt]
BraTS FG & 200/200 & 100.0 [98.2, 100.0] & 0 & 100.0 \\
\addlinespace[1pt]
RIGA Cup & 103/200 & 51.5 [44.3, 58.6] & 0 & 51.5 \\
\addlinespace[1pt]
RIGA Disc & 4/200 & 2.0 [0.5, 5.0] & 1 & 2.0 \\
\addlinespace[1pt]
ISIC-2018 & 200/200 & 100.0 [98.2, 100.0] & 0 & 100.0 \\
\addlinespace[1pt]
MSD Hippocampus & 3/200 & 1.5 [0.3, 4.3] & 60 & 2.1 \\
\addlinespace[1pt]
MSD Spleen & -- & -- & -- & -- \\
\addlinespace[1pt]
MSD Heart & -- & -- & -- & -- \\
    \bottomrule
  \end{tabular*}

\end{table}

\subsection{Encoder-family pool selection}
\label{app:pool-selection}

Table~\ref{tab:poolsel}(b) scores family-selected four-member pools on the evaluation halves of seven tasks, without assuming that different encoder families supply independent members.

For each of seven tasks (the five primary tasks with their stated splits, plus ISIC-2018 \citep{isic2019,ham10000,isic_msk2017} and MSD Hippocampus \citep{msd2022,msd2019simpson}%
), \PoolSelSplitsKvasir\ random calibration/evaluation half-splits are drawn. On each split, the four-member pool selected by encoder-family diversity on the calibration half is scored on the evaluation half at a threshold fixed from calibration. A supported split counts as low when $M_{\mathrm{eff}}<2$; unsupported splits count as not low. ACDC LV and BraTS FG are split by patient, while $M_{\mathrm{eff}}$ is computed from slice-level failure indicators within each evaluation half. ISIC uses image-level splits; images of a known patient can occur in both halves.

Calibration means weight patients equally on ACDC LV and BraTS FG. A bundle passes the 0.30 mean-Dice floor on the mean of its four heads, all of which remain candidates. We rank heads by calibration mean, select each family's best head and resolve ties lexicographically. The threshold is the empirical 33rd percentile of the best functional head's calibration case scores (slice scores on ACDC/BraTS), not of patient-averaged scores. Each task's frequency interval is a two-sided 95\% Clopper--Pearson interval over independently randomised splits; it describes Monte Carlo uncertainty conditional on the observed cohort, not uncertainty over new patients or datasets.

\FloatBarrier
\section{Broader impacts and responsible AI}
\label{app:broader-impacts}

\paragraph{Societal impact.} This is a retrospective analysis of evaluation evidence; it does not propose, deploy or recommend any clinical AI system. Its intended positive impact is better-informed evaluation of model pools for medical image segmentation: reporting absolute dependence and threshold-specific effective size alongside accuracy exposes shared failures that mean accuracy and nominal pool size do not reveal. The main risk is over-interpretation: low measured dependence on these retrospective cohorts could be mistaken for evidence that a pool is safe to deploy, although all statistics are conditional on the evaluated pools, cohorts, readouts and thresholds (Section~\ref{sec:discussion}) and are not validated for clinical deployment decisions. Because we perform no subgroup analysis, pools with low overall dependence may still share failures within patient subgroups.

\paragraph{Release and intended use.} The release (Section~\ref{sec:artifact}) contains aligned per-case scores for the reported analyses, split identifiers, panel metadata, illustrative figures, scripts with reference outputs, a SHA-256 manifest and Croissant metadata, and a reproduction guide that maps commands and output columns to the tables and figure panels. It also contains the study's preprocessing, feature-extraction, training and evaluation code.

Per-case Dice traces retain upstream de-identified subject/patient/slice IDs and deterministic hashed RIGA alignment IDs solely for case alignment and subject-level resampling. The released traces contain no machine paths, run-environment identifiers or embedded copies of source code, and no images, masks, per-pixel predictions, feature caches, checkpoints or additional demographic annotations. Users must obtain original datasets and models under their upstream terms for retraining or image-level re-evaluation.

The intended use is to recompute case-level dependence statistics from the aligned traces, inspect the aggregate evidence and reuse the protocol in evaluation research and in retrospective local redundancy checks of candidate pools (Section~\ref{sec:discussion}). Phikon-derived values are withheld, as explained below. The traces are not intended for individual profiling, re-identification or clinical diagnosis; stable alignment identifiers do not guarantee anonymity.

All upstream datasets (Kvasir-SEG, ACDC, BraTS 2024 GLI, RIGA+, ISIC-2018, MSD and AMOS22) are publicly released under their respective licences and de-identification procedures. Demographic annotations are not harmonised across the evaluated datasets; we perform no subgroup or fairness analysis and claim no subgroup robustness.

\paragraph{Asset licences.} Table~\ref{tab:licences} lists the terms under which the datasets are used for training and evaluation and under which the backbones, checkpoints and software are used; backbone, checkpoint and software licences follow the named repositories or model cards. Data attributions are given in Appendix~\ref{app:data-notes}, and software releases are specified in Appendices~\ref{app:encoder-wrappers} and~\ref{app:amos}, Table~\ref{tab:licences} and the repository's pinned requirement files. BraTS image-derived examples are confined to the illustrative figures.

\begin{table}[p]
  \caption{Licences and terms of use of third-party assets. Backbone, checkpoint and software terms follow the named repositories or model cards.}
  \label{tab:licences}
  \centering
  \begin{tabularx}{\linewidth}{@{}>{\raggedright\arraybackslash}p{0.34\linewidth}Y@{}}
    \toprule
    \bfseries Asset & \bfseries Licence or terms of use \\
    \midrule
    \multicolumn{2}{@{}l}{\itshape Datasets (training and evaluation)} \\
    RIGA+ & CC BY 4.0 (Zenodo record \href{https://doi.org/10.5281/zenodo.6325549}{doi:10.5281/zenodo.6325549}); upstream terms also apply: RIGA CC BY-NC 4.0 (\href{https://doi.org/10.7302/Z23R0R29}{Deep Blue Data}) and MESSIDOR (research and educational use only; no copying or redistribution) \\
    ACDC & CC BY-NC-SA 4.0; non-commercial scientific research only (licence terms distributed with the dataset) \\
    Kvasir-SEG & Research and educational use only; commercial use requires prior written permission (Simula) \\
    BraTS 2024 & Post-challenge CC BY-NC terms, via Synapse \\
    MSD & CC BY-SA 4.0 \\
    ISIC-2018 Task~1 & CC0 for the \href{https://challenge.isic-archive.com/data/}{challenge bulk distribution}; current Archive metadata list per-image terms \\
    AMOS22 & CC BY 4.0 \\
    \addlinespace
    \multicolumn{2}{@{}l}{\itshape Backbones and checkpoints} \\
    DINOv2 & Apache-2.0 (code and weights) \\
    CLIP ViT-B/16, ViT-B/32, ViT-L/14 & Weights MIT; \texttt{open\_clip} 3.3.0 code MIT \\
    BiomedCLIP & Checkpoint MIT \\
    MAE-B & CC BY-NC 4.0 \\
    DeiT-B & Apache-2.0 \\
    ConvNeXt-T/S/B, EfficientNet-B0, ResNet-18/34/50/101 & torchvision \texttt{IMAGENET1K\_V1} weights; torchvision code BSD-3-Clause; pretrained weights may carry dataset-derived terms; original ConvNeXt code MIT, EfficientNet code Apache-2.0 \\
    RETFound & Checkpoint gated; CC BY-NC 4.0 \\
    Phikon & Owkin non-commercial licence \\
    RAD-DINO & MIT; research use only per its model card \\
    BiomedParse v1 & Weights gated; CC BY-NC-SA 4.0; code Apache-2.0; research and development use only \\
    \addlinespace
    \multicolumn{2}{@{}l}{\itshape Software} \\
    timm 1.0.27 & Code Apache-2.0; checkpoint-specific terms apply to pretrained weights \\
    FocalNet & Code MIT \\
    MONAI 1.5.0 (AMOS22 scoring) & Apache-2.0 \\
    NiBabel 5.4.2 (5.3.2 for AMOS22 scoring) & Code MIT \\
    nnU-Net v2.6.2 & Code Apache-2.0 (\href{https://github.com/MIC-DKFZ/nnUNet/blob/v2.6.2/LICENSE}{licence file}) \\
    MRSegmentator & Code Apache-2.0 (\href{https://github.com/hhaentze/MRSegmentator/blob/0ad135b7b22063a6bbfae6a0cab0a4702914fbd2/License.txt}{licence file}) \\
    TotalSegmentator 2.4.0 & Code Apache-2.0 (\href{https://github.com/wasserth/TotalSegmentator/blob/2e0c20058df8acb17acd54a7adf25bf7a0a33c90/LICENSE}{licence file}); the release documentation states that the \texttt{total} and \texttt{total\_mr} tasks are free for any use \\
    \bottomrule
  \end{tabularx}
\end{table}

Authored code is provided under Apache-2.0; upstream assets and model-derived data retain their own terms. Phikon-derived outputs are subject to the Owkin licence, which restricts use and sharing to qualifying non-commercial research; they are not included in the release. Included scalar traces retain applicable attribution and use conditions, documented alongside the data.

\FloatBarrier

\end{document}